\documentclass[10pt,twocolumn,letterpaper]{article}

\usepackage{iccv}
\usepackage{times}
\usepackage{epsfig}
\usepackage{graphicx}
\usepackage{amsmath}
\usepackage{amssymb}
\usepackage{multirow}
\usepackage{siunitx}
\usepackage{adjustbox}
\usepackage{booktabs, makecell, multirow}

\usepackage[normalem]{ulem}
\usepackage{authblk}
\usepackage{url}

\makeatletter
\newcommand\email[2][]%
   {\newaffiltrue\let\AB@blk@and\AB@pand
      \if\relax#1\relax\def\AB@note{\AB@thenote}\else\def\AB@note{\relax}%
        \setcounter{Maxaffil}{0}\fi
      \begingroup
        \let\protect\@unexpandable@protect
        \def\thanks{\protect\thanks}\def\footnote{\protect\footnote}%
        \@temptokena=\expandafter{\AB@authors}%
        {\def\\{\protect\\\protect\Affilfont}\xdef\AB@temp{#2}}%
         \xdef\AB@authors{\the\@temptokena\AB@las\AB@au@str
         \protect\\[\affilsep]\protect\Affilfont\AB@temp}%
         \gdef\AB@las{}\gdef\AB@au@str{}%
        {\def\\{, \ignorespaces}\xdef\AB@temp{#2}}%
        \@temptokena=\expandafter{\AB@affillist}%
        \xdef\AB@affillist{\the\@temptokena \AB@affilsep
          \AB@affilnote{}\protect\Affilfont\AB@temp}%
      \endgroup
       \let\AB@affilsep\AB@affilsepx
}
\makeatother

\usepackage[breaklinks=true,bookmarks=false]{hyperref}

\iccvfinalcopy % *** Uncomment this line for the final submission

\def\iccvPaperID{****} % *** Enter the ICCV Paper ID here

\begin{document}

%%%%%%%%% TITLE
\title{Multi-Scale Tensorial Summation and Dimensional Reduction Guided Neural Network for Edge Detection}

\author[1]{Lei Xu}
\author[1]{Mehmet Yamac}
\author[1, 2]{Mete Ahishali}
\author[1]{Moncef Gabbouj}

\affil[1]{Faculty of Information Technology and Communication Sciences, Tampere University, Finland}
\email{\url{{lei.xu, mehmet.yamac, moncef.gabbouj}@tuni.fi}}
\affil[2]{School of Forest Sciences, Faculty of Science, Forestry and Technology, University of Eastern Finland}
\email{\url{{mete.ahishali}@uef.fi}}

%command\Authands{ and }

\maketitle
%\thispagestyle{empty}
%%%%%%%%% ABSTRACT
%%%%%%%%% ABSTRACT

\begin{abstract}
    Edge detection has attracted considerable attention thanks to its exceptional ability to enhance performance in downstream computer vision tasks. In recent years, various deep learning methods have been explored for edge detection tasks resulting in a significant performance improvement compared to conventional computer vision algorithms. In neural networks, edge detection tasks require considerably large receptive fields to provide satisfactory performance. In a typical convolutional operation, such a large receptive field can be achieved by utilizing a significant number of consecutive layers, which yields deep network structures. Recently, a Multi-scale Tensorial Summation (MTS) factorization operator was presented, which can achieve very large receptive fields even from the initial layers. In this paper, we propose a novel MTS Dimensional Reduction (MTS-DR) module guided neural network, \textbf{MTS-DR-Net}, for the edge detection task. The MTS-DR-Net uses MTS layers, and corresponding MTS-DR blocks as a new backbone to remove redundant information initially. Such a dimensional reduction module enables the neural network to focus specifically on relevant information (i.e., necessary subspaces). Finally, a weight U-shaped refinement module follows MTS-DR blocks in the MTS-DR-Net. We conducted extensive experiments on two benchmark edge detection datasets: BSDS500 and BIPEDv2 to verify the effectiveness of our model. The implementation of the proposed MTS-DR-Net can be found at \url{https://github.com/LeiXuAI/MTS-DR-Net.git}.
\end{abstract}

%%%%%%%%% BODY TEXT
\section{Introduction}
Edge detection aims to locate edges or boundary pixels of objects from various images, which plays a pivotal role in the area of computer vision. Usually, edges are defined as discontinuities in surface reflectance, illumination, surface normal, and depth \cite{Pu2021RINDNet:Depth} in an image. This research topic has a wide range of applications because edge detection is a critical low-level computer vision task for the beneficiary of higher-level downstream tasks, such as remote sensing image segmentation \cite{Zhang2022ISNet:Detection, You2023Boundary-AwareSegmentation}, image enhancement \cite{Chen2020RBPNET:Image}, and intelligent transportation \cite{Yang2019, Xu2024RevisitingDatasets}. There are several existing challenges of the edge detection task derived primarily from the characteristics of available benchmark datasets. Firstly, the edges defined in an image can arise in different ways \cite{Pu2021RINDNet:Depth} with diverse representations. The second is the uncertainty problem of the benchmark datasets due to multiple annotations, such as BSDS500 \cite{Arbelaes2011ContourSegmentaion} and Multicue \cite{mely2016systematic}. Moreover, the imbalance problem is a significant characteristic of these benchmark datasets that needs more consideration for robustness \cite{cetinkaya2024ranked, Xu2024RevisitingDatasets}. 

\begin{figure}[t]
\begin{center}

   \includegraphics[width=0.8\linewidth]{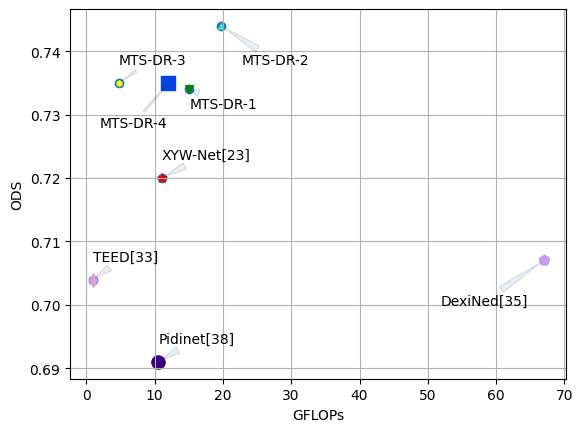}
\end{center}
   \caption{ODS vs. GFLOPs on BSDS500 dataset under ``Thin" setting.}
\label{fig:ods_flops}
\end{figure}

To effectively extract edge features from images captured in different environments, intense efforts have been devoted by the studies \cite{Canny1986ADetection, Wang2019DeepTasks, Pu2021RINDNet:Depth, cetinkaya2024ranked}. Generally, edge detection algorithms can be classified as traditional methods and deep learning-based methods. Traditional edge detection techniques are mainly based on differential operators or high pass filtration \cite{Gao2010AnDetection}, such as the Sobel operator \cite{Gao2010AnDetection} or the Canny operator \cite{Singh2015ComparisonTechniques, Kalra2016ADetection}. The widely known shortcomings of these operators are computational complexity and lack of robustness \cite{Gao2010AnDetection}. Then, various variants have been proposed that cooperate with local and/or global information from images to learn richer descriptors of edges \cite{Arbelaes2011ContourSegmentaion, ren2005scale, martin2004}. 

With the prosperous development of deep neural networks (DNN) during the last decade, various multi-scale DNN structures have been widely investigated for edge detection tasks \cite{xu2017learning, Xie2017Holistically-NestedDetection, Soria2020DenseDetection}. In these methods, the edge features are learned from coarse to fine levels by producing a final output through the fusion of multi-scale additional side estimations. Multi-scale structures have remarkable characteristics of high computational complexity, redundant parameters, and the need for pre-training \cite{Liu2019RicherDetection, Sun2021PixelDetection}. To address computational complexity, novel light-weight DNN structures are proposed to trade-off between accuracy and efficiency, such as Pixel Difference Networks (PiDiNet) \cite{Sun2021PixelDetection}, Tiny and Efficient Edge Detector (TEED) \cite{Soria2023TinyGeneralization}, XYW-Net \cite{pang2024bio}, and Lightweight Dense CNN (LDC) \cite{soria2022ldc}. Moreover, the crispness of the results is also taken into account in DiffusionEdge \cite{ye2024diffusionedge} and Ye's work \cite{ye2023delving}. In addition, label uncertainty and imbalance problems are explored using ranking-based losses in \cite{cetinkaya2024ranked}, pixel-level noise transitions \cite{xuan2023pnt}, and edge granularity estimation \cite{zhou2024muge}. The main shortcomings of these state-of-the-art (SOTA) proposals are as follows: (i) it is difficult to ensure both the efficiency of the model and high qualitative results \cite{ye2024diffusionedge, pu2022edter}, (ii) a post-processing step, i.e., non-maximum suppression morphological scheme (NMS) \cite{Soria2023TinyGeneralization, Sun2021PixelDetection, pang2024bio} is used to ensure the crispness and precision of the results, and (iii) transfer learning is indispensable in the current SOTA for challenging BSDS500 dataset.

To overcome the above drawbacks, we propose a novel Multi-Scale Tensorial Summation and Dimensional Reduction (MTS-DR) guided Neural Network for edge detection called the MTS-DR-Net. The proposed architecture comprises a novel MTS-DR module as the backbone and a refinement network. The MTS layer is the fundamental unit of the MTS-DR module, which is inspired by the summation of tensors on multiple scales \cite{Mehmet2023} based on the Tucker-Decomposition \cite{Tucker1966SomeMN}. The MTS layer works as a compact substitute for a complex CNN or transformer structure. The MTS-DR module is feasible to learn required submanifolds using multiple patch sizes strategy as input and generate the summation of multi-scale feature maps as output. In addition, a Multi-Head Gate operation (MHG) is introduced into the MTS-DR module to add nonlinearity. In general, the main contributions of our proposal are as follows.
\begin{itemize}
\item To the best of our knowledge, this is a pioneer work utilizing tensorial summation strategy to solve edge detection tasks. 
\item Compared to the SOTA methods \cite{Sun2021PixelDetection, soria2022ldc, Soria2023TinyGeneralization, pang2024bio, Soria2020DenseDetection}, the novelty of our model is to first remove unnecessary features rather than extract the necessary features directly from the input. 
\item The proposed MTS-DR-Net approach can satisfy the efficiency of the model and the precision of the results simultaneously without any post-processing step. 
\item The proposed approach can surpass previous methods on benchmark datasets even without the use of transfer learning.
\end{itemize}

%-------------------------------------------------------------------------

\section{Related works}

\subsection{Edge Detection}
As a long-standing low-level vision task, early edge detection works aim to utilize gradients or derivative operators to discrete edge pixels from an image. For example, Canny proposes a Canny edge detector based on the Gaussian filter for arbitrary edges \cite{Canny1986ADetection}. In order to address the drawbacks existing in gradient operators, Martin et al. \cite{martin2004} use a combination of brightness and texture cues for local boundary detection with pixel-level posterior probability maps. Bertasius et al. \cite{bertasius2015deepedge} propose a unified multi-scale DNN approach, which uses high-level object features with a multi-scale bifurcated convolutional neural network. A holistically-nested network \cite{Xie2017Holistically-NestedDetection} is proposed with a single-stream convolutional neural network along with multiple side outputs \cite{Xie2017Holistically-NestedDetection}. The multi-scale structures \cite{ Liu2019RicherDetection} are popular for learning richer representations for edge detection tasks. More recently, lightweight structures have been explored with promising performance. For example, Sun et al. \cite{Sun2021PixelDetection} propose a lightweight pixel difference convolutions network (PiDiNet) derived from the extended local binary patterns for edge detection. The proposed TEED \cite{Soria2023TinyGeneralization} with double fusion and double loss is notable for its simplicity and efficiency for edge detection. More recently, Pang et al. propose a lightweight encoding-decoding structure named XYW-Net \cite{pang2024bio} for edge detection. The XYW-Net is designed based on physiological mechanisms with three parallel pathways. 

\subsection{Efficiency in Neural Networks}
Efficiency is an essential concern in the success of the deployment of neural networks to solve real-world tasks on edge devices. Hence, researchers have investigated solutions to improve the efficiency of the model without sacrificing the performance of the model. Firstly, various compact neural network structures have been explored by optimizing the redundancy and complexity of standard network structures. MobileNets \cite{Sanjay2019MobileNetTinyAD} proposed by Howard et al. is built on the basis of a streamlined architecture for mobile and embedded vision applications. The proposed structure uses the depthwise separable convolutions strategy \cite{sifre2014rigid}, which can be trained with highly optimized general matrix multipliers. Han et al. propose a lightweight GhostNet \cite{Han2019GhostNetMF}, in which a novel Ghost module generates more features by fewer parameters, and a series of lightweight linear transformations generates more feature maps. Sun et al. propose a dynamic group convolution \cite{su2020dynamic} to select connection paths from input channels to feature selector groups adaptively. The structure of Mixture-of-Experts (MoE) proposed by Jacobs et al. \cite{Jacobs1991AdaptiveMO} aims to learn a complete set of training data in different subsets with separate networks initially. With the rise of large models, the MoE strategy has been widely used to scale the capacity of the model while suppressing the surge in computational costs \cite{du2022glam}. 

%%%%%%%%%%%%%%%%%%%%%%%%%%%%%%%%%%%%%%%%%%mts_layer
\begin{figure}
\begin{center}
    \includegraphics[width=1.\linewidth]{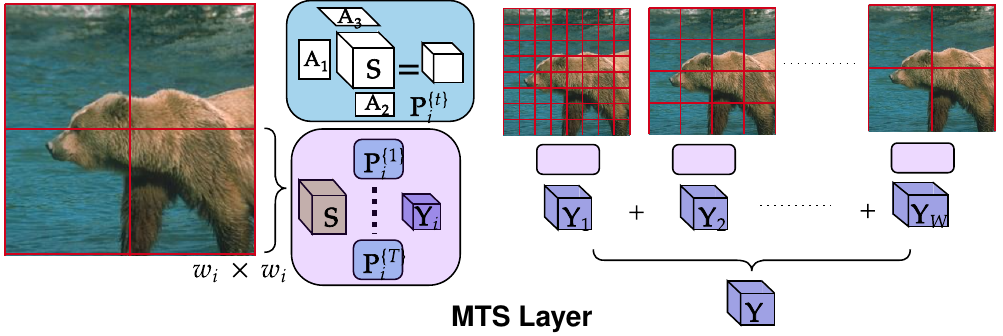}
\end{center}
   \caption{An Example of the MTS Layer with Window Scales [8, 16, 32]}
\label{fig:mts_layer}
\end{figure}

\section{Learnable Tensor Decomposition}
Tensor decomposition \cite{phan2020stable, panagakis2021tensor} is extended from conventional linear operation to multilinear ones, i.e., multilinear dimensional reduction \cite{oh2019high} for higher-order tensors. The tensor decomposition strategy has been intensively explored for deep neural network compression using tensor mappings with efficient network structures and a lower computational cost. To reduce the memory need of the fully-connected layers, Novikov et al. \cite{novikov2015tensorizing} propose tensorizing neural networks on the basis of Tensor-Train decomposition \cite{Oseledets_2011_TensorTrain}. Tucker convolutions \cite{panagakis2021tensor} using the Tucker decomposition \cite{Tucker1966SomeMN} can replace standard convolutional layers with low-rank layers. Phan et al. propose \cite{phan2020stable} a stable compressive CNN with Canonical Polyadic decomposition. Although tensor decomposition can achieve efficient CNN structures \cite{zniyed2024enhanced, yin2021towards}, information loss is an inevitable problem that limits the widespread application of the strategy. More recently, Yamac et al. introduce a generalized tensor summation strategy \cite{Mehmet2023} as a backbone structure, which can directly extract features in the spatial domain. The generalized tensor summation has $T$ number of different map tensors representing the learned features in a multi-linear way, which corresponds to linear representation equivalent in summed factorization of multiple mod products.

%%%%%%%%%%%%%%%%%%%%%%%%%%%%%%%%%%%%This part is for figures
%%%%%%%%%%%%%%%%%%%%%%%%%overall framework
\begin{figure*}
\begin{center}
    \includegraphics[width=1.\linewidth]{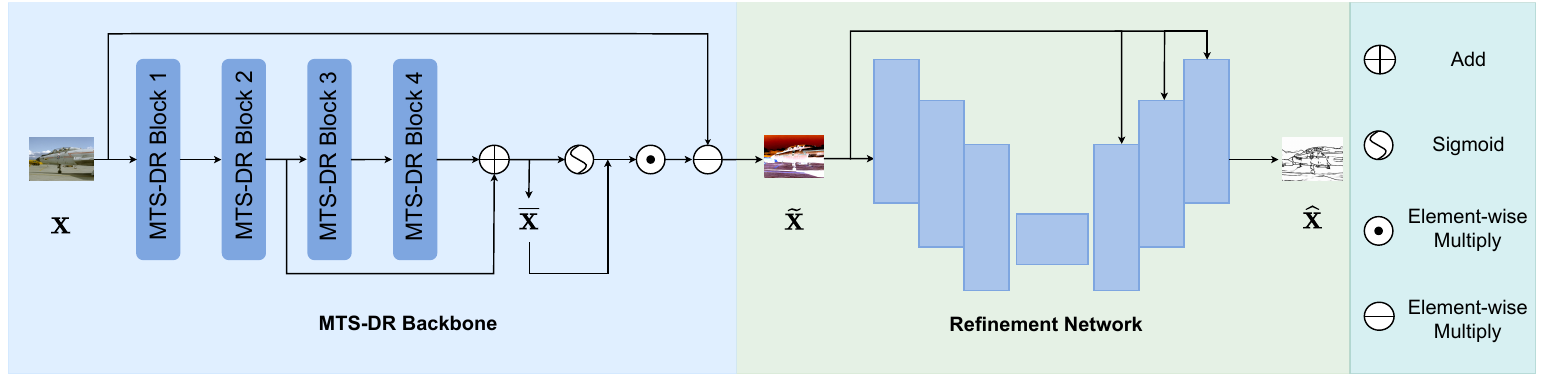}
\end{center}
   \caption{Overall framework of the proposed MTS-DR-Net. It consists of two modules: a MTS-DR backbone and a refinement network.}
\label{fig:overall}
\end{figure*}

\subsection{MTS Layer for multi-linear feature extraction}
Let \(\mathbf{X} \in \mathbb{R}^{n_1 \times n_2 \times \ldots \times n_J}\) represent a multi-dimensional signal, where J indicates the dimensionality of the signal on which we wish to perform a linear operation. For example, in the case of an RGB image, we have \(J=3\) and \(n_3=3\). If \(\mathbf{X}\) is a feature map with dimensions \(H \times W \times C\), then it follows that \(n_1 = H\), \(n_2 = W\), and \(n_3 = C\). In a traditional MLP layer, matrix multiplication is applied on vectorized tensor, \( \text{vec}(\mathbf{X} ) \), i.e., \(MLP(\mathbf{X}), \mathbf{A}  = A \text{vec}(\mathbf{X})\). However, such an operation is infeasible for large-scale signal-like feature maps in computer vision tasks.  In \cite{Mehmet2023}, the authors proposed a tensorial summation factorization of such linear mappings, i.e., 
\begin{equation}
    GTS(X) = \sum_{t=1}^{T} \mathbf{{X}} \times_1 \mathbf{A_1^{(t)}} \times_2 \mathbf{A_2^{(t)}} \ldots \times_{J-1} \mathbf{A_{J-1}^{(t)}} \times_J \mathbf{A_J^{(t)}}, \label{tensorsum1}
\end{equation}
where multi-linear transformation is done via learnable weight matrices, \( \mathbf{A_1^{(t)}}\).
The study \cite{Mehmet2023} was proposed for a learnable compressive sensing system, focusing solely on RGB images for dimensional reduction. However, applying the GTS operation to each feature map in a neural network becomes computationally infeasible when dealing with large-resolution feature maps. To overcome this issue, the authors of \cite{yamaç2025multiscaletensorsummationfactorization} propose to apply GTS operation on patch-wise. They also proposed to apply such an operation for varying patch-sizes in order to enhance the representation power of the new neural network layer which is equipped with a large receptive field and multi-scale view. Mathematically speaking, 
\begin{equation}
\begin{aligned}
MTS(\mathbf{X}) = \sum_{sc=1}^{SC} f^{-1}_{w_{sc}} \Bigg( \sum_{t=1}^{T} & f_{w_{sc}}(\mathbf{X}) 
\times_1 \mathbf{A}_1^{(t,sc)} \\
\times_2 \mathbf{A}_2^{(t,sc)}  
\times_3 \dots 
& \times_{J-1} \mathbf{A}_{J-1}^{(t,sc)} 
\times_J \mathbf{A}_J^{(t,sc)} \Bigg) \label{MTS}
\end{aligned}
\end{equation}
where \(f_{w_{sc}}\) patch-embedding of size \(w_{sc}\), \(f^{-1}_{w_{sc}}\) is inverse of this operation, \(SC\) is the number of different patch (or window sizes) sizes and T is the number of summed tensor in each scale. For instance, if window size list is set as \(w = [8, 16, 32]\), it means that GTS operation with learnable matrices is applied to \(8\times 8\), \(16\times 16\) and \(32\times 32\) paths of the feature map independently.

\subsection{MHG Layer as Activation Free Non-linear Operator}
The study of \cite{yamaç2025multiscaletensorsummationfactorization} proposes a multi-head-gate operation. In recent years, there has been a notable trend within the machine learning community, particularly in the domain of computer vision, to replace traditional non-linear activation functions, e.g., Rectified Linear Units \cite{relu} (ReLU), with gated operations. This shift is primarily due to the considerable improvement in network performance across various applications that the gated operations offer. Gated Linear Units \cite{GLN}  (GLU) are nothing but element-wise multiplication of outputs of two linear units, where at least one of these is applied to a non-linear activation,   
\(
    Gate(\mathcal{X}, g_1, g_2, \sigma) = g_1(\mathcal{X}) \odot \sigma(g_2(\mathcal{X})), \) where \(\odot\) is element-wise multiplication, 
\(g_1\) and \(g_2\) are linear operators, and \(\sigma\) is a non-linear activation function, e.g., Sigmoid. The non-linearity is also applied directly on feature maps in many works, i.e., Gaussian Error Linear Unit \cite{gelu} (GELU) is defined as \(GELU(\mathcal{X}) = \mathcal{X} \odot \Psi(\mathcal{X})\), where
\(\Psi\) is the cumulative distribution function of Gaussian distribution. On the other hand, recent works \cite{Nafnet} shows that simple gate operations, \(SGate(\mathcal{X}, g_1, g_2, \sigma) = g_1(\mathcal{X}) \odot (g_2(\mathcal{X}))\), where \(g_1\) and \(g_2\), are learned linear maps implemented by \(1 \times 1 ~ conv\) are enough for improved performance in many computer vision tasks. 

Such a non-linear operator simply aims to approximate the true nature of curved spaces where the signal of interest lives. We may interpret a gate operation as a second-order Taylor approximation with a constrained Hessian tensor. In addition, while the \(1 \times 1 ~ conv\) are computationally efficient in low-channel settings, they can still be costly when the number of feature maps is large.  To increase the representative power of the gate operation, multi-head-gate operation introduced in \cite{anonymous2025}, applies 
\begin{equation}
    MHG(\mathcal{X}, g_1, g_2) = \sum_{i=1}^{H} \left (  f_{2i} \left ( g_{1i}(\mathcal{X}) \right )\right) \odot \left (  f_{2i}(g_{2i}(\mathcal{X})) \right )
\end{equation}
where, \(H\) is the number of heads, \(g_{ji}\), \(f_{ji}\) are implemented via \(1 \times 1 ~ gconv\) and
\(3 \times 3 ~ dconv\), respectively. In this implementation, the group convolution split feature maps on the channel dimension, and the depthwise convolutions collect information from adjacent pixels. The corresponding MHG layer is illustrated in Figure \ref{fig:mts_dr}.
%------------------------------------------------------------------------

\section{Methods}
In this section, we first describe the overall architecture of MTS-DR-Net. The MTS-DR backbone and the refinement network are then presented in detail. Finally, we describe the loss function.  

%-------------------------------------------------------------------------
\subsection{MTS-DR-Net Framework}

As illustrated in Figure \ref{fig:overall}, the proposed MTS-DR-Net consists of a MTS-DR backbone and a refinement network. The MTS-DR backbone constitutes four MTS-DR blocks, which are built mainly by the MTS layer and the MHG layer shown in Figure \ref{fig:mts_dr}. The proposed backbone aims to eliminate unnecessary major-class features from the input images. The backbone produces an MTS feature map $\mathbf{\widetilde{X}}$ as the input of the refinement network. The refinement network is built using a lightweight U-net architecture with a side fusion of three scales \cite{Xie2017Holistically-NestedDetection, Sun2021PixelDetection} to learn multi-scale edge features from coarse to fine. The basic unit of the refinement network is a convolution bottleneck structure \cite{he2016deep} without residual connection. Finally, the predicted edge map $\mathbf{\widehat{X}}$ is generated by side fusion.

\subsection{MTS-DR Blocks}
\begin{figure}
\begin{center}
    \includegraphics[width=1.\linewidth]{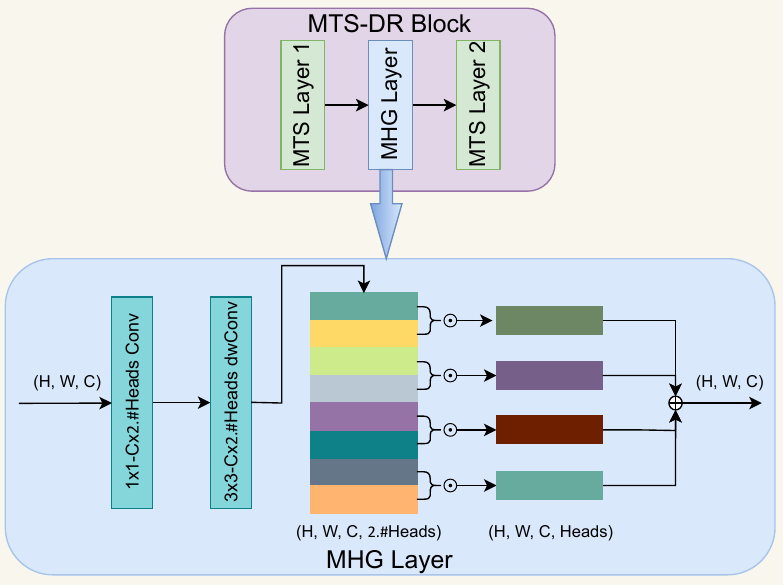}
\end{center}
   \caption{MHG layer and MTS Dimension Reduction (MTS-DR) block implementations}
\label{fig:mts_dr}
\end{figure}

MTS-DR block is an essential unit of the backbone, which consists of an MTS layer followed by an MHG layer and another MTS layer sequentially, as shown in Figure \ref{fig:mts_dr}. Given an image input $\mathbf{X} \in \mathbb{R}^{H \times W \times 3}$, MTS Layer-1 in the MTS-DR block reduces the dimension of $\mathbf{X}$ to $\mathbf{\dot{X}} \in \mathbb{R}^{h \times w \times C_1}$, where $h << H, w << W$. The $h$ and $w$ are calculated based on the compressing ratio. Then, the MHG layer is used to introduce non-linearity in $\mathbf{\dot{X}}$. Finally, MTS Layer 2 maps the $\mathbf{\dot{X}}$ back to the original size $H \times W \times C$ where $C=2 \times C_1$. The output of the MTS-DR backbone is $\mathbf{\overline{X}} \in \mathbb{R}^{H \times W \times 3}$, which is expected to contain major-class features. Then, a MTS feature map is generated as follows:
\begin{equation}
\label{eq:mts_map}
\mathbf{\widetilde{X}} = \mathbf{X} - \textbf{Sigmoid}(\mathbf{\overline{X}}) \otimes \mathbf{\overline{X}}.  
\end{equation}

%%%%%%%%%%%%%%%%%%%%%%%%%%%%%%%%%%%%%%%%%%mts_dr
%%%%%%%%%%%%%%%%%%%%%%%%%%%%%%%%%%%%%%%%%%refinement network
\begin{figure*}
\begin{center}
    \includegraphics[width=0.9\linewidth]{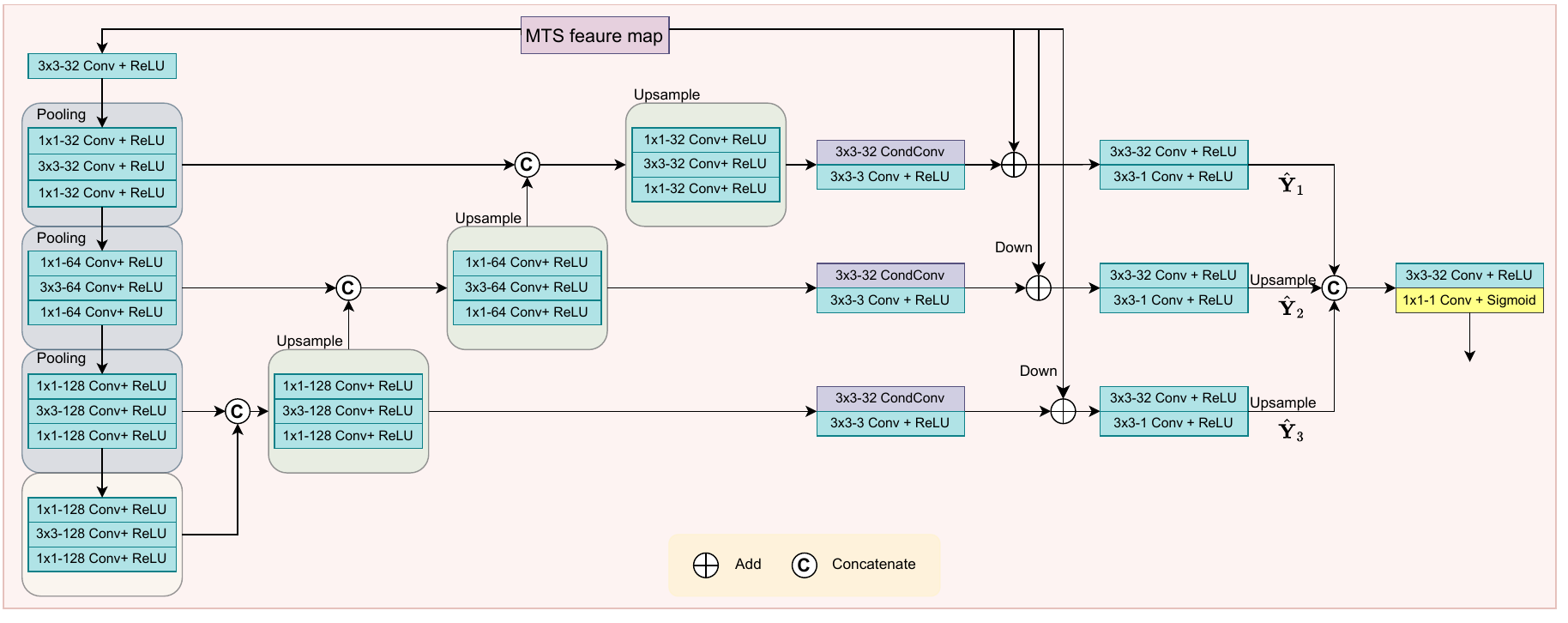}
\end{center}
   \caption{The architecture of the Refinement Network}
\label{fig:refinement}
\end{figure*}
\subsection{Refinement Network}

The U-shaped refinement network comprises an encoder and a decoder with three scales, as shown in Figure \ref{fig:refinement}. The MTS feature map $\mathbf{\Tilde{X}}$ is used as input and side fusion input on three scales. This design is capable of generating a final hierarchical feature map for edge representations from the learned necessary subspace with efficiency. Starting from the input, a $3 \times 3$ convolution layer followed by a ReLU layer is used to extract shallow features from  $\mathbf{\widetilde{X}}$. Then, the shallow features are gradually downsampled at three scales with a growing number of channels for the encoder part. At each scale, the decoder part takes the concatenated features from the corresponding encoder and upsamples the learned features. Especially a conditional convolutional layer (CondConv) with four experts \cite{yang2019condconv} is used for side feature fusion to increase the learning capacity of each scale. The three side feature maps are concatenated after upsampling to the original size as $\mathbf{\widehat{Y}_i}, i = 1, 2, 3$. Finally, a fused edge map $\mathbf{\widehat{Y}}$ is produced using two convolutional layers followed by a Sigmoid function. The values in the edge maps range from 0 to 1 as probability maps. In this work, upsampling is implemented by transposed convolution with strides, and downsampling is max-pooling. 

\subsection{Loss Function}
%-------------------------------------------------------------------------
To highlight the importance of edge pixels in minors, we adopt the weighted binary cross-entropy loss function as in \cite{Liu2019RicherDetection, Sun2021PixelDetection}. The loss $\mathcal{L}$ is calculated with three side edge maps and the fused edge map. The weighted binary cross-entropy is defined as 
\begin{equation}
\label{eq:loss_fun}
\mathcal{L}_{bce}(Y, \hat{Y}) = \begin{cases}
   \alpha \cdot \sum\limits_{i=0}^N log(1 - \hat{y}_i),  &\:\:\: \text{if} \:\: y_i = 0\\
   
   0, &\:\:\: \text{if} \:\: 0 < y_i < \eta \\

   \beta \cdot \sum\limits_{i=0}^N log(\hat{y}_i),  &\:\:\: otherwise,
\end{cases} 
\end{equation}
where $\eta$ is a threshold to mitigate uncertainty problems caused by multiple annotations in the ground-truth edge maps $\mathbf{Y}$ and $\hat{y}_i$ is the probability value on the $i^{th}$ pixel in the predicted edge map $\mathbf{\widehat{Y}}$. The weight parameters $\alpha$ and $\beta$ are used to balance the contribution between positive pixels $y^+$ (edges) and negative pixels $y^-$ (non-edges), defined as:
\begin{equation}
\label{eq:alpha_beta}
\begin{split}
 \alpha & = \lambda \frac{\sum y_i^+}{\sum y_i^+ + \sum y_i^-}, \\
 \beta & = \frac{\sum y_i^-}{\sum y_i^+ + \sum y_i^-}.
\end{split}
\end{equation}
Then the total loss function is 
\begin{equation}
\label{eq:loss}
\mathcal{L} = \sum\limits_{i=1}^3 \mathcal{L}_{bce}(Y, \widehat{Y}_i) + \mathcal{L}_{bce}(Y, \widehat{Y})
\end{equation}

\section{Experiments}
We evaluate the MTS-DR-Net on two benchmark datasets: BSDS500 \cite{Arbelaes2011ContourSegmentaion} and BIPEDv2 \cite{Soria2020DenseDetection} using statistics metrics and semantic segmentation metrics. The experimental results are presented with four competing methods: TEED \cite{Soria2023TinyGeneralization}, Pidinet \cite{Sun2021PixelDetection}, DexiNed \cite{Soria2020DenseDetection}, and XYW-Net \cite{pang2024bio}.
%------------------------------------------------------------------------

%%%%%%%%%%%%%%%%%%%%%%%%%%%%%%%%%%%%%%%%%%%%%%%%%%%%%%%%%%%%%%%%%%%%%%%%%
%%%%%%%%%%%%%%%%%%%%%%%%%%%%%Figures
%%%%%%%%%%%%%%%%%%%%%%%%%%%%%%%%%%%%%%%%%%%%%%%%%%%%%%%%%%%%%%%%%%%%%%%%%%
 %MTS MAP ON BSDS500
%%%%%%%%%%%%%%%%%%%%%%%%%%%%%%%%%%%%%%%%%%%%%%%%%%%%%%%%%%%%%%%%%%%%%%%%%%
%bsds
\begin{figure*}
\centering
\footnotesize
\renewcommand{\tabcolsep}{1pt} % adjust horizontal space
\renewcommand{\arraystretch}{0.2} % adjust vertical space
\begin{tabular}{cccccccc}
    \raisebox{2.0\normalbaselineskip}[0pt][0pt]{\rotatebox[origin=c]{0}{}} &  
    \includegraphics[width=0.12\linewidth, height=2.2cm]{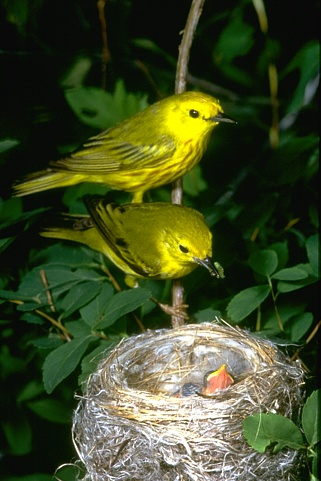} &
    \includegraphics[width=0.12\linewidth, height=2.2cm]{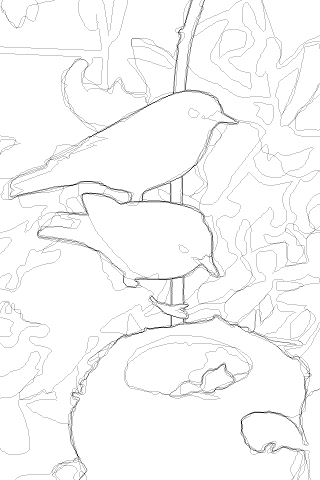} & 
    \includegraphics[width=0.12\linewidth, height=2.2cm]{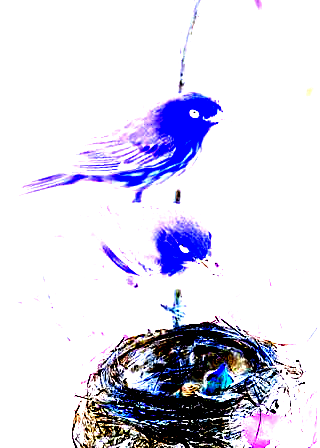} & 
    \includegraphics[width=0.12\linewidth, height=2.2cm]{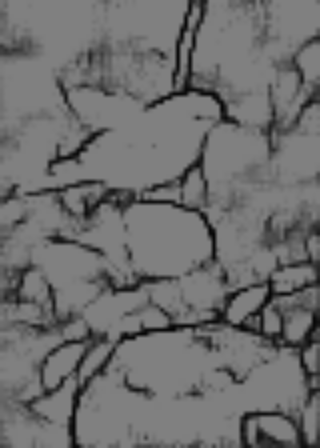} & 
    \includegraphics[width=0.12\linewidth, height=2.2cm]{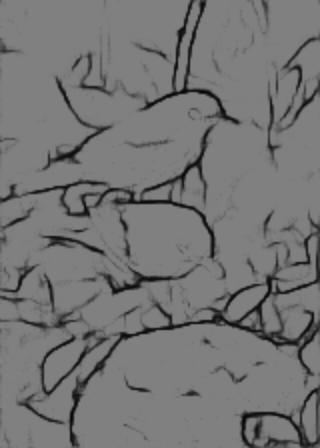} &
    \includegraphics[width=0.12\linewidth, height=2.2cm]{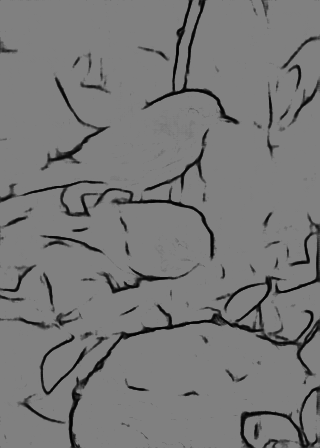} & 
    \includegraphics[width=0.12\linewidth, height=2.2cm]{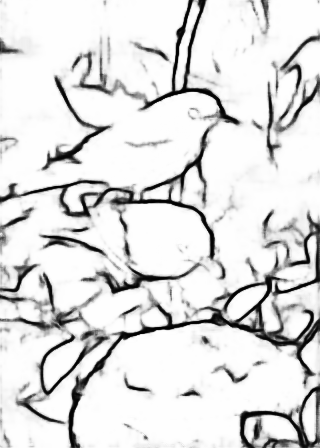}
    \\
    
\iffalse
    \raisebox{2.0\normalbaselineskip}[0pt][0pt]{\rotatebox[origin=c]{0}{}} &  
    \includegraphics[width=0.12\linewidth, height=2.5cm]{figs/bsds/maps/140088.jpg} &
    \includegraphics[width=0.12\linewidth, height=2.5cm]{figs/bsds/gt/140088.png} & 
    \includegraphics[width=0.12\linewidth, height=2.5cm]{figs/bsds/maps/140088_mts.png} & 
    \includegraphics[width=0.12\linewidth, height=2.5cm]{figs/bsds/maps/140088_side1.png} & 
    \includegraphics[width=0.12\linewidth, height=2.5cm]{figs/bsds/maps/140088_side2.png} &
    \includegraphics[width=0.12\linewidth, height=2.5cm]{figs/bsds/maps/140088_side3.png} & 
    \includegraphics[width=0.12\linewidth, height=2.5cm]{figs/bsds/maps/140088.png}
    \\
\fi
    \raisebox{2.0\normalbaselineskip}[0pt][0pt]{\rotatebox[origin=c]{0}{}} &  
    \includegraphics[width=0.12\linewidth, height=2.2cm]{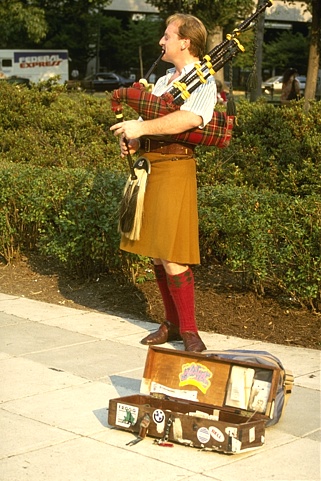} &
    \includegraphics[width=0.12\linewidth, height=2.2cm]{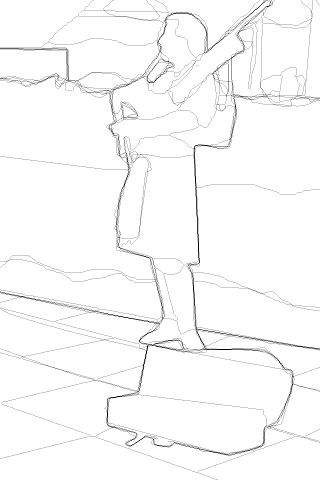} & 
    \includegraphics[width=0.12\linewidth, height=2.2cm]{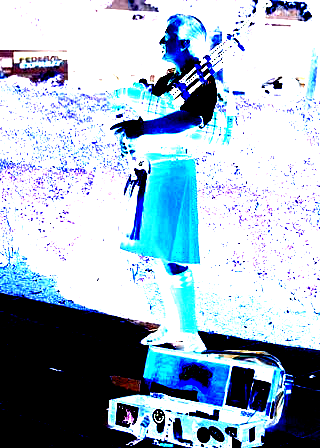} & 
    \includegraphics[width=0.12\linewidth, height=2.2cm]{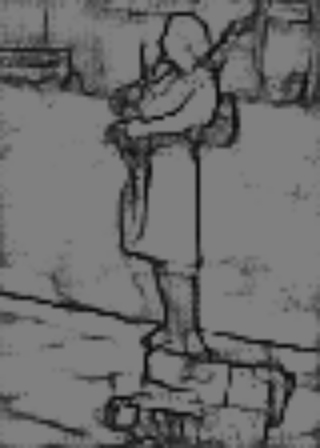} & 
    \includegraphics[width=0.12\linewidth, height=2.2cm]{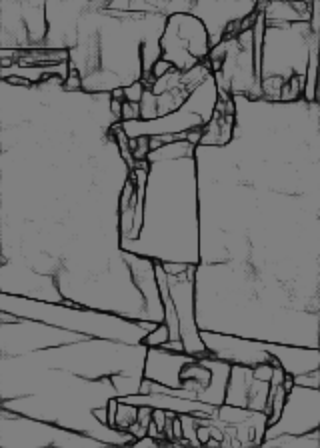} &
    \includegraphics[width=0.12\linewidth, height=2.2cm]{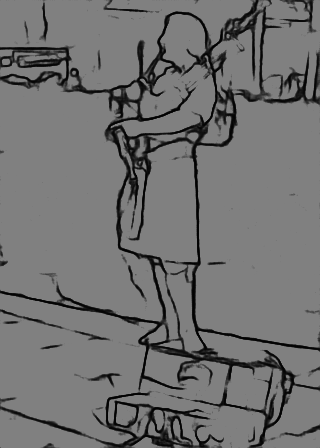} & 
    \includegraphics[width=0.12\linewidth, height=2.2cm]{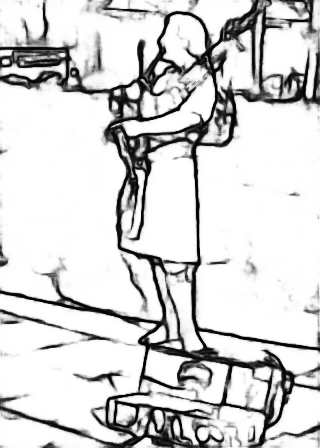}
    \\

    &(a) Original image. &(b) Ground truth. &(c) MTS feature map. 
    & (d) Side 1.& (e) Side 2. & (f) Side 3.& (g) Fused output
\end{tabular}

\caption{ \footnotesize {\textbf{Some examples of feature maps on BSDS500 with MTS-DR-1.} }}
\label{fig:feature_bsds}
\end{figure*}

%%%%%%%%%%%%%%%%%%%%%%%%%%%%%%%%%%%%%%%%%%%%%%%%%%%%%%%%%%%%%%%%%%%%%%%%%%
 %MTS MAP ON BIPED
%%%%%%%%%%%%%%%%%%%%%%%%%%%%%%%%%%%%%%%%%%%%%%%%%%%%%%%%%%%%%%%%%%%%%%%%%%
%biped
\begin{figure*}
\centering
\footnotesize
\renewcommand{\tabcolsep}{1pt} % adjust horizontal space
\renewcommand{\arraystretch}{0.2} % adjust vertical space
\begin{tabular}{cccccccc}
    \raisebox{2.5\normalbaselineskip}[0pt][0pt]{\rotatebox[origin=c]{0}{}} &  
    \includegraphics[width=0.135\linewidth]{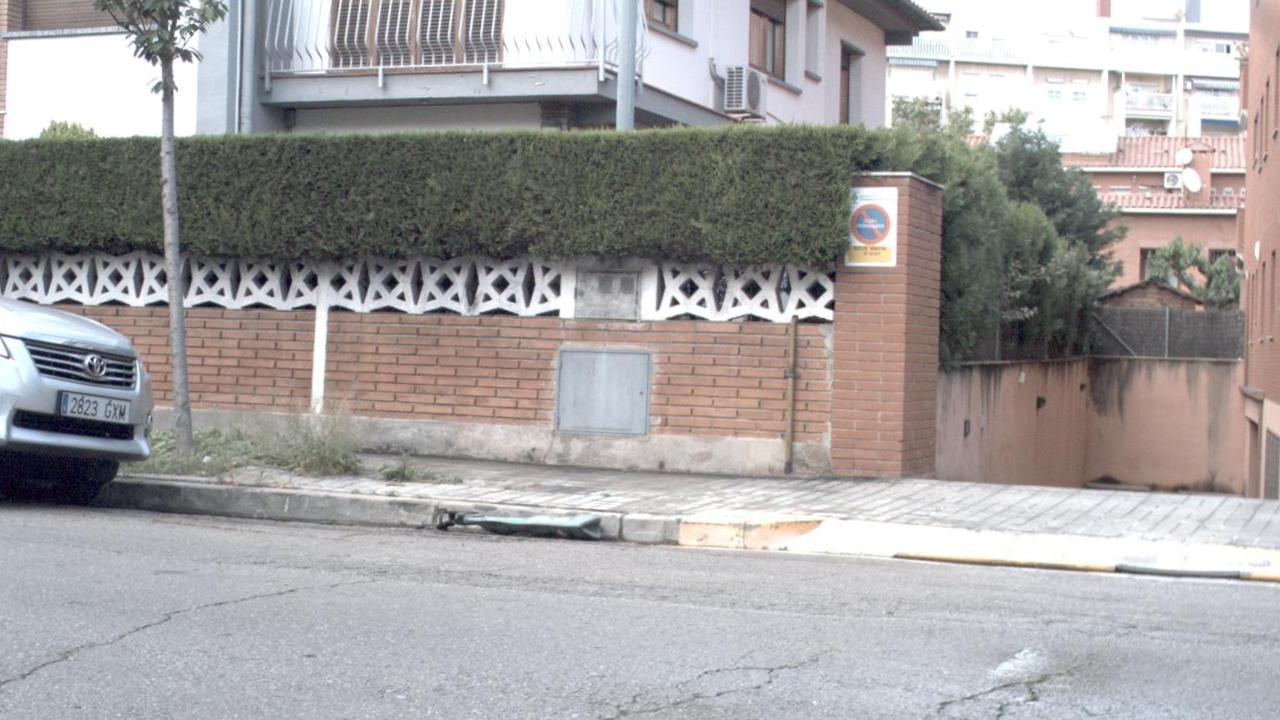} &
    \includegraphics[width=0.135\linewidth]{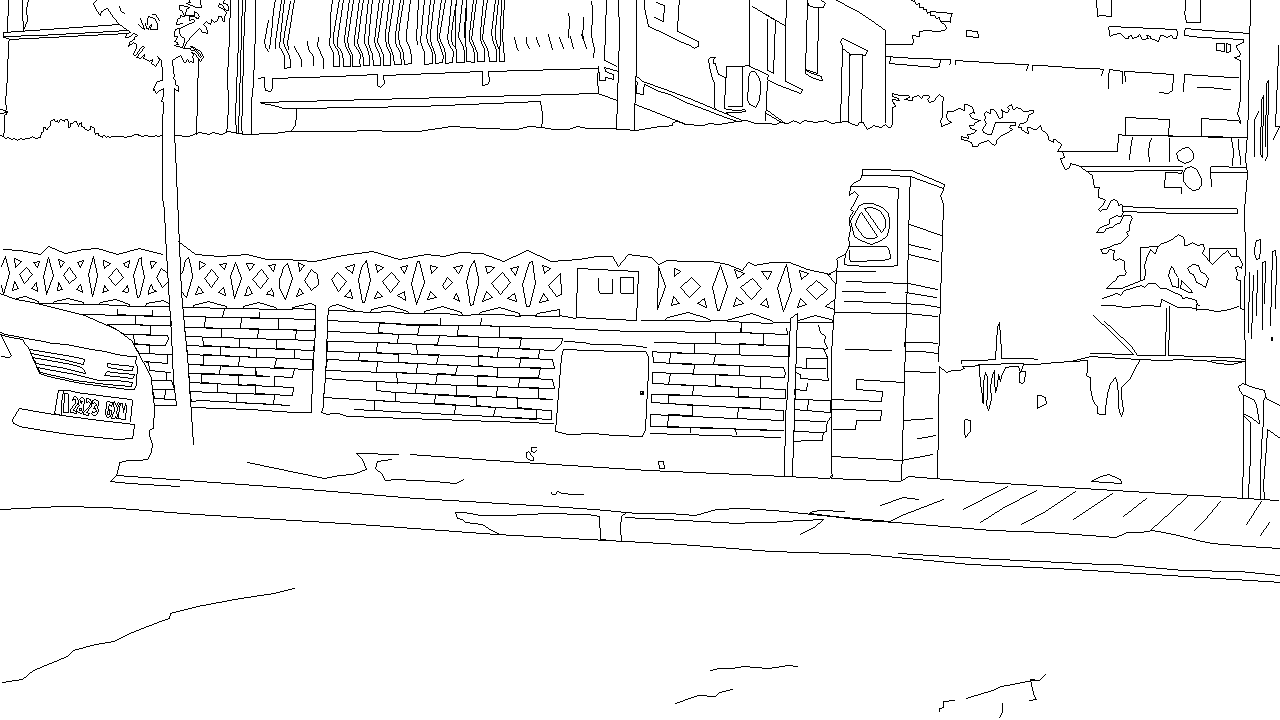} & 
    \includegraphics[width=0.135\linewidth]{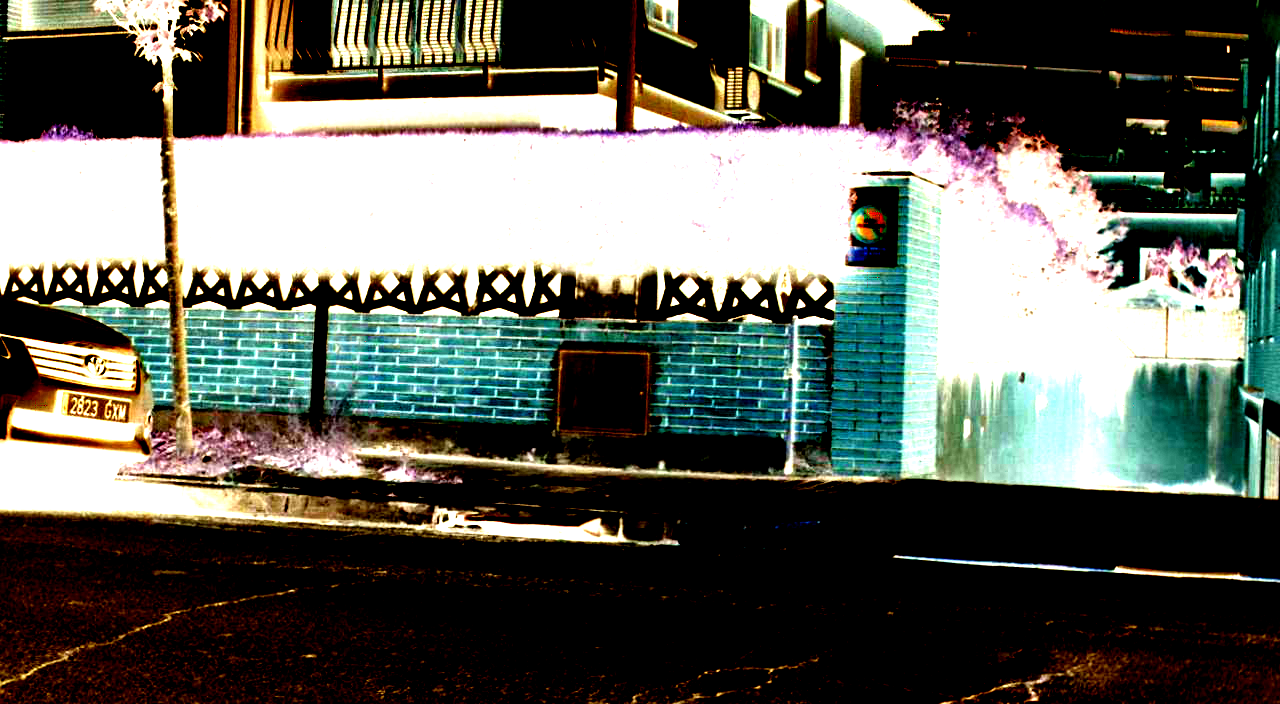} & 
    \includegraphics[width=0.135\linewidth]{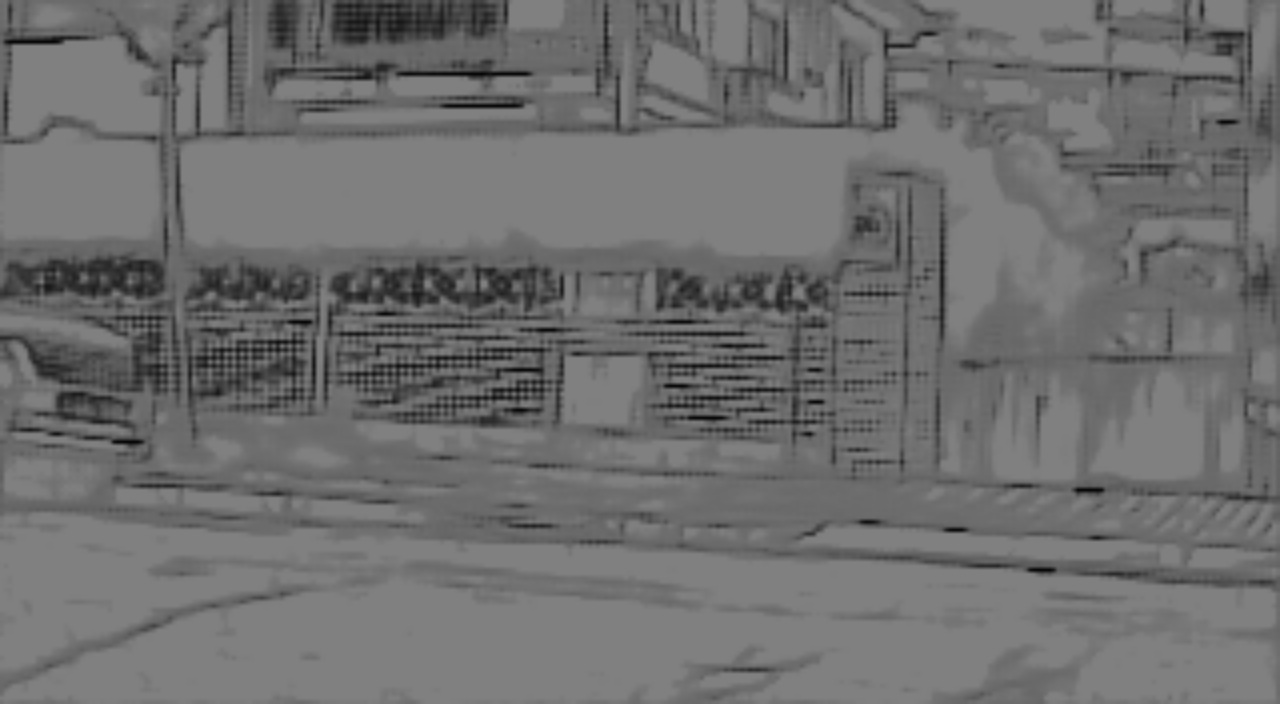} & 
    \includegraphics[width=0.135\linewidth]{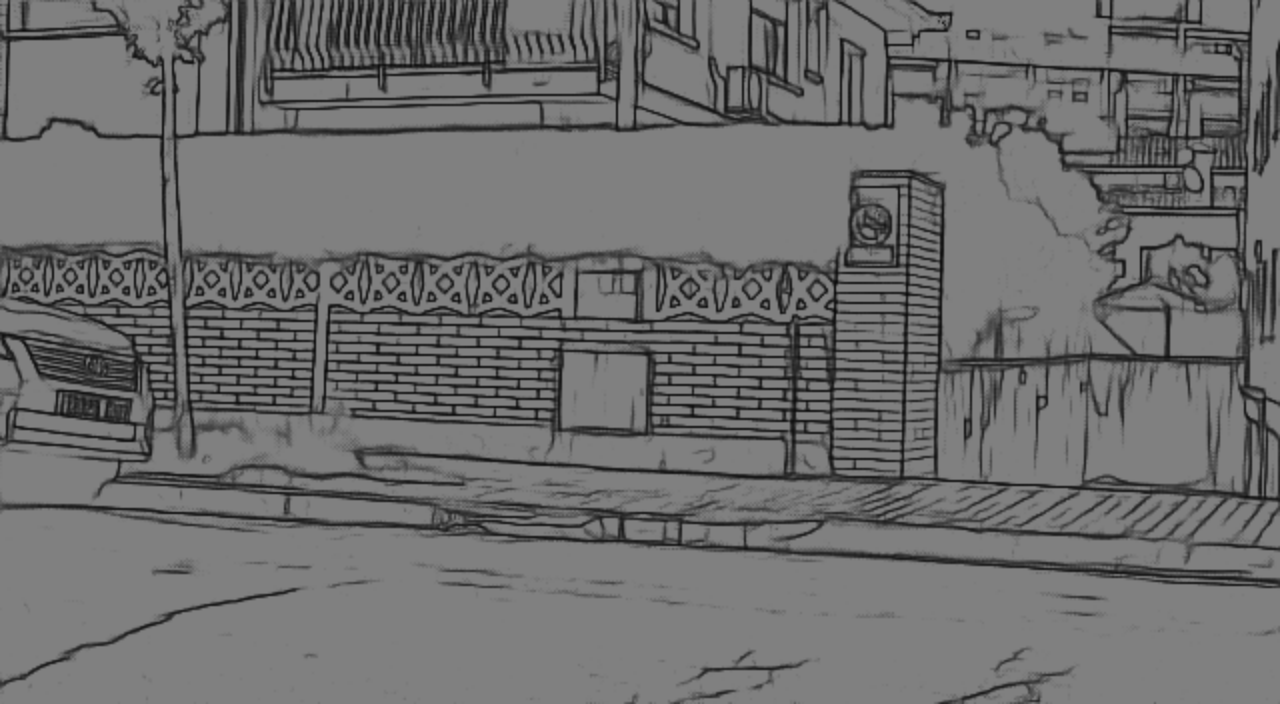} &
    \includegraphics[width=0.135\linewidth]{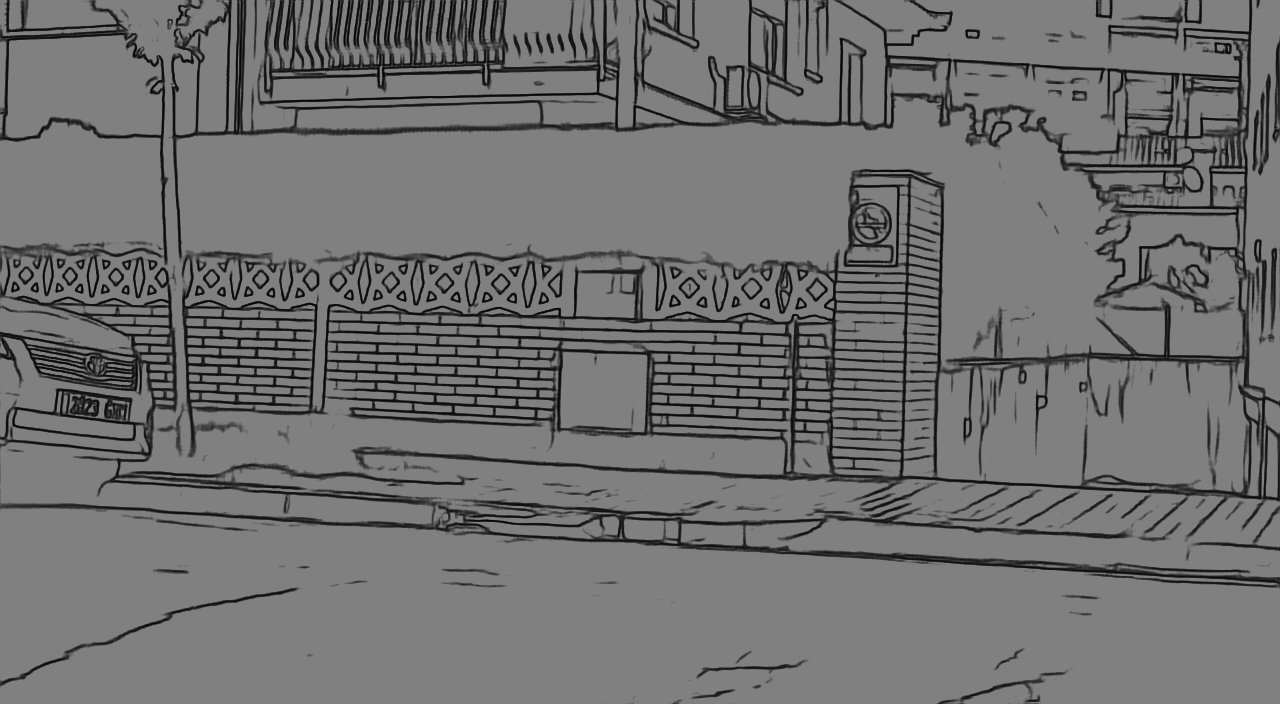} & 
    \includegraphics[width=0.135\linewidth]{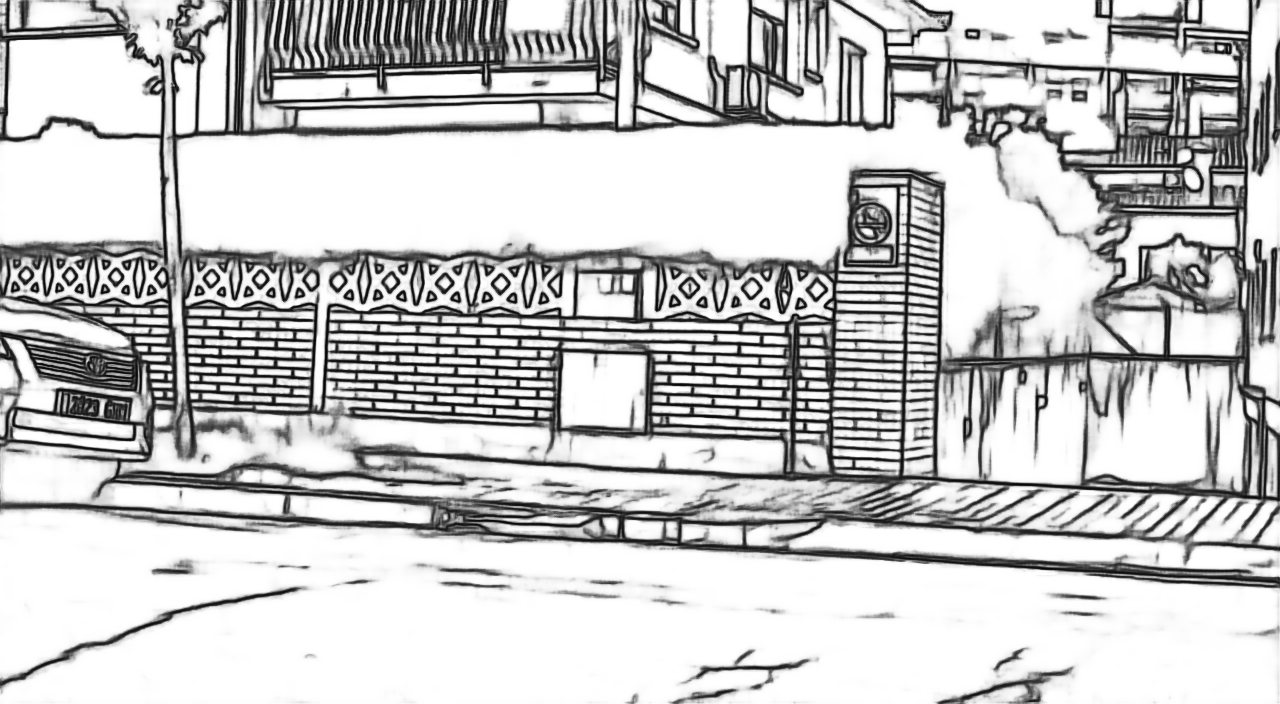}
    \\
\iffalse
    \raisebox{2.5\normalbaselineskip}[0pt][0pt]{\rotatebox[origin=c]{0}{}} &  
    \includegraphics[width=0.135\linewidth]{figs/biped/maps/RGB_192.jpg} &
    \includegraphics[width=0.135\linewidth]{figs/biped/gt/RGB_192.png} & 
    \includegraphics[width=0.135\linewidth]{figs/biped/maps/RGB_192_mts.png} & 
    \includegraphics[width=0.135\linewidth]{figs/biped/maps/RGB_192_side1.png} & 
    \includegraphics[width=0.135\linewidth]{figs/biped/maps/RGB_192_side2.png} &
    \includegraphics[width=0.135\linewidth]{figs/biped/maps/RGB_192_side3.png} &
    \includegraphics[width=0.135\linewidth]{figs/biped/maps/RGB_192.png}
    \\
\fi

    \raisebox{2.5\normalbaselineskip}[0pt][0pt]{\rotatebox[origin=c]{0}{}} &  
    \includegraphics[width=0.135\linewidth]{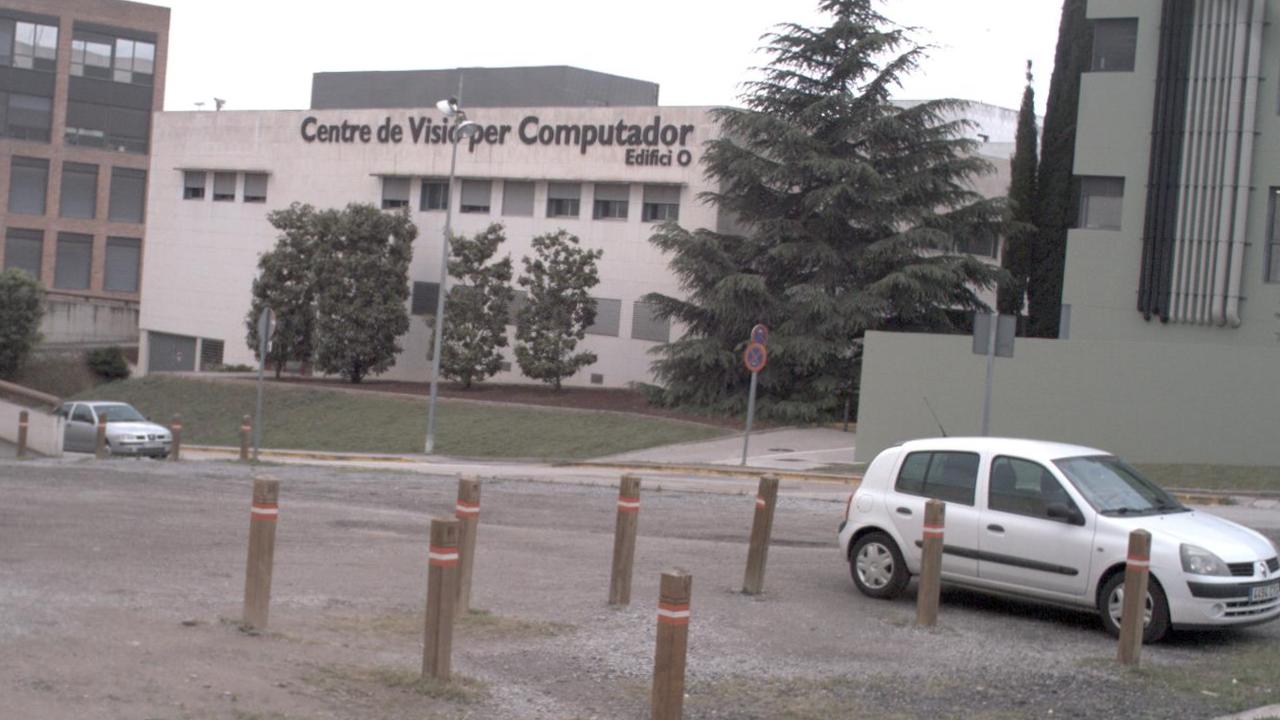} &
    \includegraphics[width=0.135\linewidth]{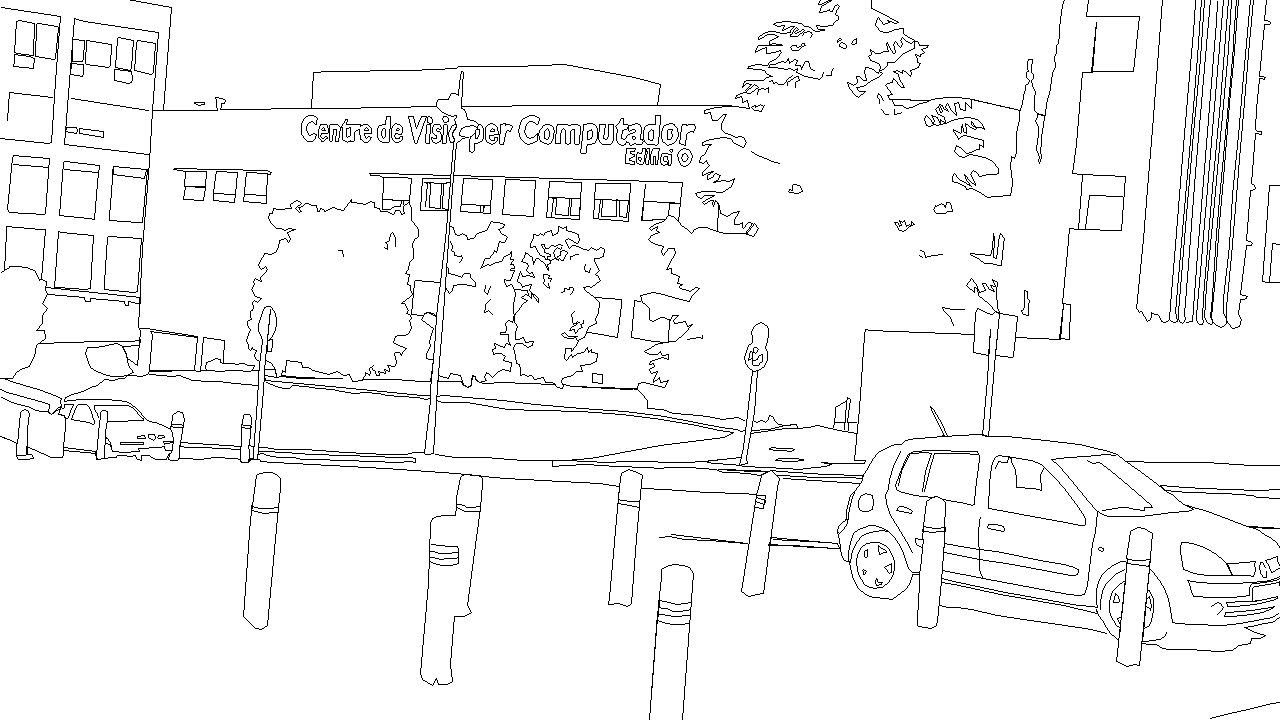} & 
    \includegraphics[width=0.135\linewidth]{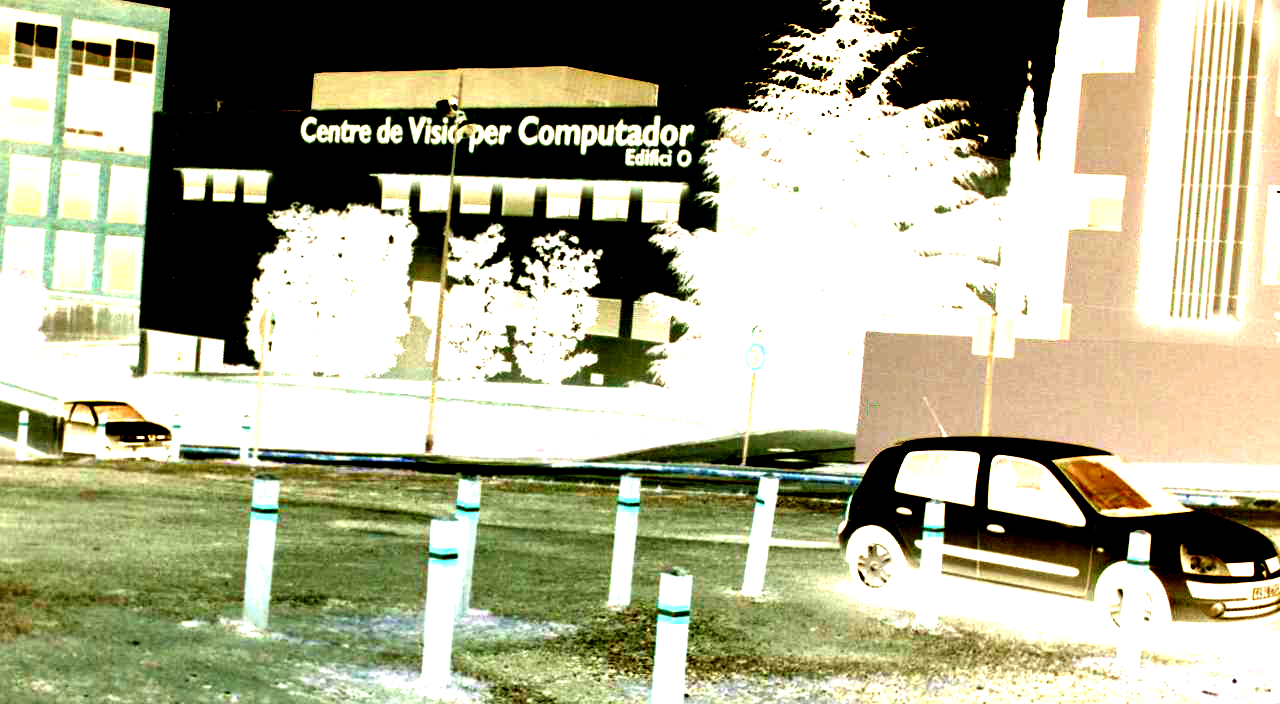} & 
    \includegraphics[width=0.135\linewidth]{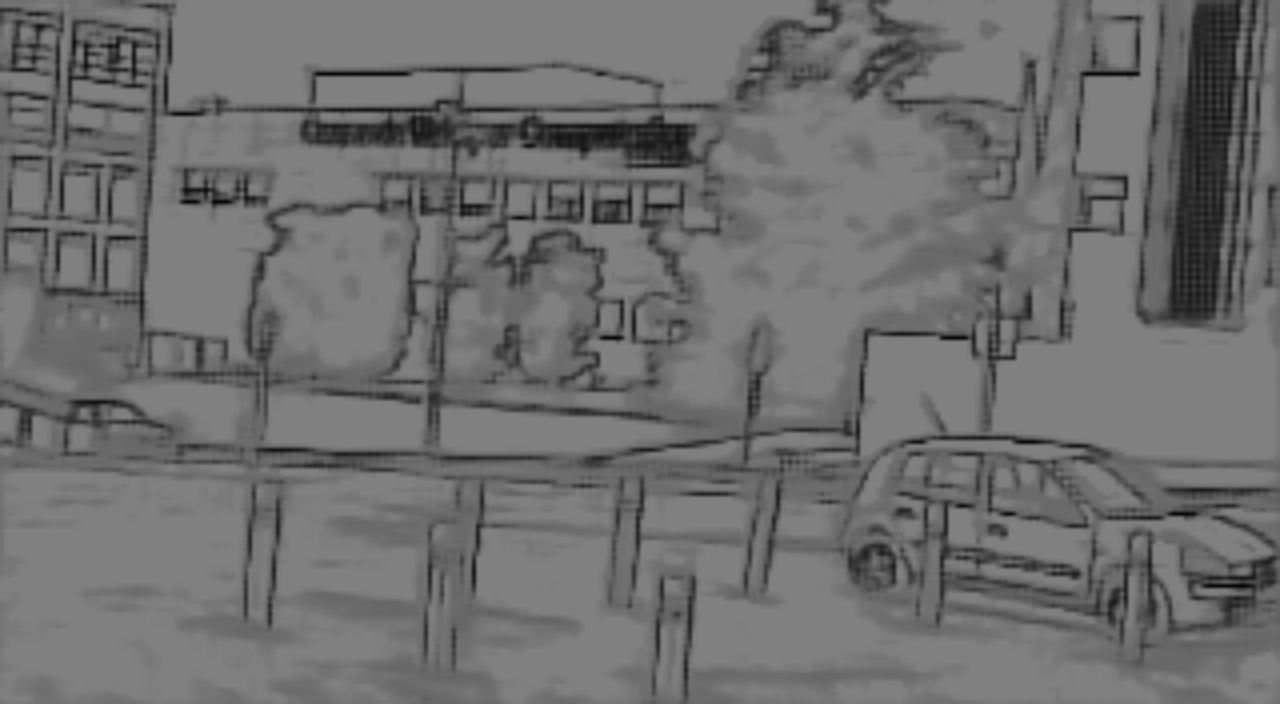} & 
    \includegraphics[width=0.135\linewidth]{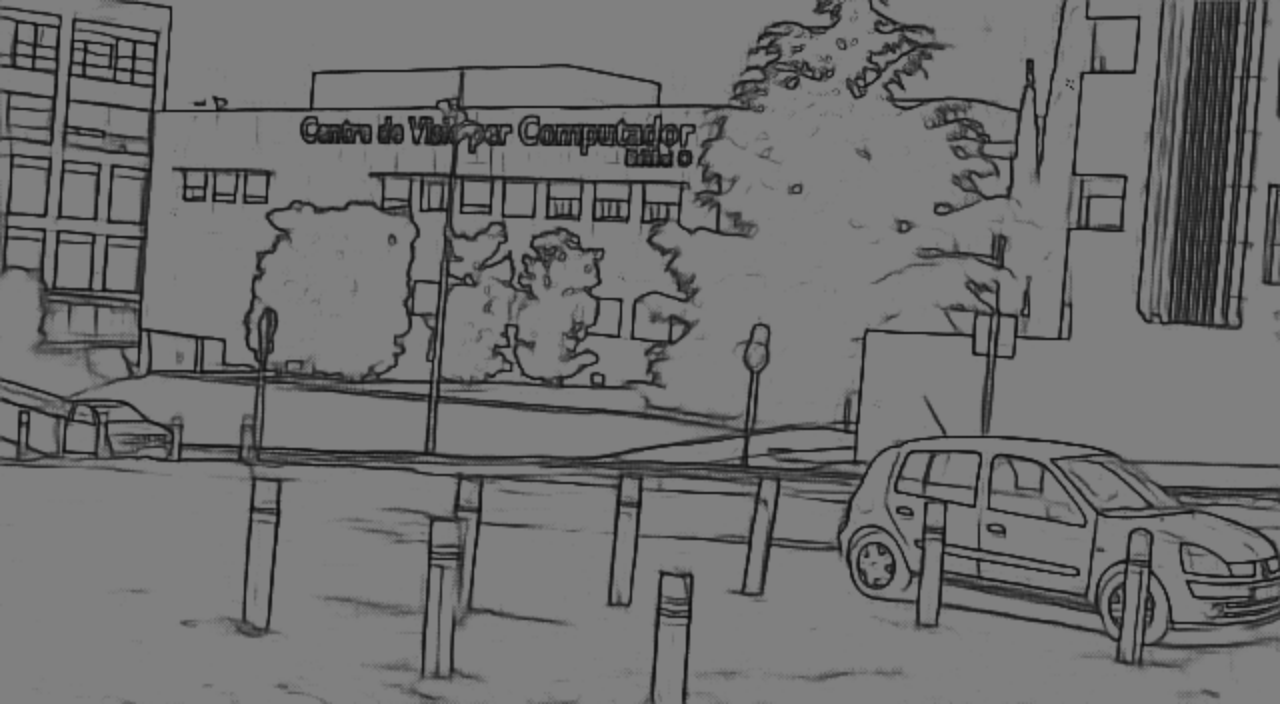} &
    \includegraphics[width=0.135\linewidth]{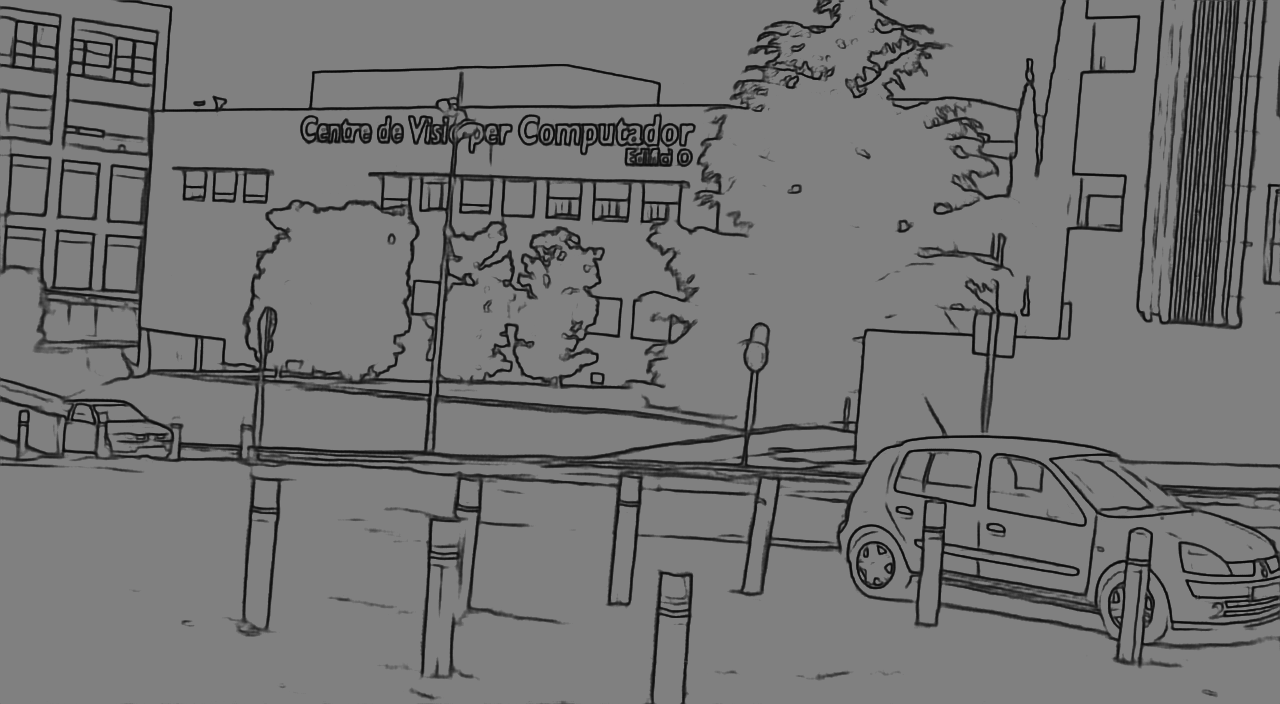} &
    \includegraphics[width=0.135\linewidth]{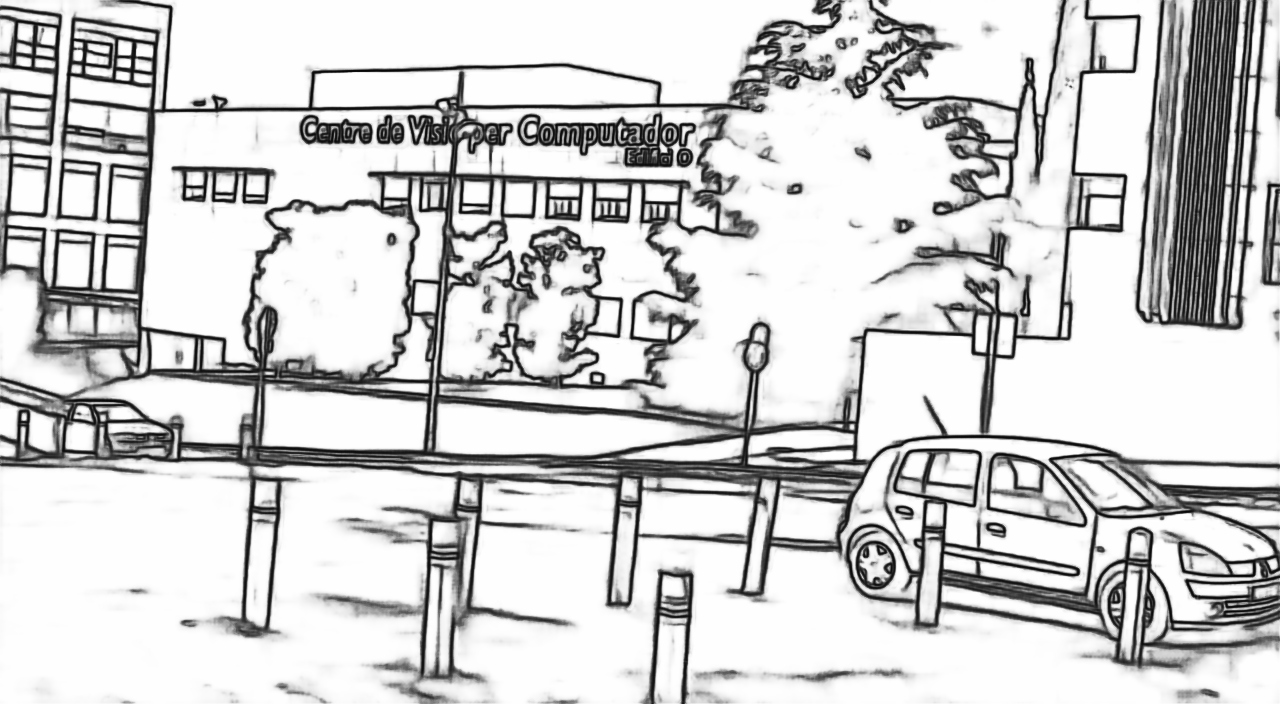}
    \\
    &(a) Original image. &(b) Ground truth. &(c) MTS feature map. 
    & (d) Side 1.& (e) Side 2. & (f) Side 3.& (g) Fused output
\end{tabular}

\caption{ \footnotesize {\textbf{Some examples of feature maps on BIPEDv2 with MTS-DR-1.} }}
\label{fig:feature_biped}
\end{figure*}

%%%%%%%%%%%%%%%%%%%%%%%%%%%%%%%%%%%%%%%%%%%%%%%%%%%%%%%%%%%%%%%%%%%%Result on BSDS
%%%%%%%%%%%%%%%%%%%%%%%%%%%%%%%%%%%%%%%%%%%%%%%%%%%%%%%%%%%%%%%%%%%%Result on BSDS
\begin{table*}
\begin{center}
\begin{tabular}{ l  cc  cc  cc  cc }
\hline
\hline
\multirow{2}{*}[2pt]{Method}
    & \multicolumn{2}{c}{\thead{Thin}}
    & \multicolumn{2}{c}{\thead{Raw}}
        &  \multirow{2}{*}{\thead{mean\\ Precision}}
        &  \multirow{2}{*}{\thead{mean\\ IOU}}
        &  \multirow{2}{*}{\thead{Params}}
        &  \multirow{2}{*}{\thead{GFLOPs}}\\ 
\cline{2-5}
 &ODS&OIS  &ODS&OIS\\
\hline
%RCF \cite{Liu2019RicherDetection} &  & & & & 0.8606&0.6050 &14.804M  & 50.914& \\
TEED \cite{Soria2023TinyGeneralization} & 0.704 &0.711 &  0.614 & 0.614 & 0.8294 & 0.5378  & \textcolor{blue}{58.91K}  & \textcolor{blue}{0.955} \\
Pidinet \cite{Sun2021PixelDetection}& 0.691 &0.693 &  0.612  & 0.613&0.7926&0.5848 & 710.149K &10.56  \\
DexiNed \cite{Soria2020DenseDetection}&0.707  &0.710   &0.636 &0.642 & 0.8073 & 0.5913&35.215M & 66.943  \\
XYW-Net \cite{pang2024bio}&0.720 &0.737  &0.639 & 0.648  & 0.8202&0.5861&808.831K  &11.042 \\
\cline{1-9}
\cline{1-9}
\thead{MTS-DR-1 \\($M=2, C=32$)}&0.734 &\textcolor{red}{0.750}&0.677 &\textcolor{red}{0.683} &0.8404 &  \textcolor{red}{0.6036} &1.599M &15.112  \\

\thead{MTS-DR-2 \\($M=4, C=32$)} &\textcolor{blue}{0.744} &\textcolor{blue}{0.758}&\textcolor{blue}{0.692} &\textcolor{blue}{0.695}& \textcolor{blue}{0.8524} &\textcolor{blue}{0.6079} & 1.717M& 19.735 \\

\thead{MTS-DR-3 \\($M=2, C=16$)}  &\textcolor{red}{ 0.735} &\textcolor{red}{0.750} &0.674 &0.682 & 0.8416&0.6023 & \textcolor{red}{478.419K} &\textcolor{red}{4.758} \\

\thead{MTS-DR-4 \\ ($M=4, C=16$)}  &\textcolor{red}{0.735} &\textcolor{red}{0.750} &\textcolor{red}{0.686}  &\textcolor{blue}{0.695} & \textcolor{red}{0.8428}&0.6004&610.579K & 12.027\\
\hline
\hline
\end{tabular}
\end{center}
\caption{ Comparison with competed methods on BSDS500 dataset.}\label{tab:bsds500}
\end{table*}

%%%%%%%%%%%%%%%%%%%%%%%%%%%%%%%%%%%%%%%%RESULTS ON BIPED
\begin{table*}
\begin{center}
\begin{tabular}{ l  cc  cc  cc  cc }
\hline
\hline
\multirow{2}{*}{Method}
    & \multicolumn{2}{c}{\thead{Thin}}
    & \multicolumn{2}{c}{\thead{Raw}}
        &  \multirow{2}{*}{\thead{mean\\ Precision}}
        &  \multirow{2}{*}{\thead{mean\\ IOU}}
        &  \multirow{2}{*}{\thead{Params}}
        &  \multirow{2}{*}{\thead{GFLOPs}} \\ 
\cline{2-5}
 &ODS&OIS &ODS&OIS\\      
\hline
%RCF \cite{Liu2019RicherDetection} & & & & &0.8590&0.5751& & 115.22& \\
TEED \cite{Soria2023TinyGeneralization} & 0.849&0.853& 0.480&0.480 &0.8124&0.5384&\textcolor{blue}{58.91K}& \textcolor{blue}{2.144} \\
Pidinet \cite{Sun2021PixelDetection}& 0.889&0.896 & 0.773&0.786&0.7926&0.5848&710.149K & 23.76   \\
DexiNed \cite{Soria2020DenseDetection}&0.893  &0.899    &  0.810 &0.820 & 0.8600  & 0.5812&35.215M & 148.081 \\
XYW-Net \cite{pang2024bio}&\textcolor{red}{0.899} &\textcolor{blue}{0.909}& 0.730&0.800 &0.8202&0.5861 & 808.831K &24.845\\
\cline{1-9}
\cline{1-9}
\thead{MTS-DR-1 \\($M=2, C=32$)}  &0.897 &\textcolor{red}{0.905} &\textcolor{red}{0.811} &\textcolor{red}{0.829} &\textcolor{red}{0.8689}& \textcolor{blue}{0.6038}&1.599M & 33.999  \\
\thead{MTS-DR-2\\ ($M=4, C=32$)}  &\textcolor{blue}{0.903}& \textcolor{blue}{0.909} &\textcolor{blue}{0.824}& \textcolor{blue}{0.838}& \textcolor{blue}{0.8698}& \textcolor{red}{0.6022} &1.717M & 44.403 \\
\thead{MTS-DR-3 \\ ($M=2, C=16$)}  &0.894 &0.901 & 0.810& 0.822 &0.8632&0.5973 &\textcolor{red}{478.419K}& \textcolor{red}{10.706}   \\
\thead{MTS-DR-4\\ ($M=4, C=16$)}  &0.883 &0.898 &0.797 &0.826 &0.8626 &0.5957 & 610.579K &27.060 \\
\hline
\hline
\end{tabular}
\end{center}
\caption{ Comparison with competed methods on BIPEDv2 dataset.}\label{tab:biped}
\end{table*}

\subsection{Datasets}
\textbf{BSDS500 \cite{Arbelaes2011ContourSegmentaion}:} initially consists of 200 training images, 100 validation images, and 200 test images of size $321 \times 481$. This dataset is annotated by 4 to 9 participants. We follow the previous publications by taking the average values of the multiple annotations as the ground truth. In this work, 300 images from the training and validation datasets are cropped and augmented for training. Original images are cropped with overlap to $256\times256$ and then flipped and rotated. Eventually, 11600 images are generated for training, while 200 images are used for testing.

\textbf{BIPEDv2 \cite{Soria2020DenseDetection}:} contains 250 outdoor images with annotations for edge detection tasks specifically. Each initial image has 1280 $\times$ 720 pixels, 200 images for training, and 50 for testing. In this work, we randomly cropped 20 patches with 384 $\times$ 384 pixels of each image of the train set. In addition, each cropped image is rotated by $90^\circ$, $180^\circ$ and $270^\circ$. Hence, the total number of images for training is 16000. This dataset is deliberately annotated pixel by pixel for edge detection. 

\subsection{Implementation Details}
We carry out our experiments with Pytorch on the NVIDIA GPU cluster platform. The proposed models are trained for 30 epochs on BIPEDv2 and 20 epochs on BSDS500 with a batch size of 4. The models are trained with the Adam optimizer \cite{KingmaB14}. The learning rate is initially set to $0.005$ and decays at epoch 15 with an exponential schedule. In this work, the final results are from a single scale fused feature map and non-pre-training involved. The hyperparameter value of $\lambda$ is set to 1.1 for both datasets, whereas $\eta$ is 0.3 for the BSDS500 dataset. 

The number of MTS-DR Blocks in the Backbone is $M$, and the number of channels in MTS-DR-Net is $C$. We adopt different $M$ and $C$ to perform ablation studies. The backbone contains three additional hyperparameters: $T$ is the number of mode products in the matrix factorization, the compressing ratio (CR) presents the percentage of information preserved, and the window scales (WS) in the MTS layer. The hyperparameters of the MTS-DR-1 baseline are set as: $M=2$, $C=32$, $CR=0.4$, $WS = [8, 16, 32, 64]$, and $T=3$.

\begin{figure}
\centering
\footnotesize
\renewcommand{\tabcolsep}{1pt} % adjust horizontal space
\renewcommand{\arraystretch}{0.2} % adjust vertical space
\begin{tabular}{ccccccc}
    \raisebox{0.8\normalbaselineskip}[0pt][0pt]{\rotatebox[origin=c]{0}{(a)}} &  
    \includegraphics[width=0.15\linewidth]{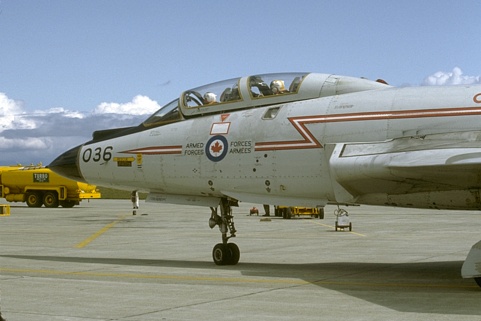} &
    \includegraphics[width=0.15\linewidth]{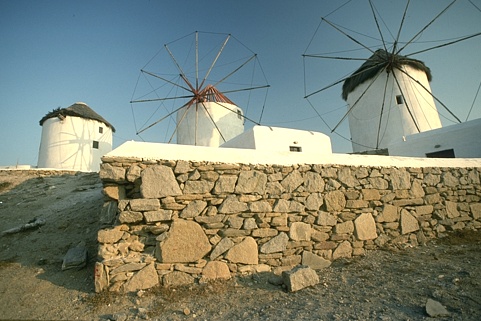} & 
    \includegraphics[width=0.15\linewidth]{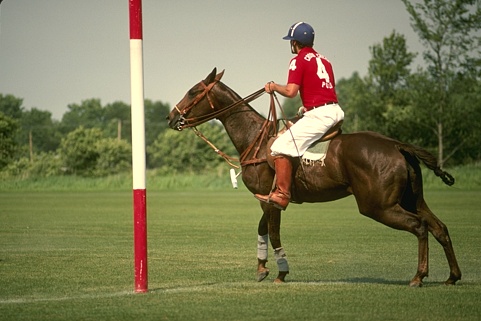} & 
    \includegraphics[width=0.15\linewidth]{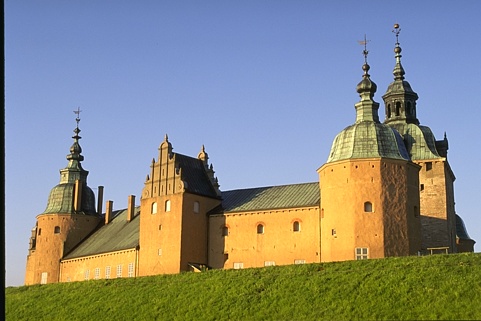} &
    \includegraphics[width=0.15\linewidth]{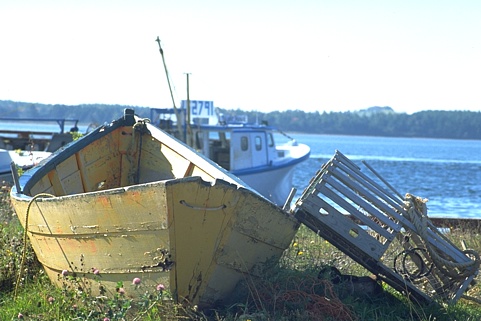} &
    \includegraphics[width=0.15\linewidth]{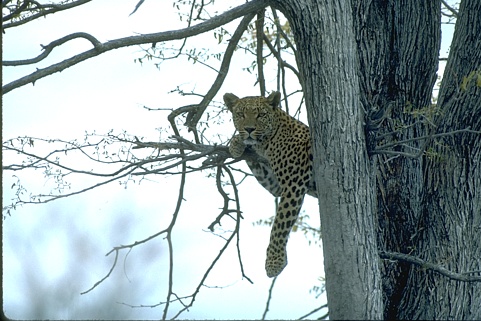} 
    \\

     \raisebox{0.8\normalbaselineskip}[0pt][0pt]{\rotatebox[origin=c]{0}{(b)}} &  
    \includegraphics[width=0.15\linewidth]{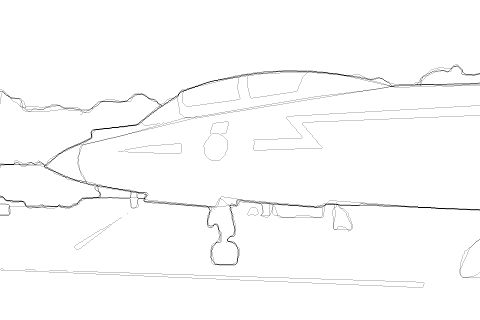} & 
    \includegraphics[width=0.15\linewidth]{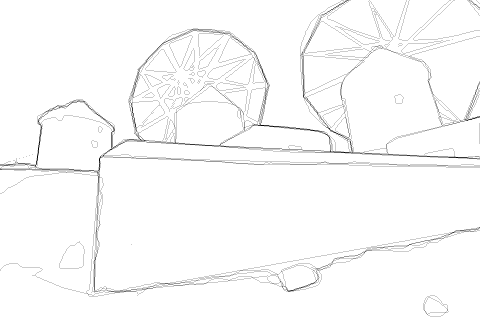} & 
    \includegraphics[width=0.15\linewidth]{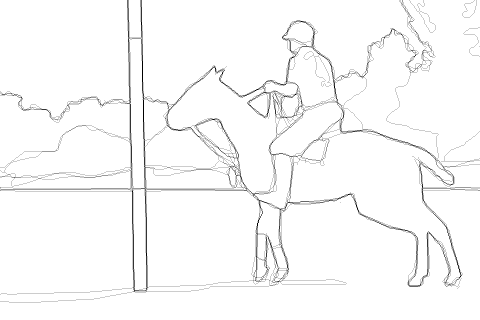} & 
    \includegraphics[width=0.15\linewidth]{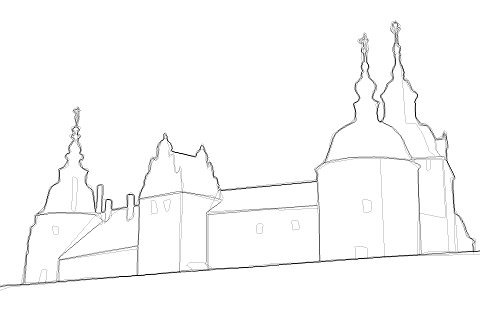} &
    \includegraphics[width=0.15\linewidth]{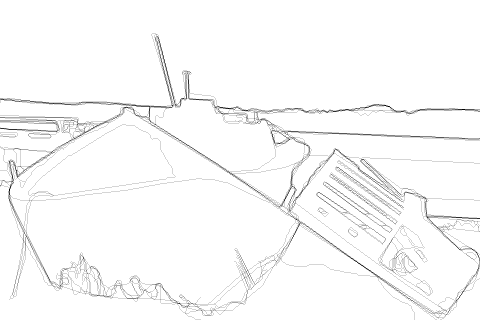} &
    \includegraphics[width=0.15\linewidth]{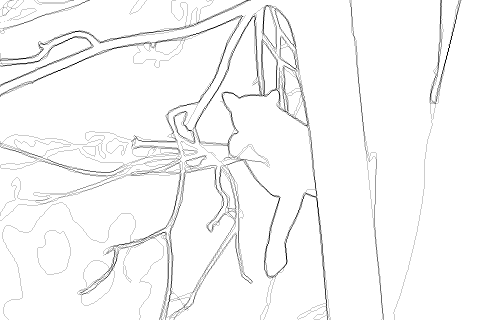} 
    \\

    \raisebox{0.8\normalbaselineskip}[0pt][0pt]{\rotatebox[origin=c]{0}{(c)}} &  
    \includegraphics[width=0.15\linewidth]{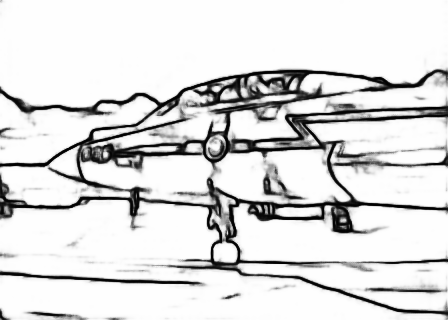} & 
    \includegraphics[width=0.15\linewidth]{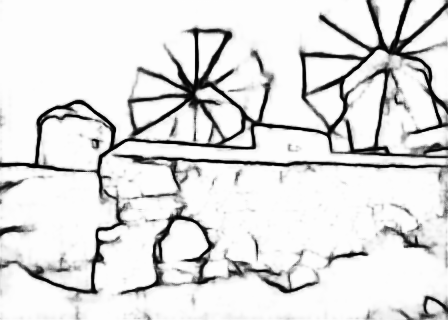} & 
    \includegraphics[width=0.15\linewidth]{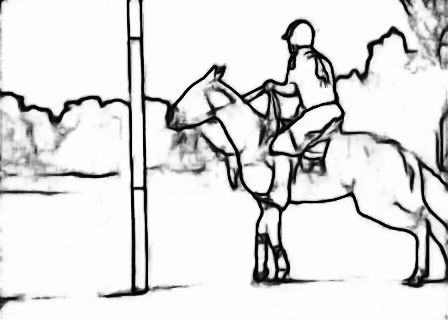} & 
    \includegraphics[width=0.15\linewidth]{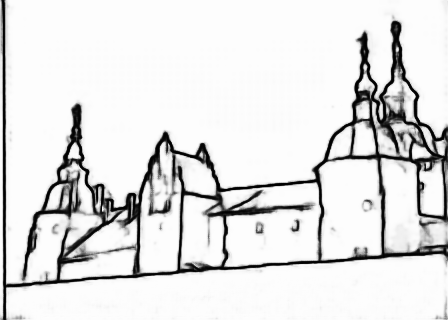} &
    \includegraphics[width=0.15\linewidth]{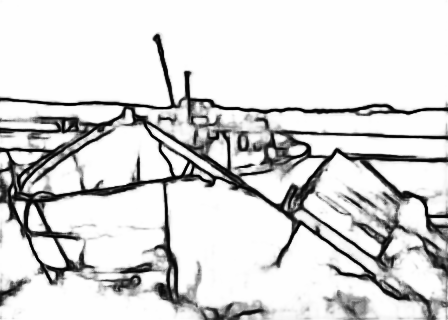} &
    \includegraphics[width=0.15\linewidth]{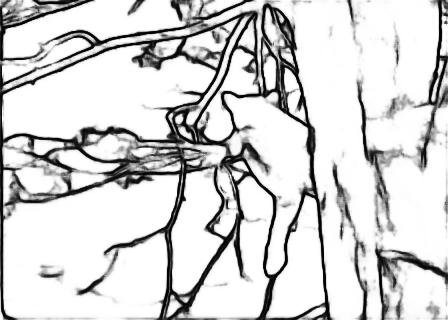}
    \\
    \raisebox{0.8\normalbaselineskip}[0pt][0pt]{\rotatebox[origin=c]{0}{(d)}} &  
    \includegraphics[width=0.15\linewidth]{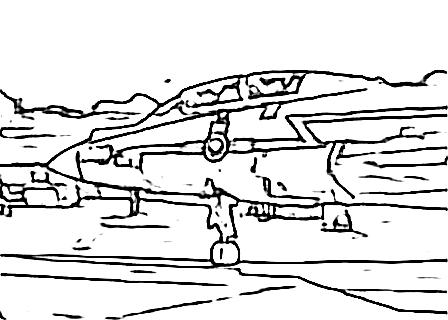} & 
    \includegraphics[width=0.15\linewidth]{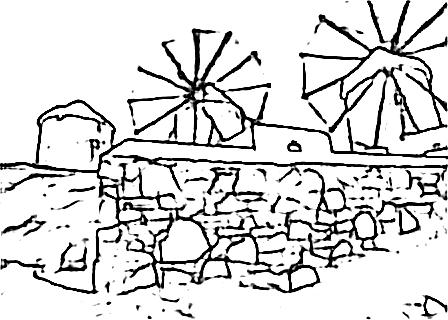} & 
    \includegraphics[width=0.15\linewidth]{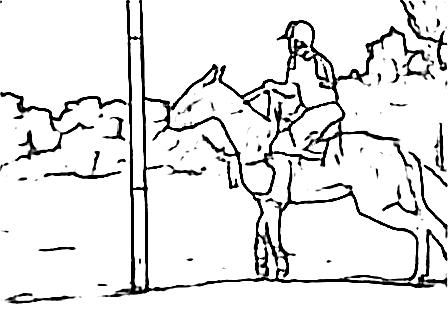} & 
    \includegraphics[width=0.15\linewidth]{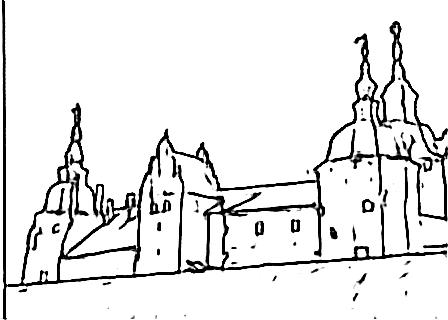} &
    \includegraphics[width=0.15\linewidth]{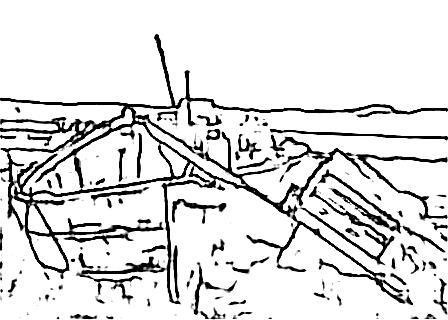} &
    \includegraphics[width=0.15\linewidth]{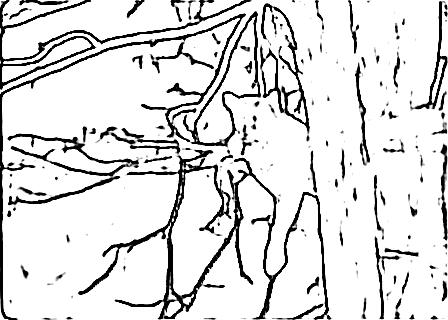} 
    \\

    \raisebox{0.8\normalbaselineskip}[0pt][0pt]{\rotatebox[origin=c]{0}{(e)}} &  
    \includegraphics[width=0.15\linewidth]{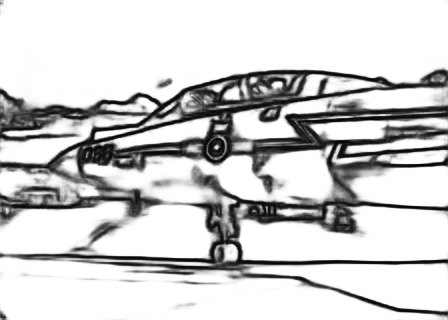} & 
    \includegraphics[width=0.15\linewidth]{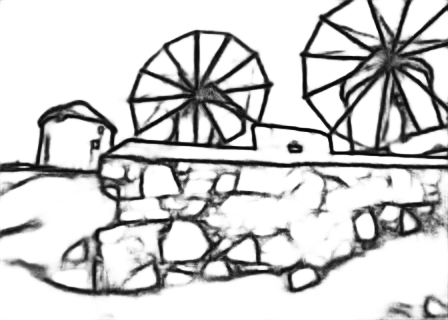} & 
    \includegraphics[width=0.15\linewidth]{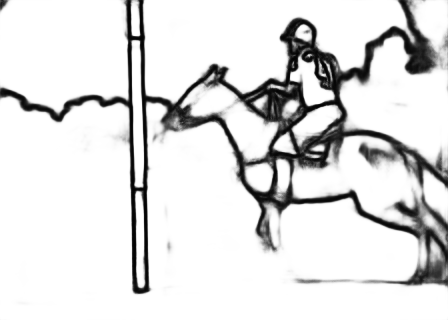} & 
    \includegraphics[width=0.15\linewidth]{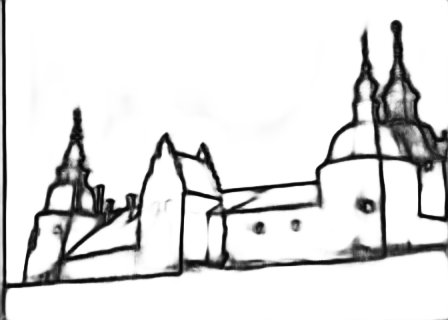} &
    \includegraphics[width=0.15\linewidth]{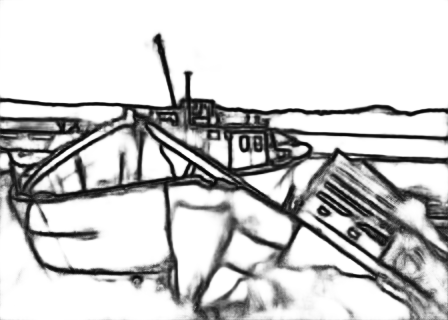} &
    \includegraphics[width=0.15\linewidth]{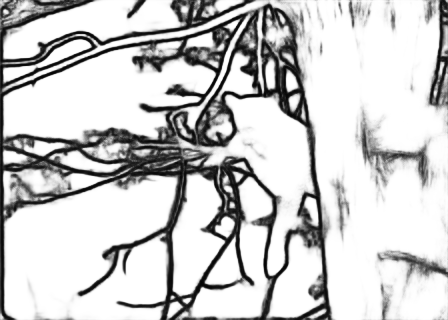} 
    \\

     \raisebox{0.8\normalbaselineskip}[0pt][0pt]{\rotatebox[origin=c]{0}{(f)}} &  
    \includegraphics[width=0.15\linewidth]{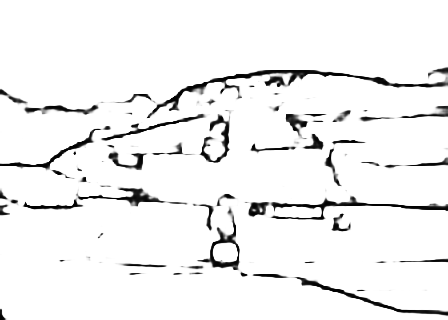} & 
    \includegraphics[width=0.15\linewidth]{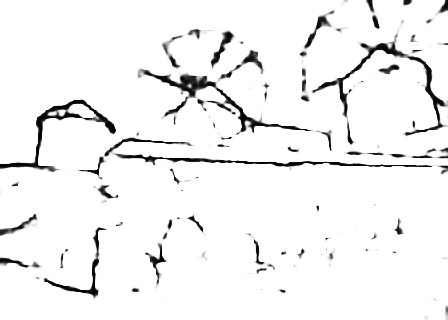} & 
    \includegraphics[width=0.15\linewidth]{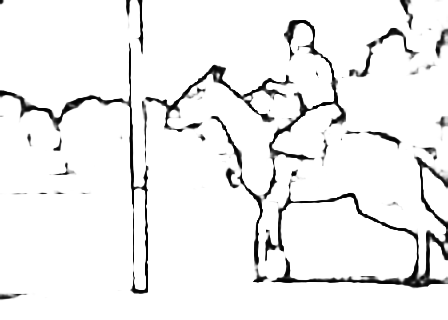} & 
    \includegraphics[width=0.15\linewidth]{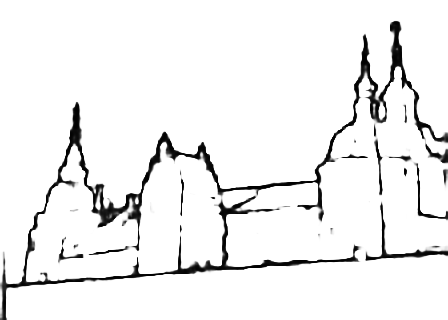} &
     \includegraphics[width=0.15\linewidth]{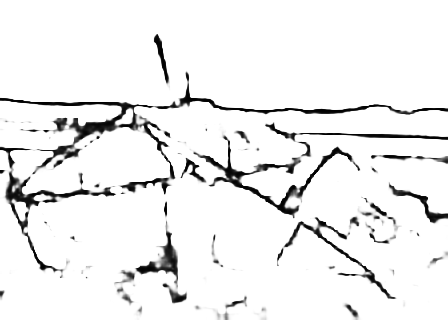} &
    \includegraphics[width=0.15\linewidth]{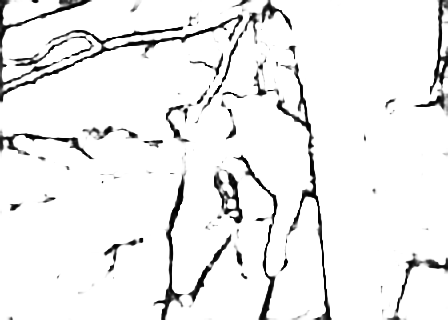}  
    \\

    \raisebox{0.8\normalbaselineskip}[0pt][0pt]{\rotatebox[origin=c]{0}{(g)}} &  
    \includegraphics[width=0.15\linewidth]{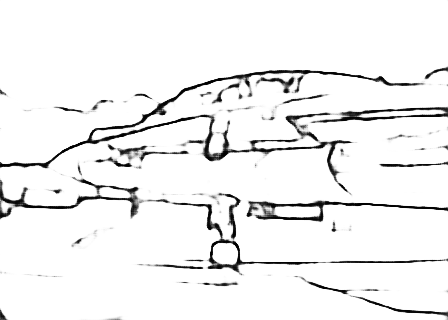} & 
    \includegraphics[width=0.15\linewidth]{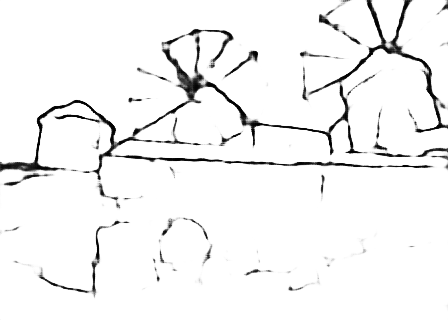} & 
    \includegraphics[width=0.15\linewidth]{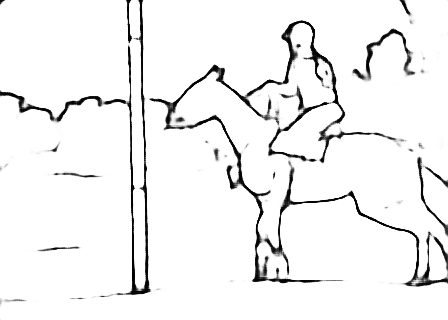} & 
    \includegraphics[width=0.15\linewidth]{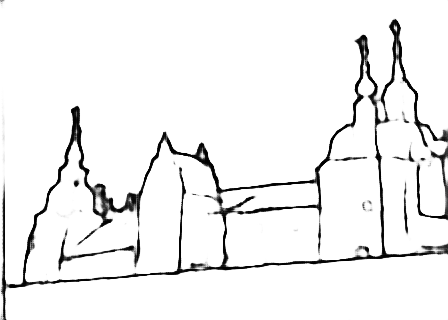} &
     \includegraphics[width=0.15\linewidth]{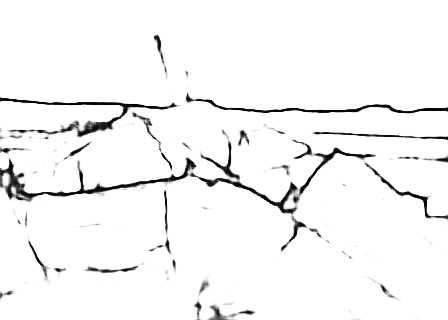} &
    \includegraphics[width=0.15\linewidth]{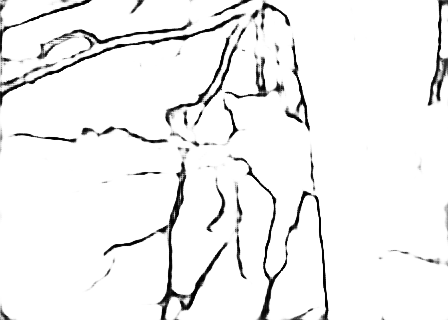} 
    \\

    %\raisebox{2.5\normalbaselineskip}[0pt][0pt]{\rotatebox[origin=c]{0}{(k)}} &  
    %\includegraphics[width=0.12\linewidth]{figs/bsds/rcf/10081_ss.png} & 
    %\includegraphics[width=0.12\linewidth]{figs/bsds/rcf/118031_ss.png} & 
    %\includegraphics[width=0.12\linewidth]{figs/bsds/rcf/206062_ss.png}& 
    %\includegraphics[width=0.12\linewidth]{figs/bsds/rcf/16068_ss.png}& 
    %\includegraphics[width=0.12\linewidth]{figs/bsds/rcf/201080_ss.png}&
    %\includegraphics[width=0.12\linewidth]{figs/bsds/rcf/253016_ss.png}&
    %\includegraphics[width=0.12\linewidth]{figs/bsds/rcf/393035_ss.png} &
    %\includegraphics[width=0.12\linewidth]{figs/bsds/rcf/335094_ss.png}\\
\end{tabular}
\caption{ \footnotesize {\textbf{Visual results on BSDS500.} (a) Input image. (b) Ground truth. (c) MTS-DR-1. (d) TEED \cite{Soria2023TinyGeneralization}. (e) XYW-Net \cite{pang2024bio}. (f) Pidinet \cite{Sun2021PixelDetection}. (g) DexiNed \cite{Soria2020DenseDetection}.}
}
\label{fig:visual_bsds}
\end{figure}
%biped
\begin{figure}
\centering
\footnotesize
\renewcommand{\tabcolsep}{1pt} % adjust horizontal space
\renewcommand{\arraystretch}{0.2} % adjust vertical space
\begin{tabular}{cccccc}
    \raisebox{0.8\normalbaselineskip}[0pt][0pt]{\rotatebox[origin=c]{0}{(a)}} &  
    \includegraphics[width=0.18\linewidth]{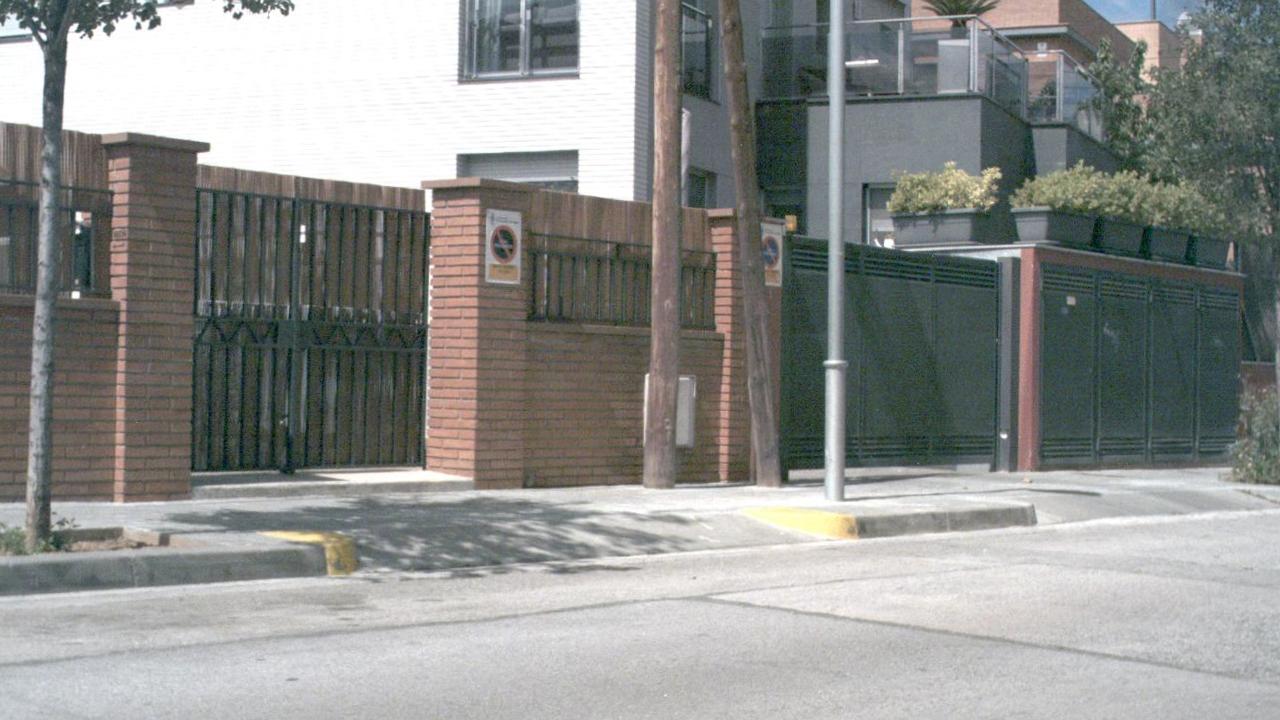} &
    \includegraphics[width=0.18\linewidth]{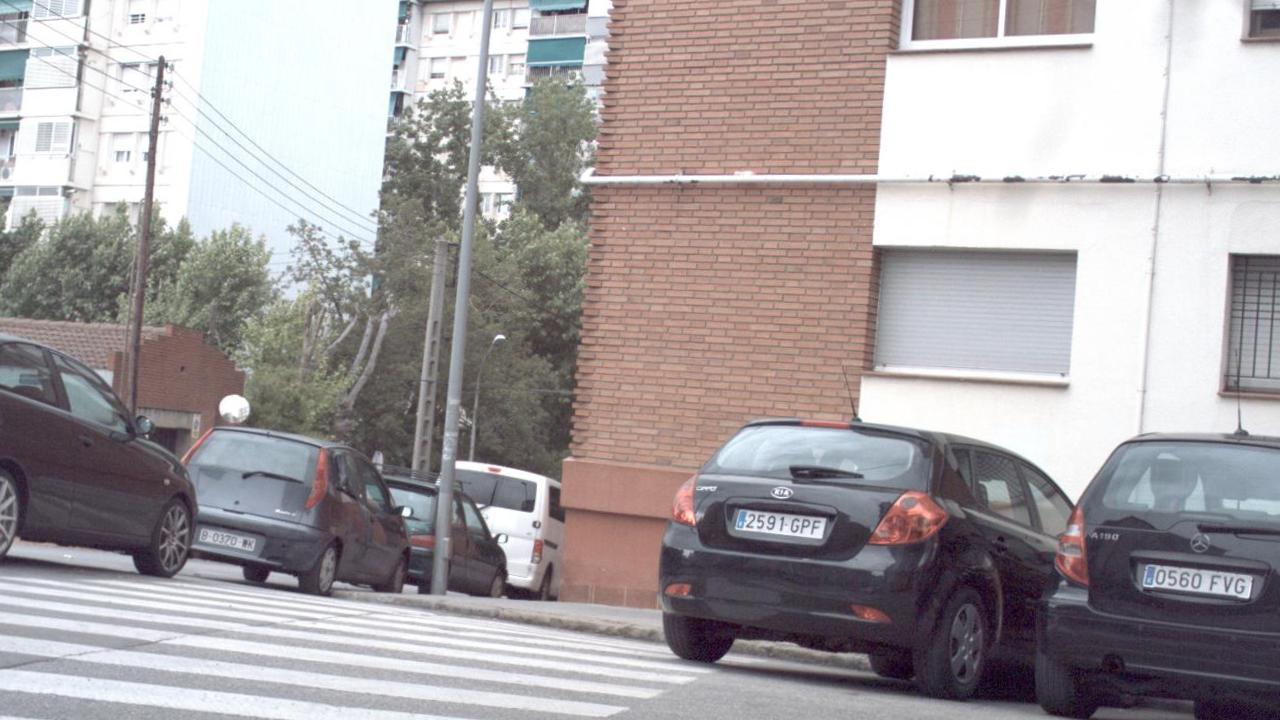} & 
    \includegraphics[width=0.18\linewidth]{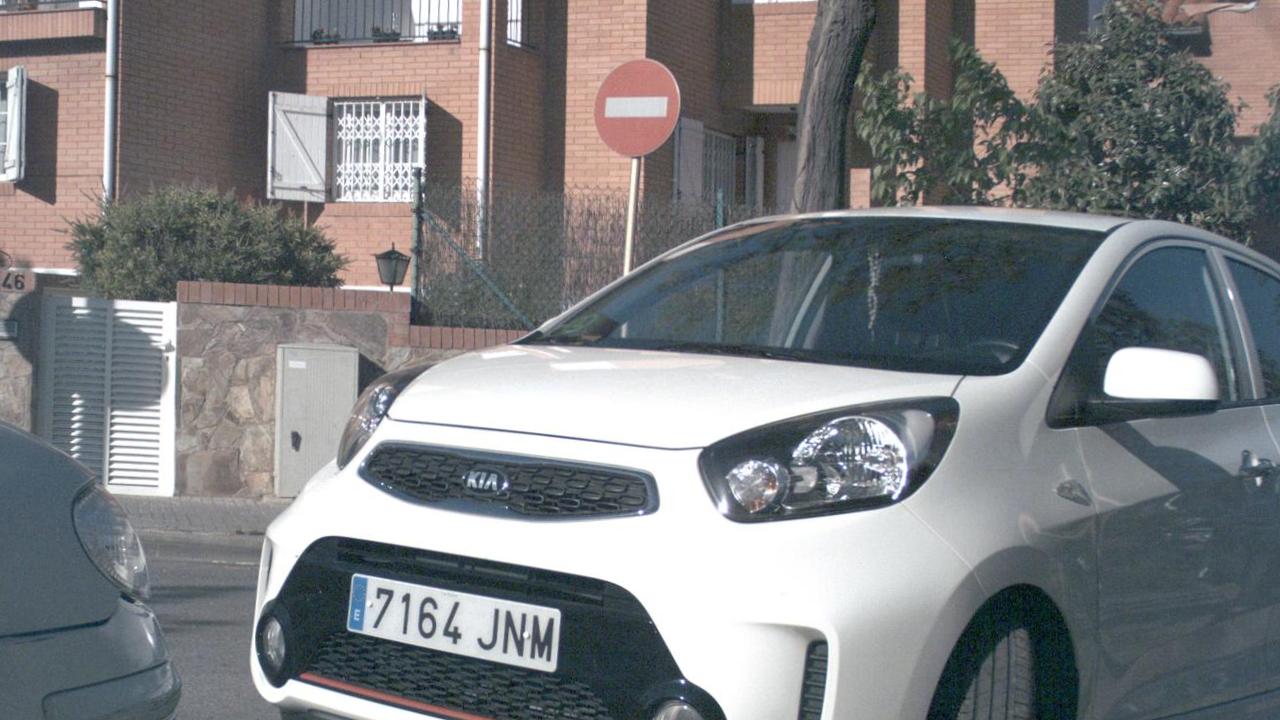} & 
    \includegraphics[width=0.18\linewidth]{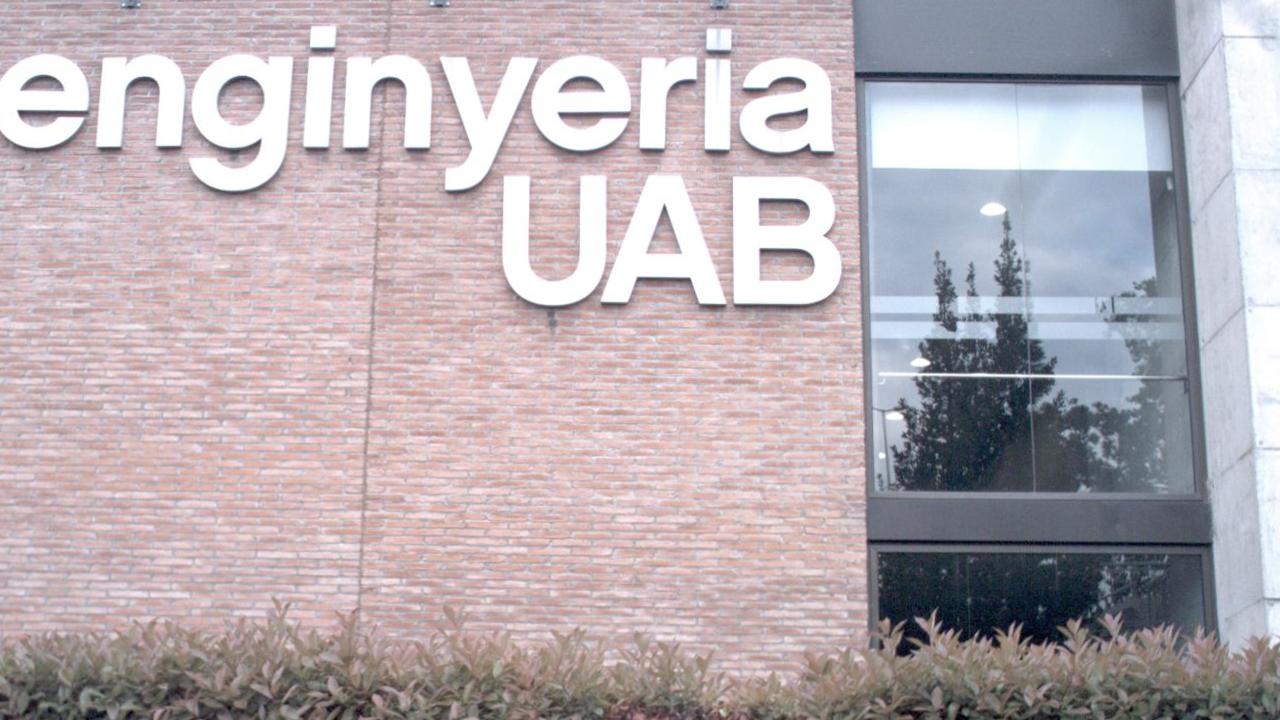} &
    \includegraphics[width=0.18\linewidth]{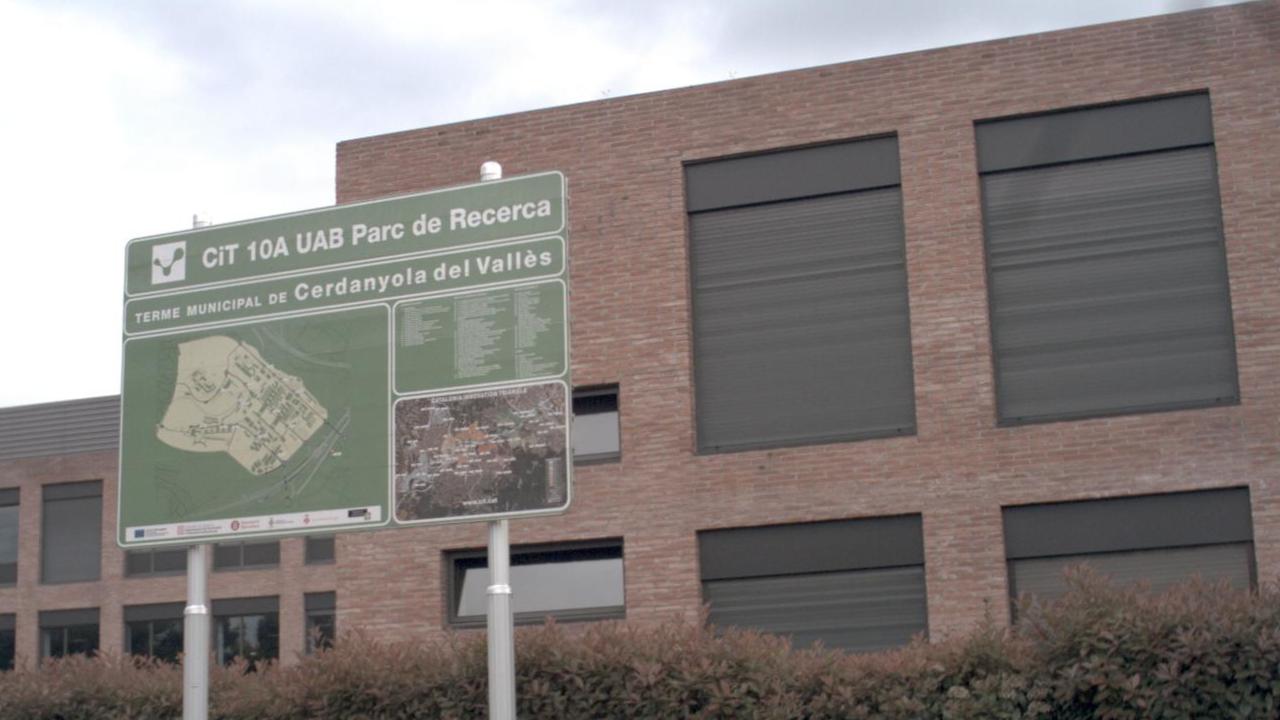}
    \\

    \raisebox{0.8\normalbaselineskip}[0pt][0pt]{\rotatebox[origin=c]{0}{(b)}} &  
    \includegraphics[width=0.18\linewidth]{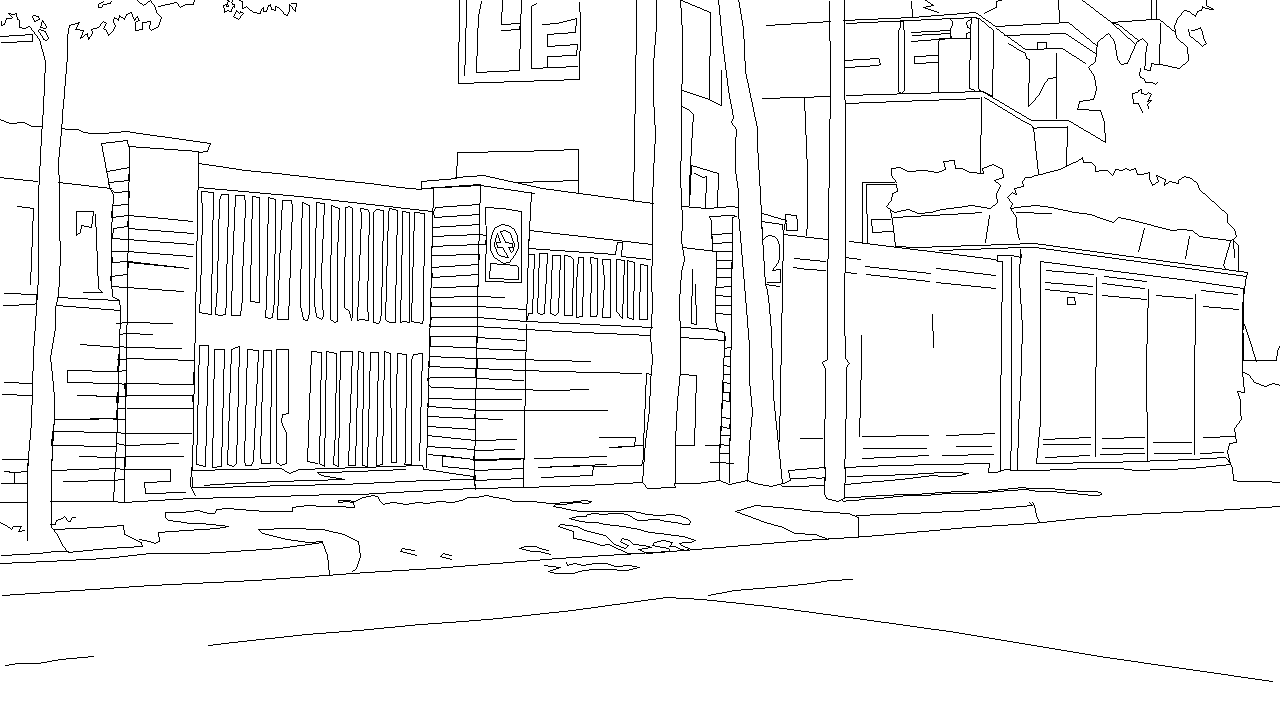} &
    \includegraphics[width=0.18\linewidth]{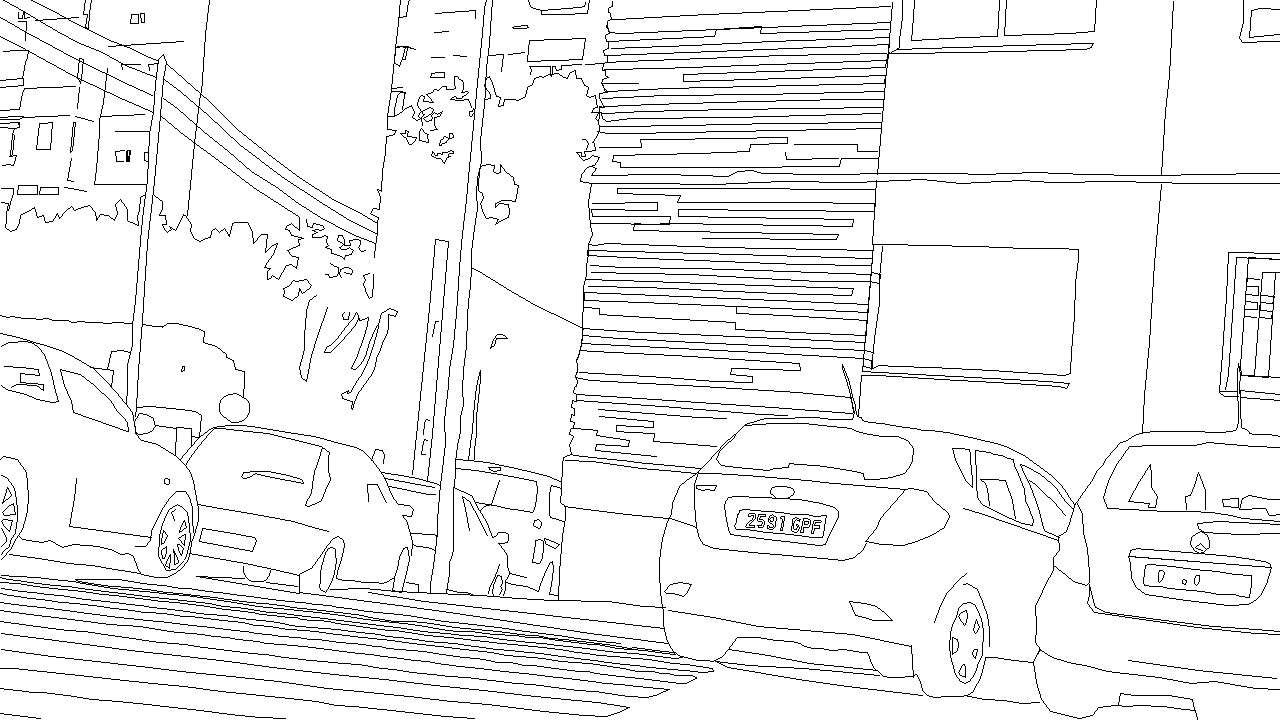} & 
    \includegraphics[width=0.18\linewidth]{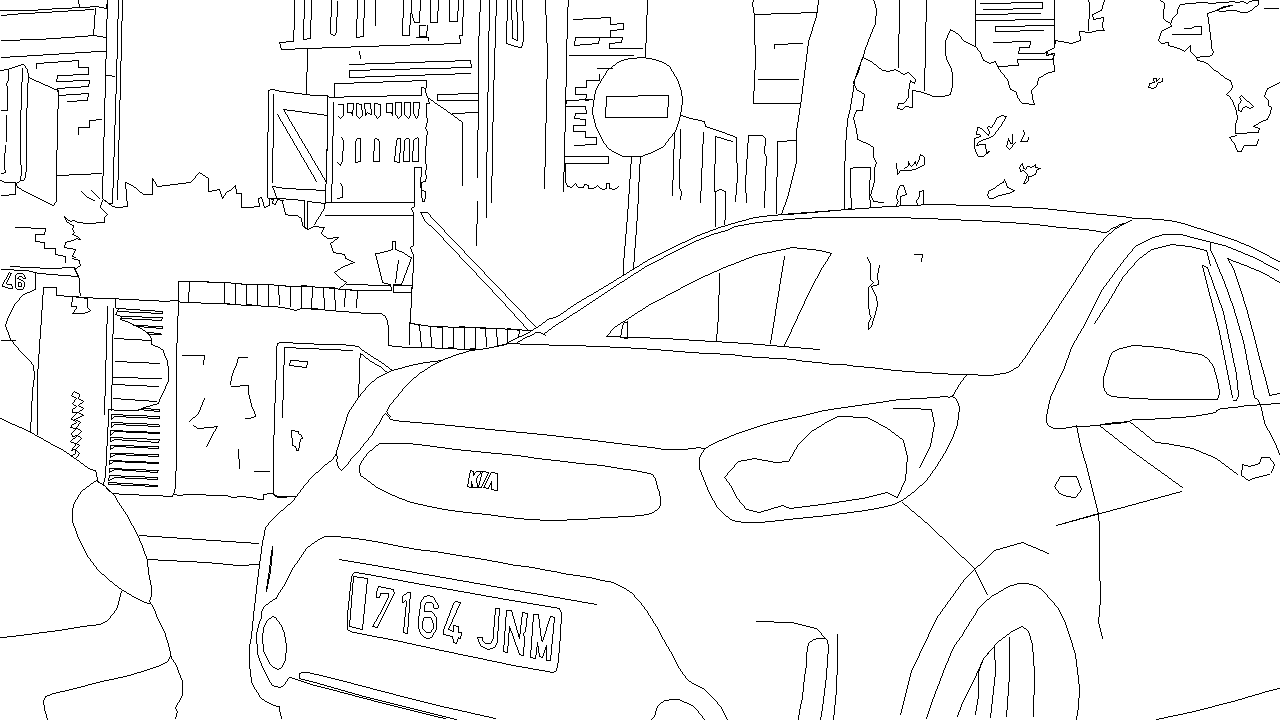} & 
    \includegraphics[width=0.18\linewidth]{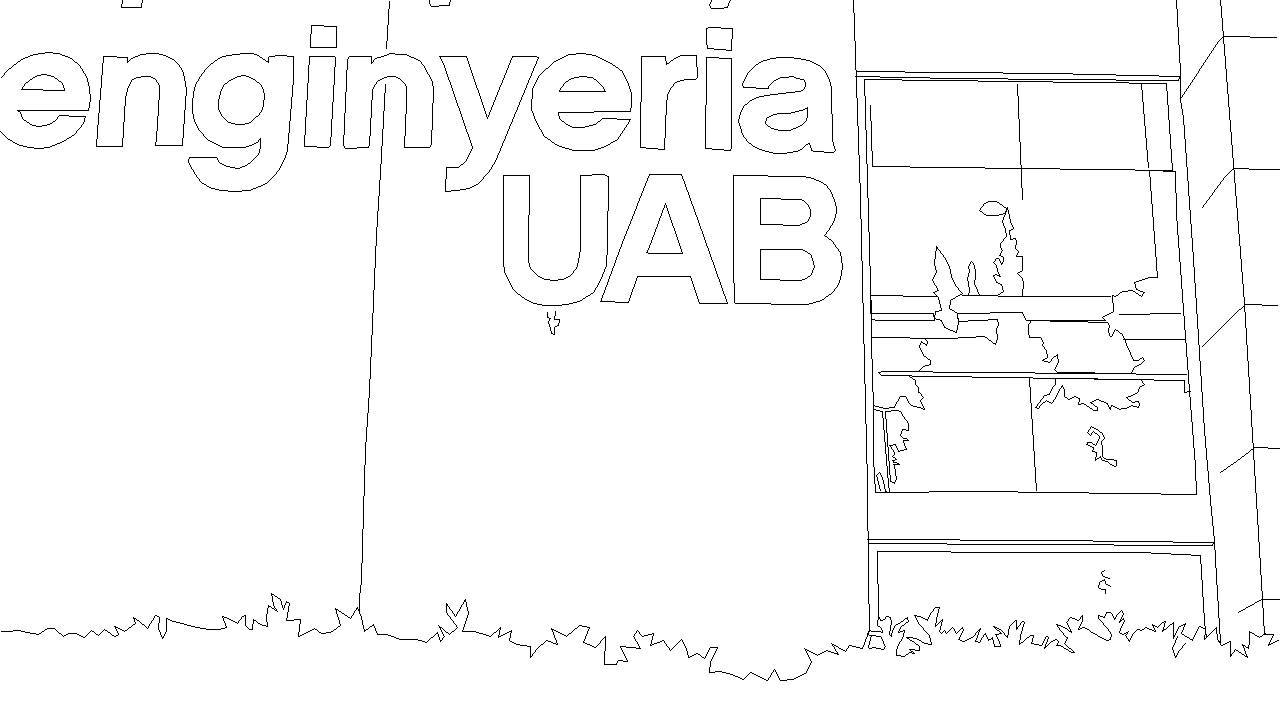} &
     \includegraphics[width=0.18\linewidth]{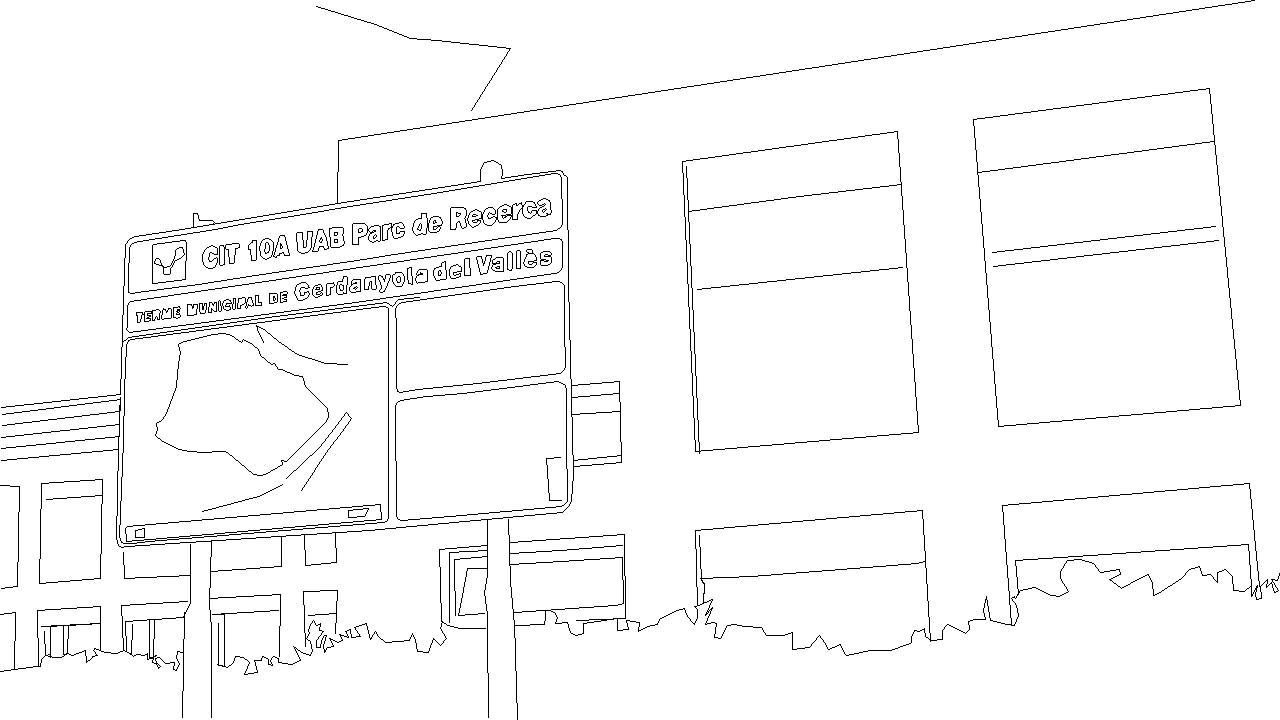}
    \\

    \raisebox{0.8\normalbaselineskip}[0pt][0pt]{\rotatebox[origin=c]{0}{(c)}} &  
    \includegraphics[width=0.18\linewidth]{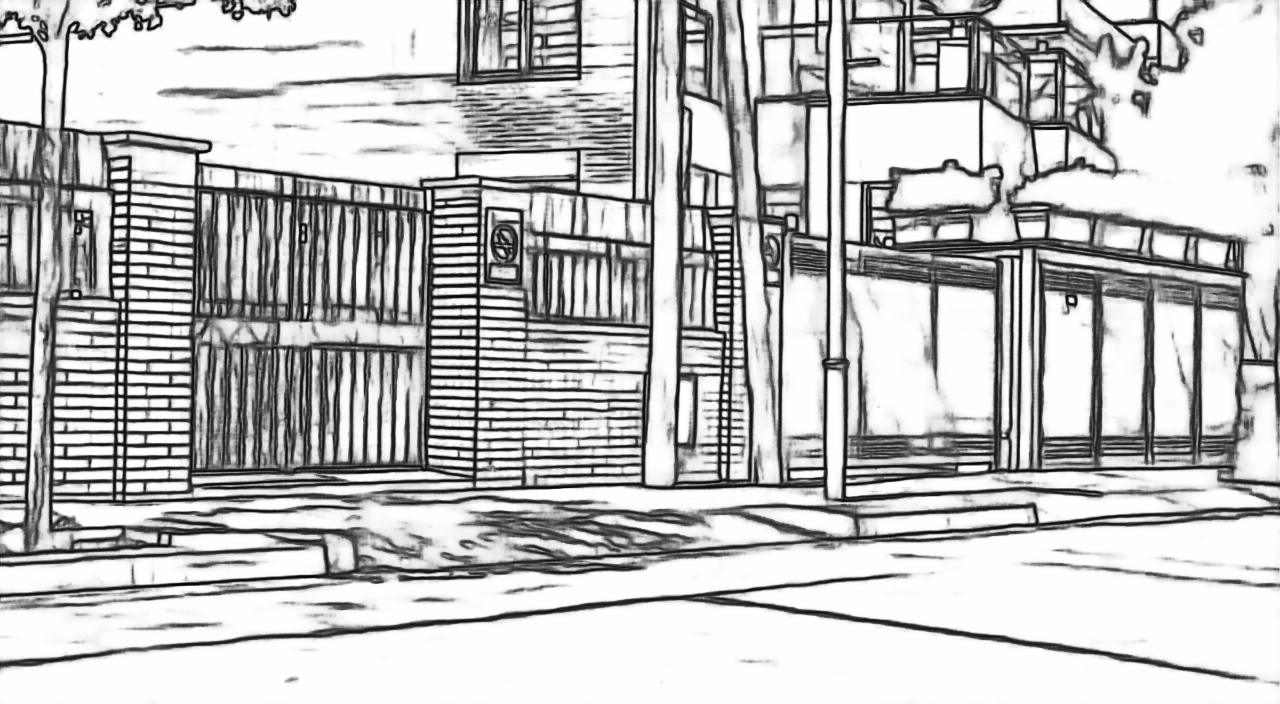} &
    \includegraphics[width=0.18\linewidth]{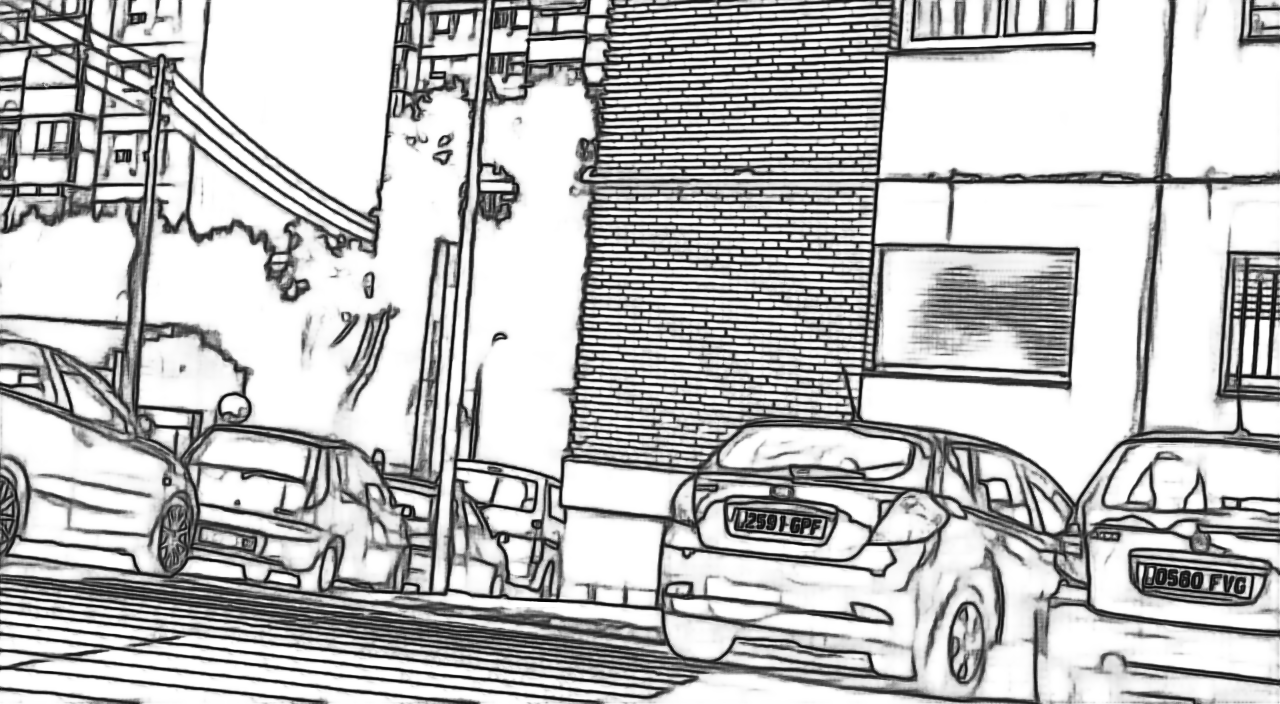} & 
    \includegraphics[width=0.18\linewidth]{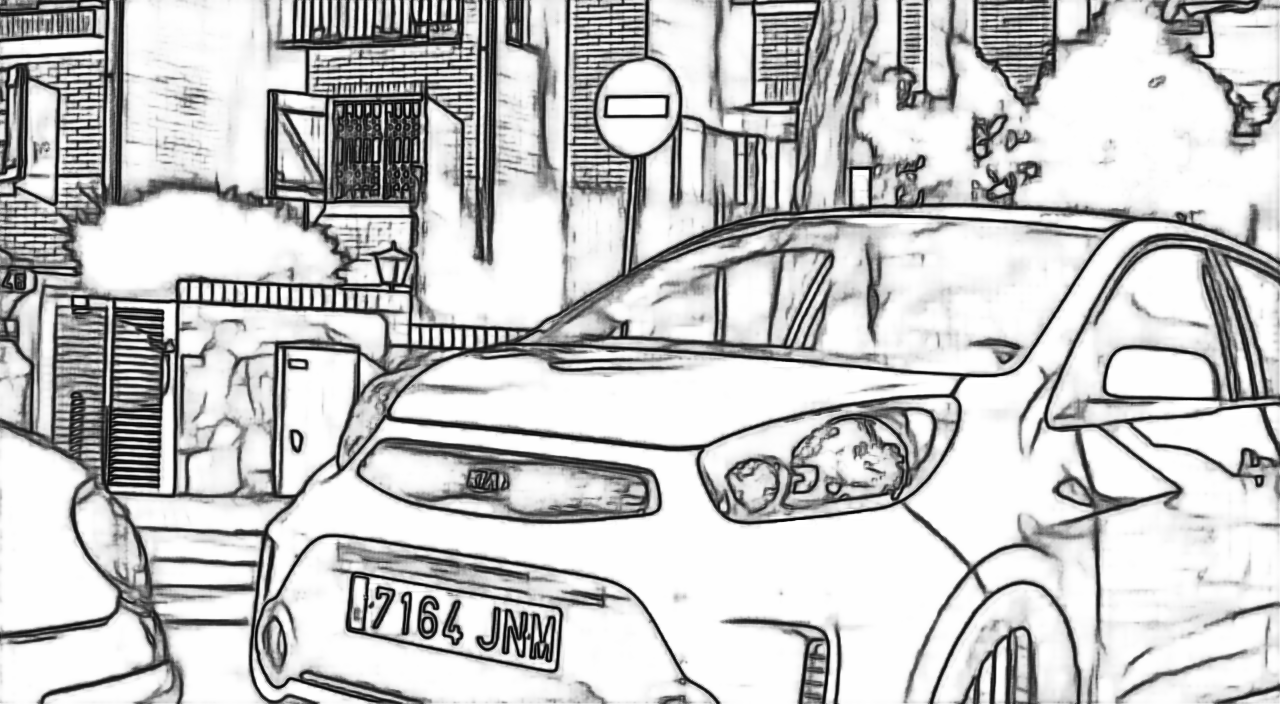} & 
    \includegraphics[width=0.18\linewidth]{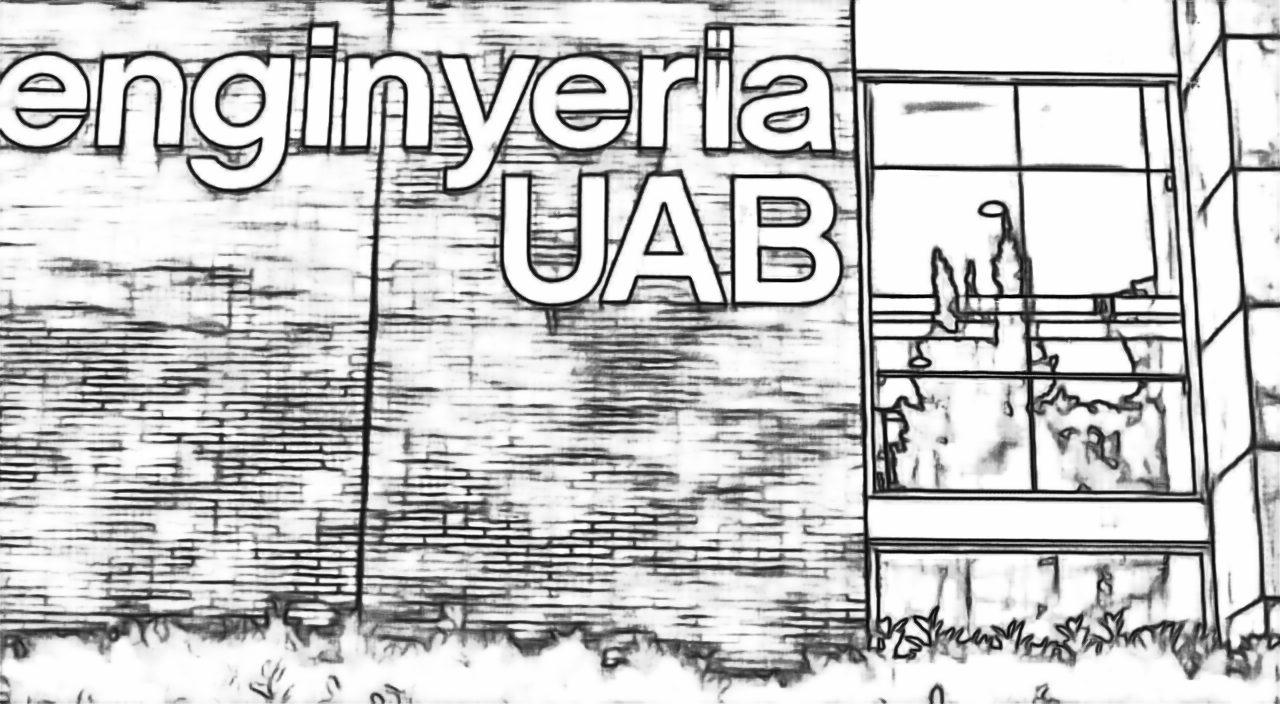} &
    \includegraphics[width=0.18\linewidth]{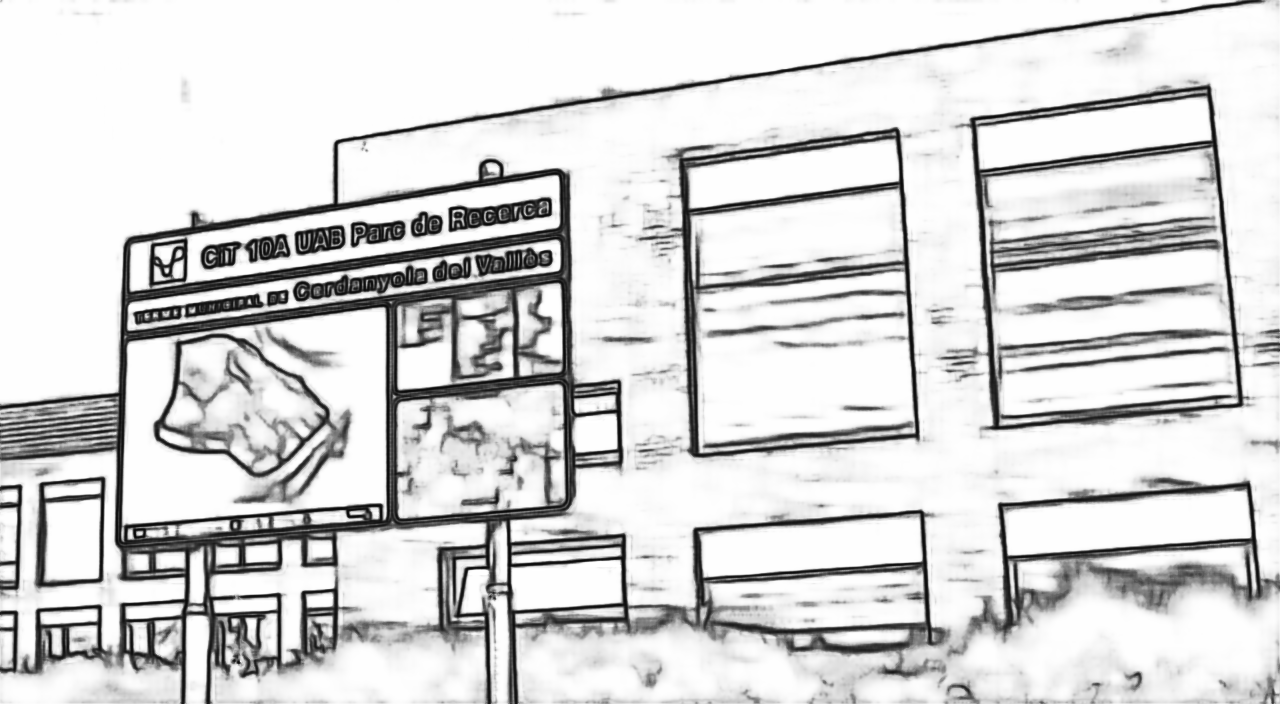}
    \\

     \raisebox{0.8\normalbaselineskip}[0pt][0pt]{\rotatebox[origin=c]{0}{(d)}} &  
    \includegraphics[width=0.18\linewidth]{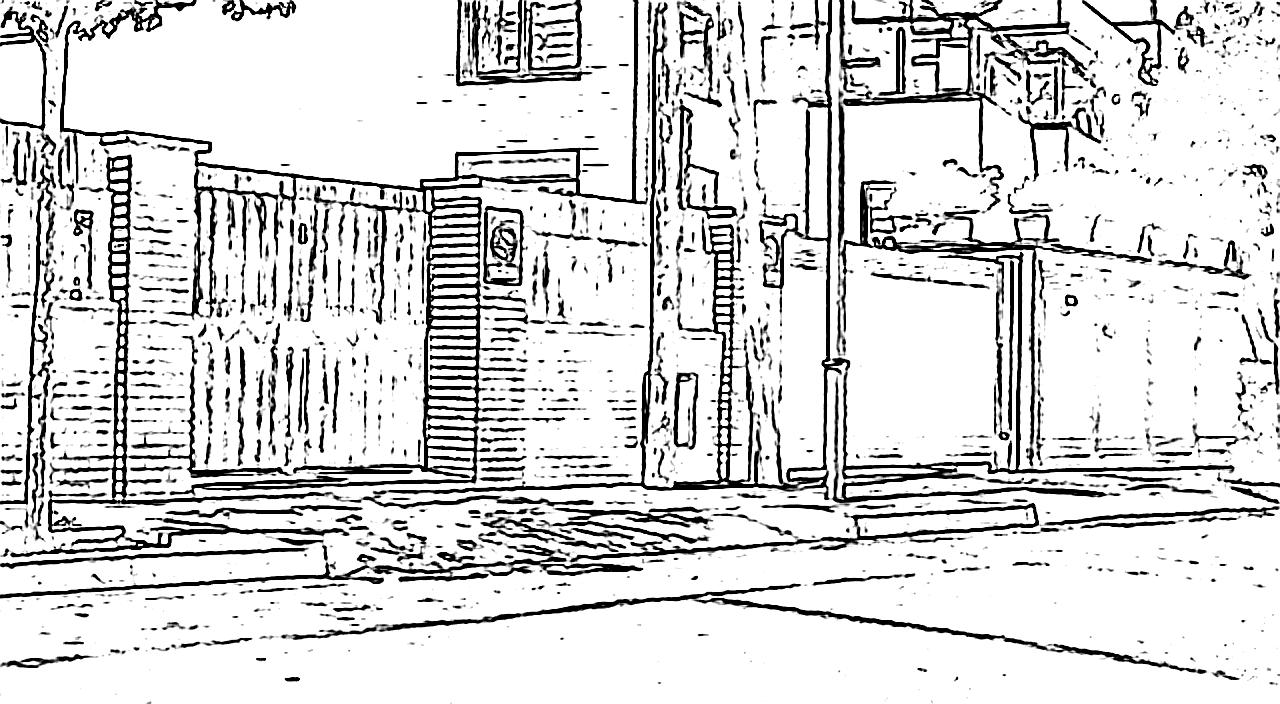} &
    \includegraphics[width=0.18\linewidth]{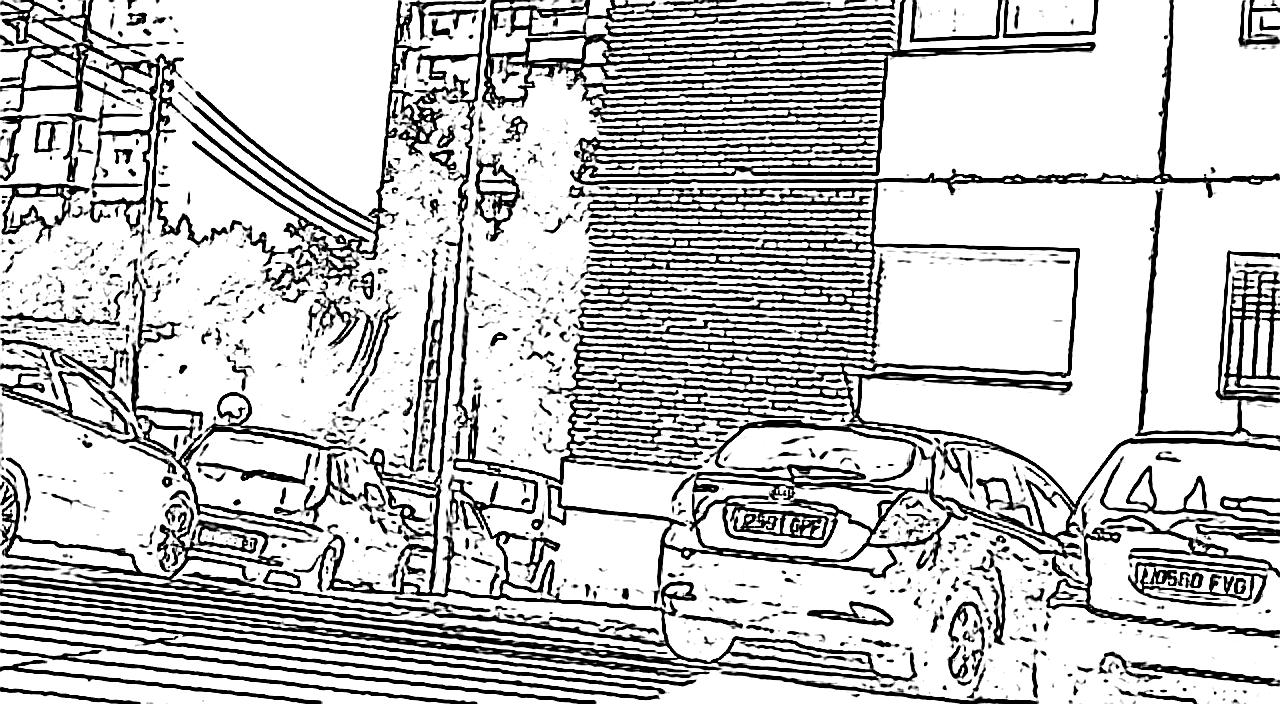} & 
    \includegraphics[width=0.18\linewidth]{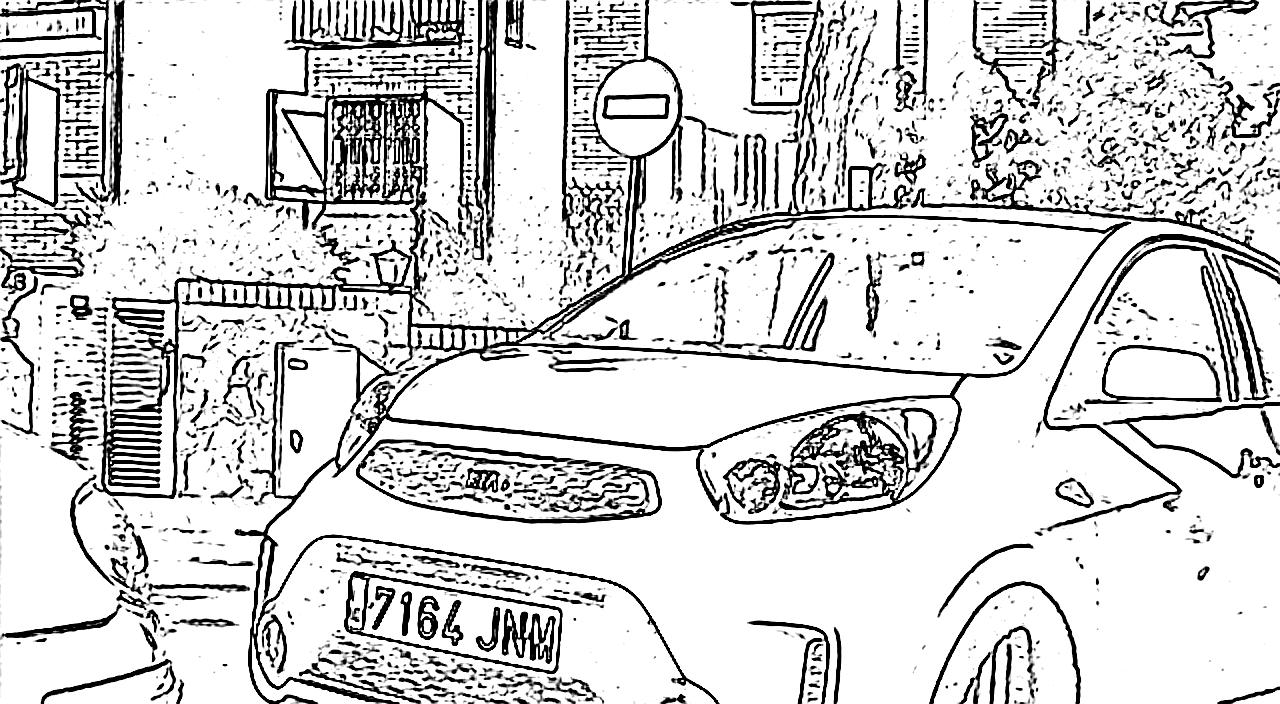} & 
    \includegraphics[width=0.18\linewidth]{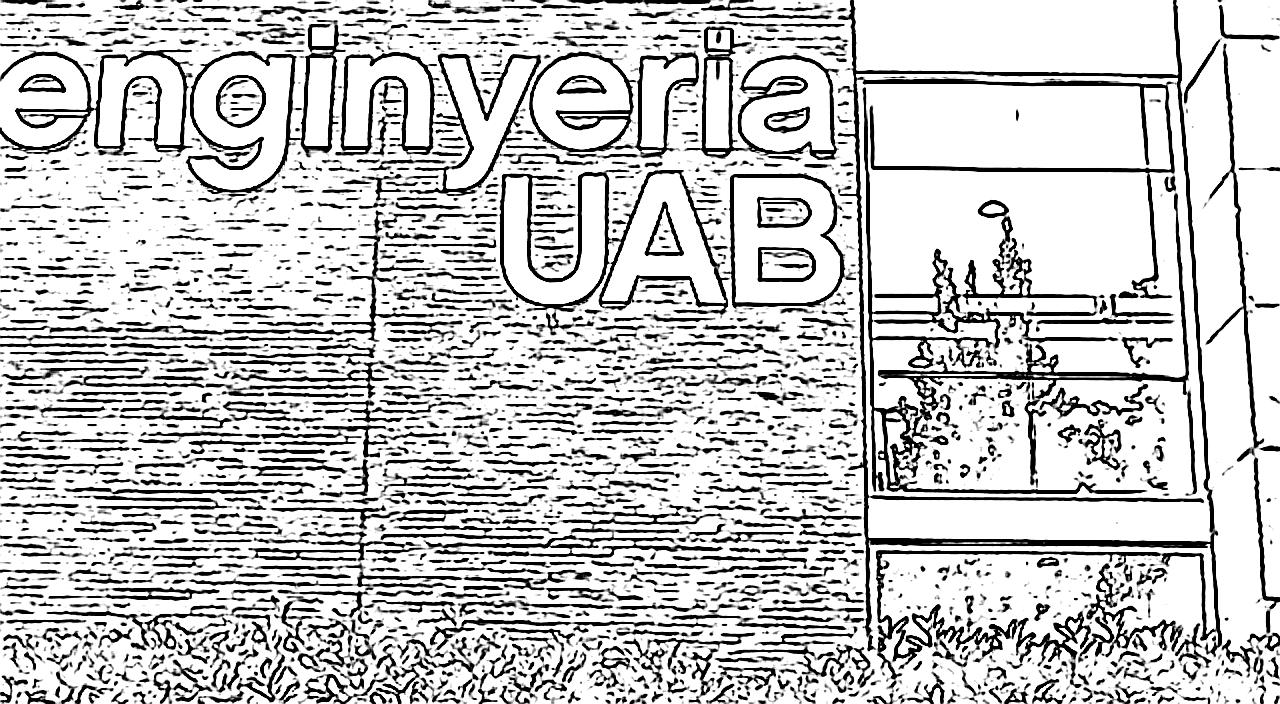} &
    \includegraphics[width=0.18\linewidth]{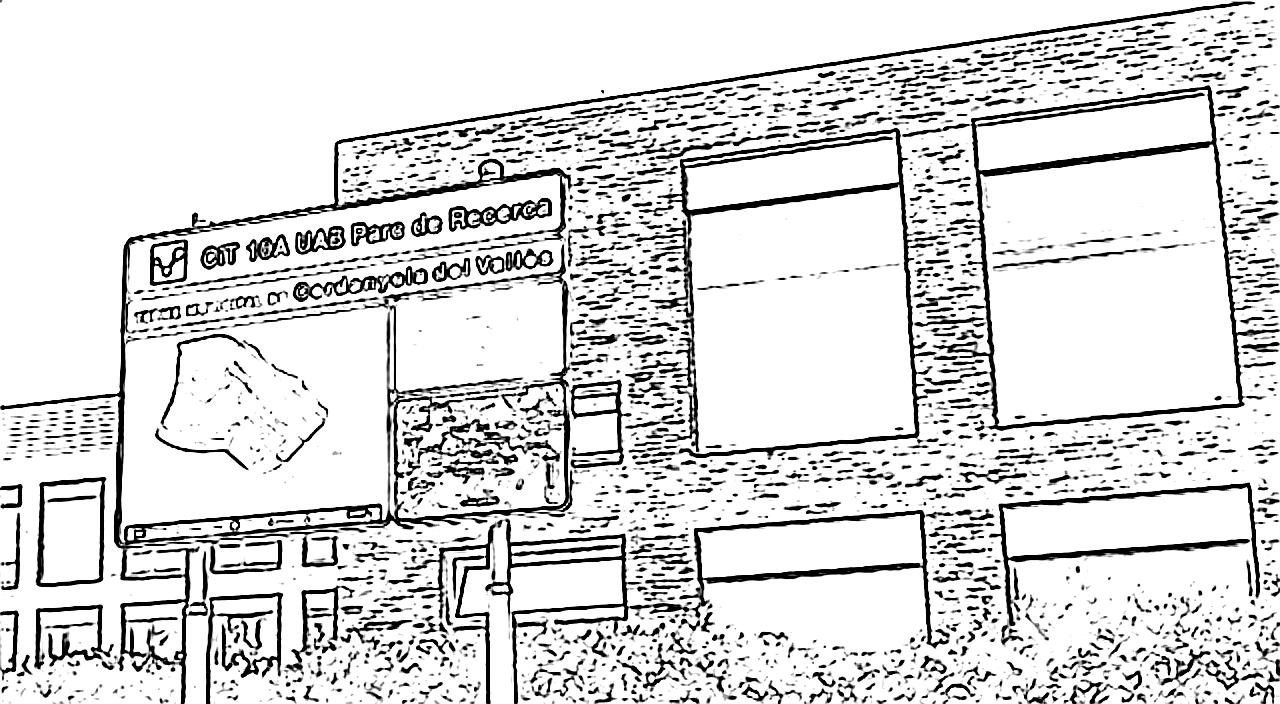}
    \\

   \raisebox{0.8\normalbaselineskip}[0pt][0pt]{\rotatebox[origin=c]{0}{(e)}} &  
    \includegraphics[width=0.18\linewidth]{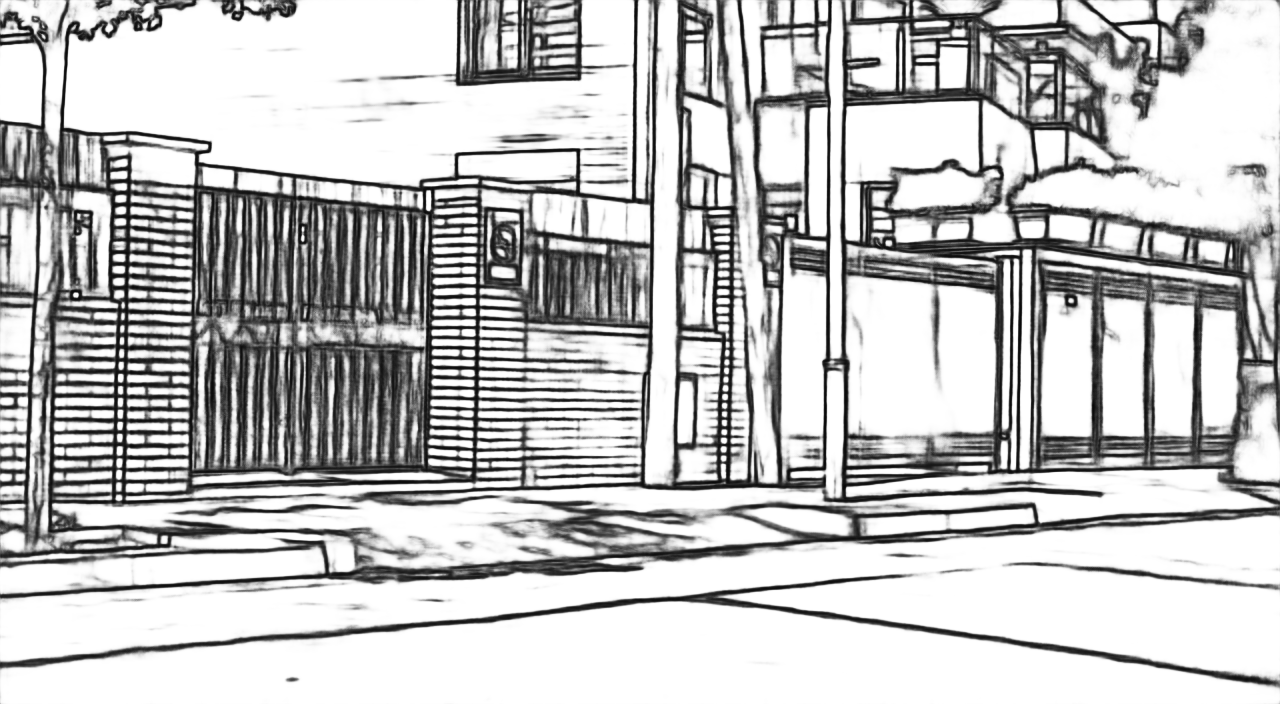} &
    \includegraphics[width=0.18\linewidth]{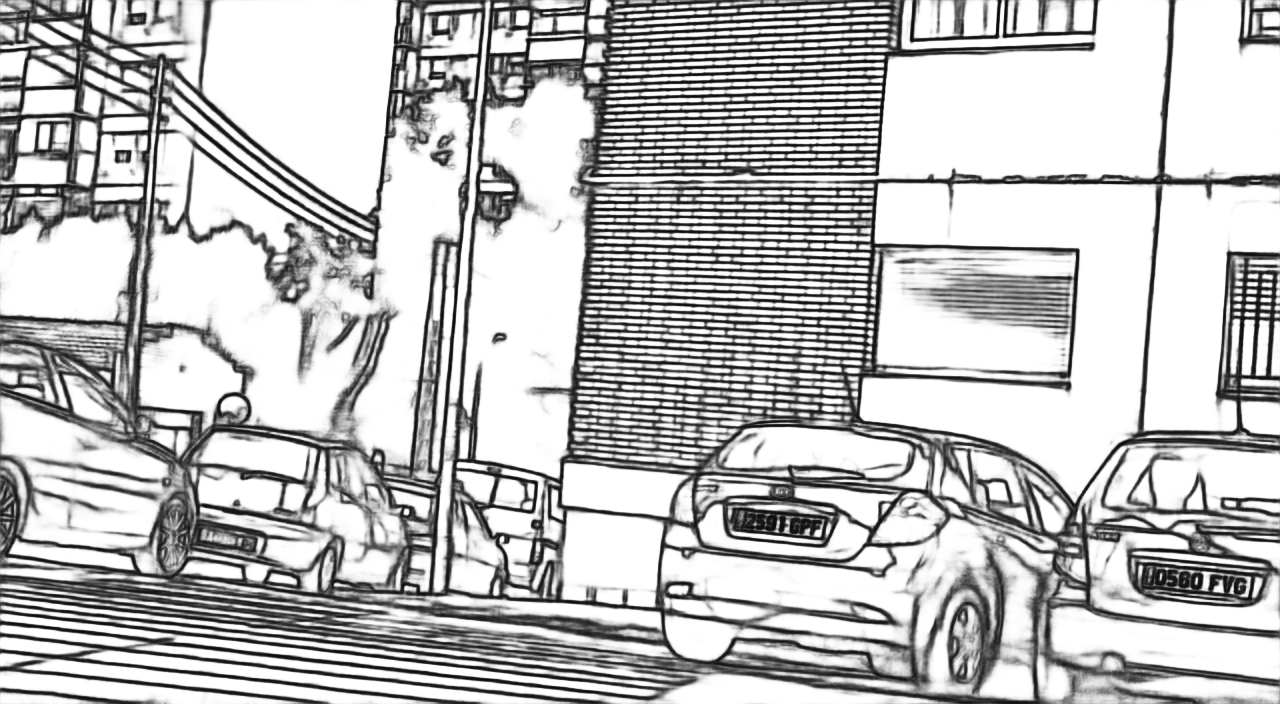} & 
    \includegraphics[width=0.18\linewidth]{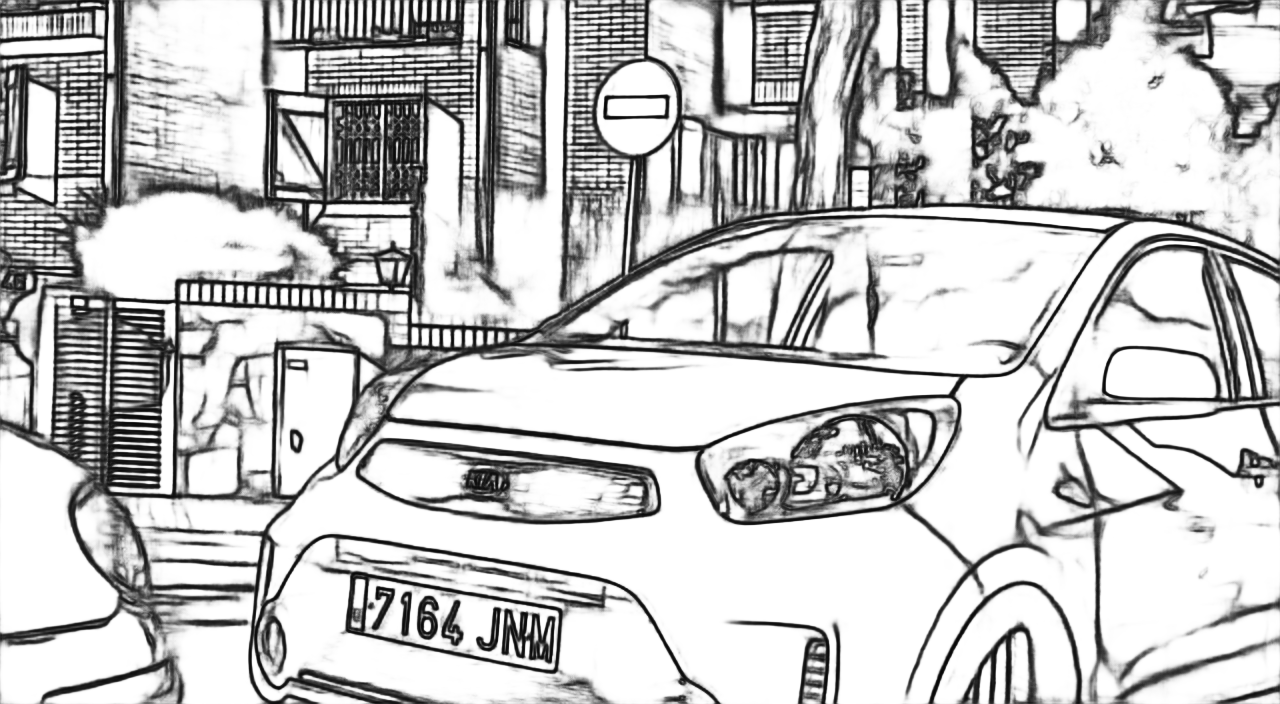} & 
    \includegraphics[width=0.18\linewidth]{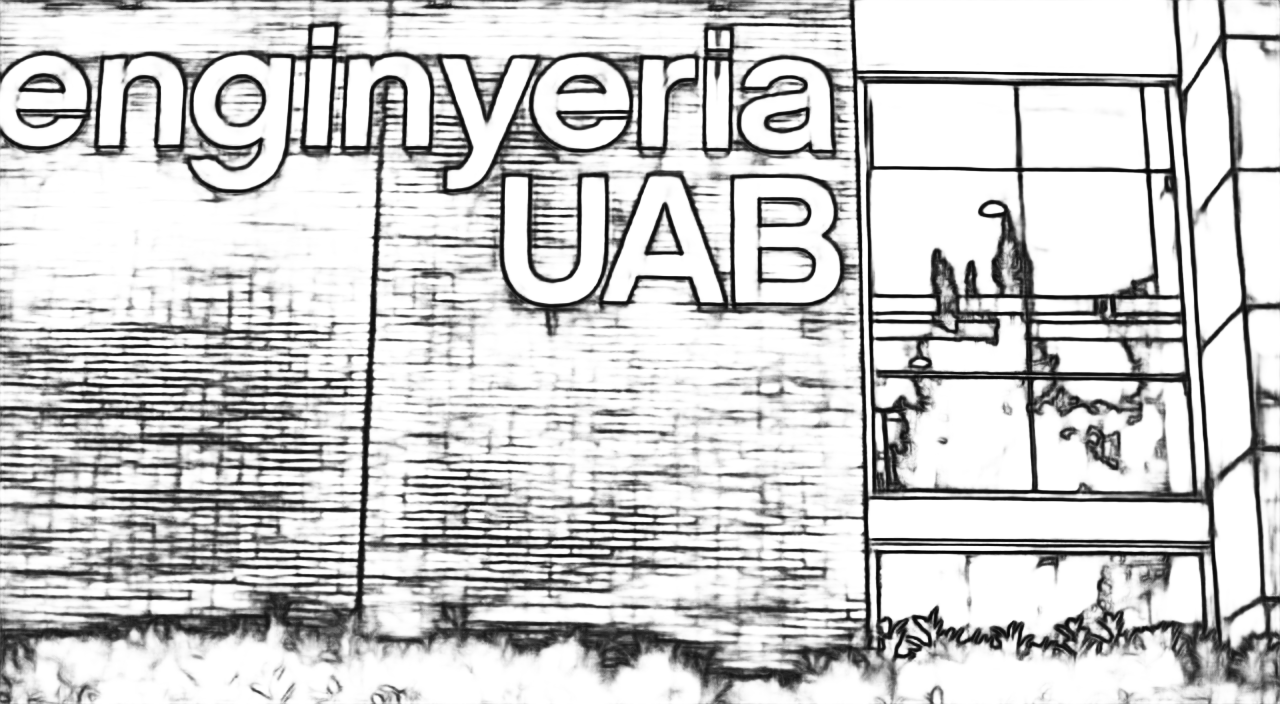} &
    \includegraphics[width=0.18\linewidth]{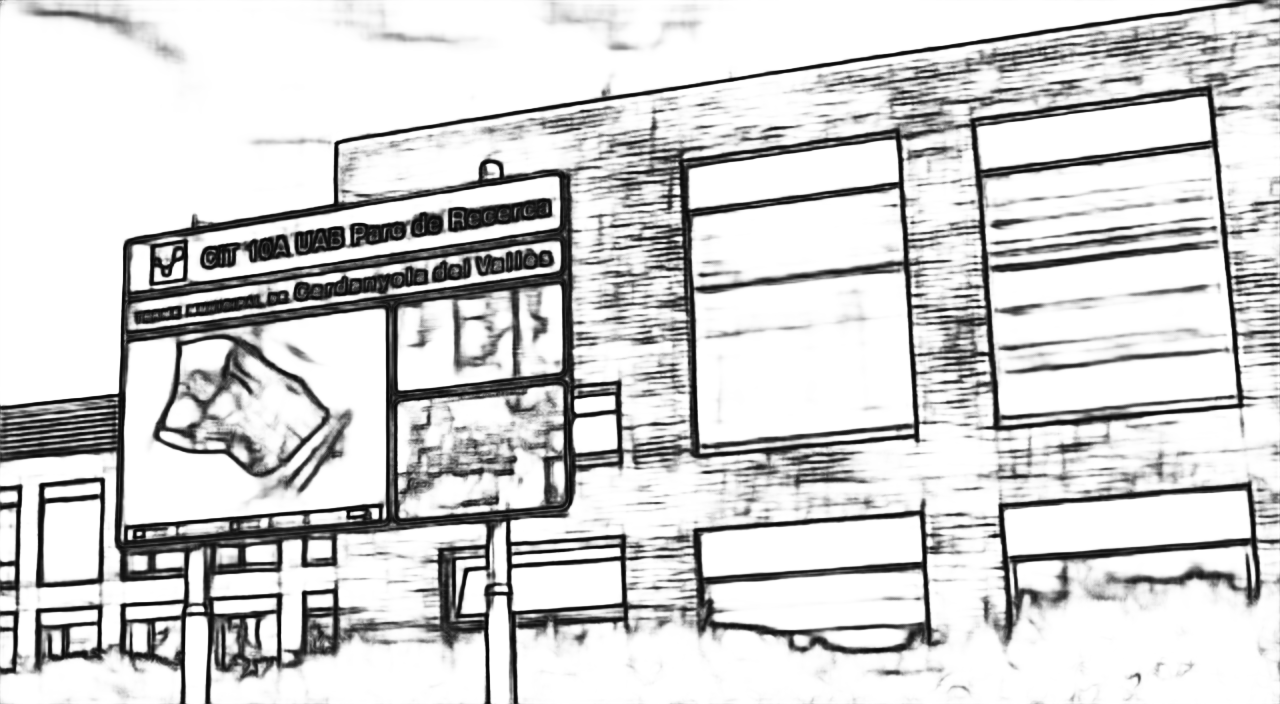} 
    \\

   \raisebox{0.8\normalbaselineskip}[0pt][0pt]{\rotatebox[origin=c]{0}{(f)}} &  
    \includegraphics[width=0.18\linewidth]{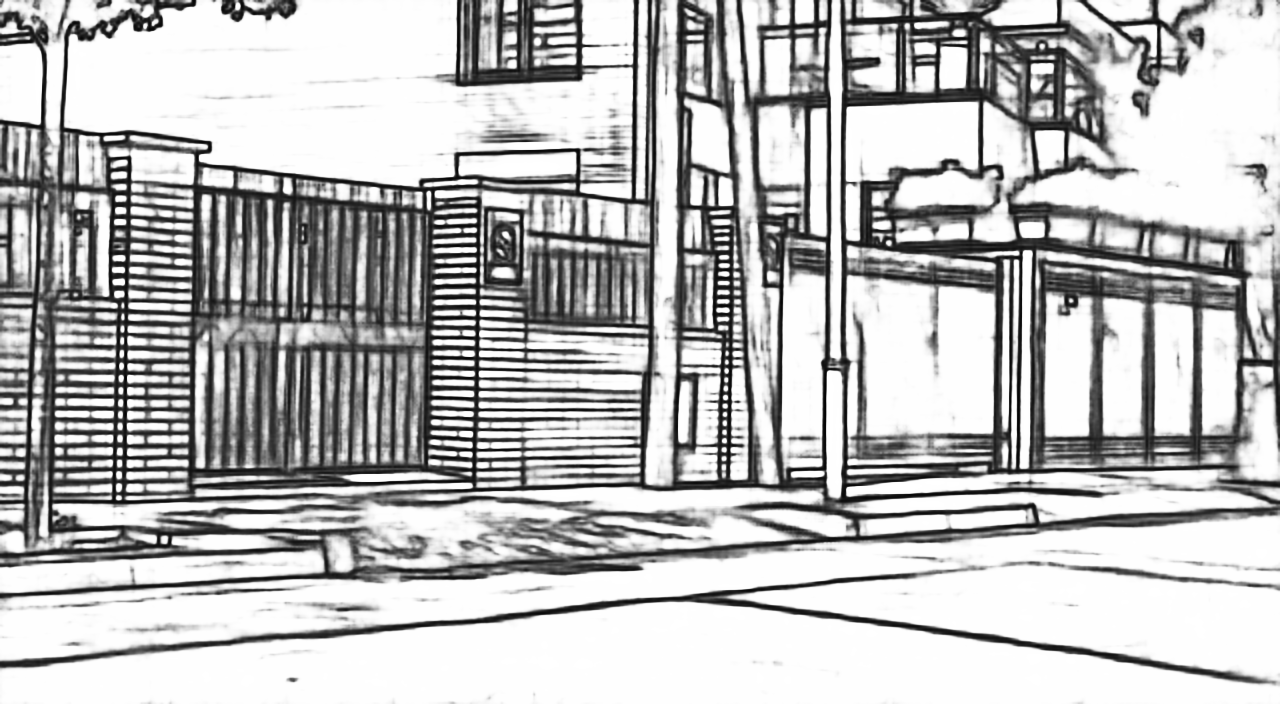} &
    \includegraphics[width=0.18\linewidth]{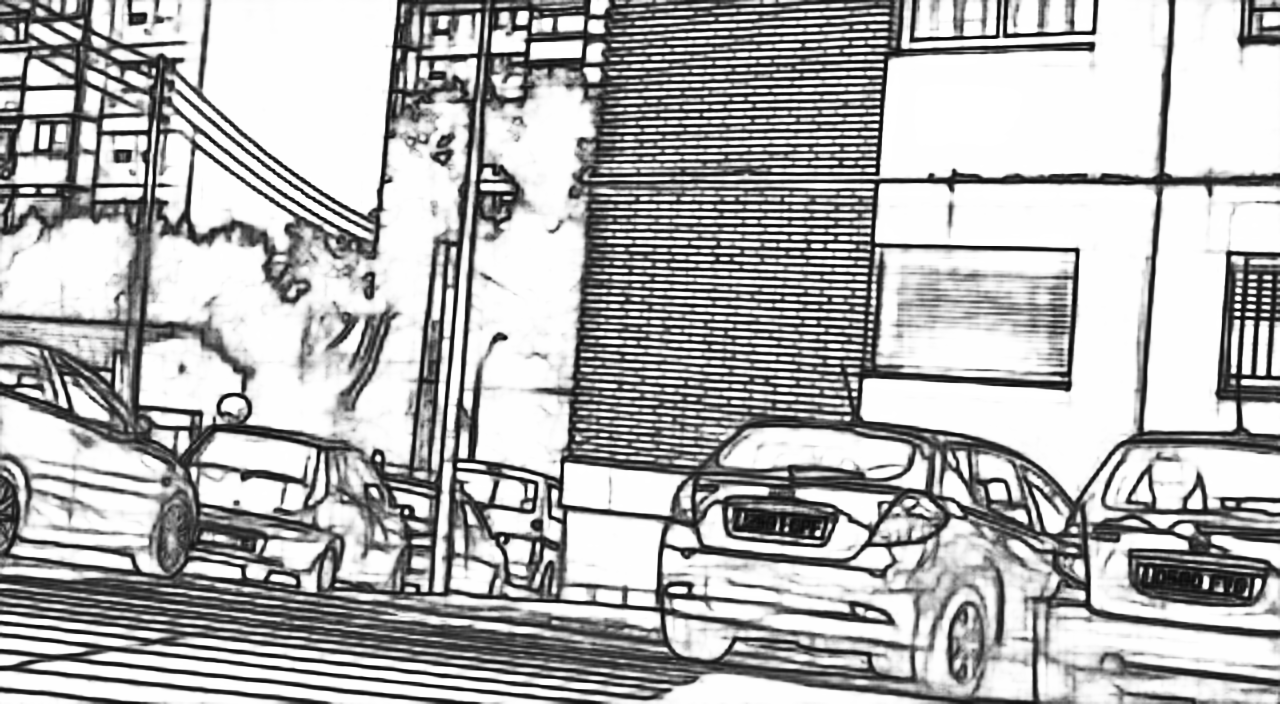} & 
    \includegraphics[width=0.18\linewidth]{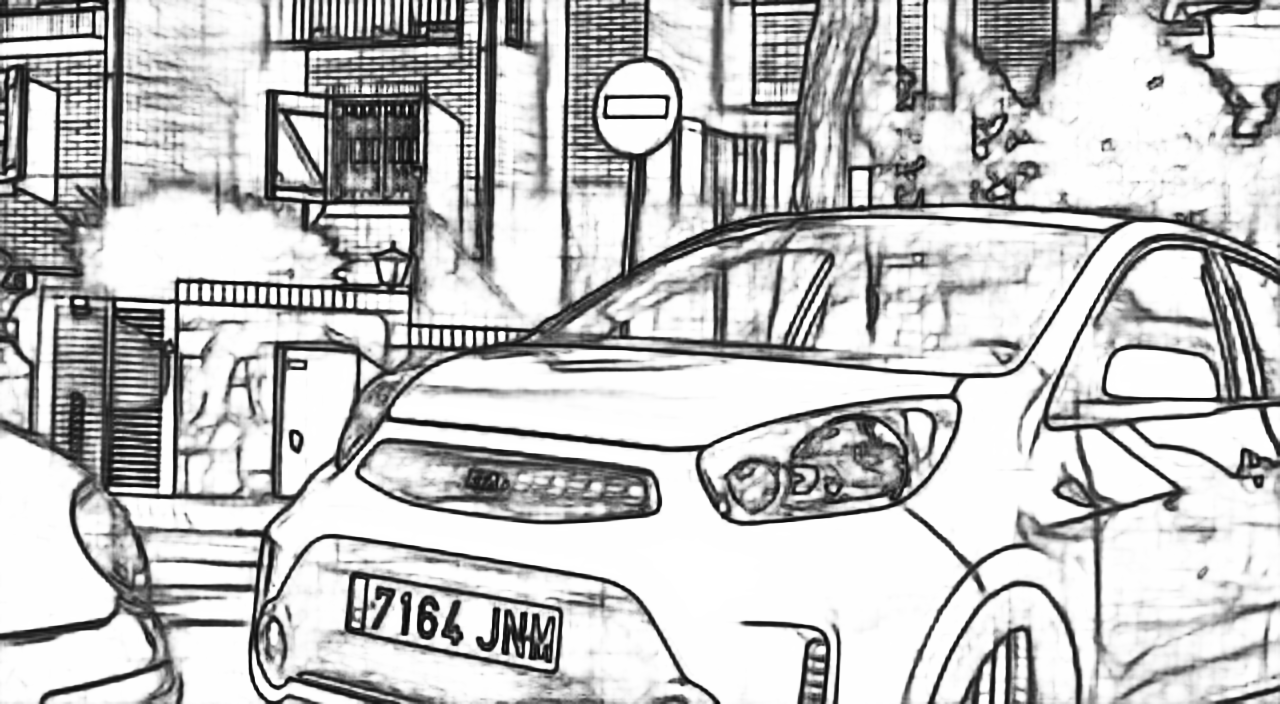} & 
    \includegraphics[width=0.18\linewidth]{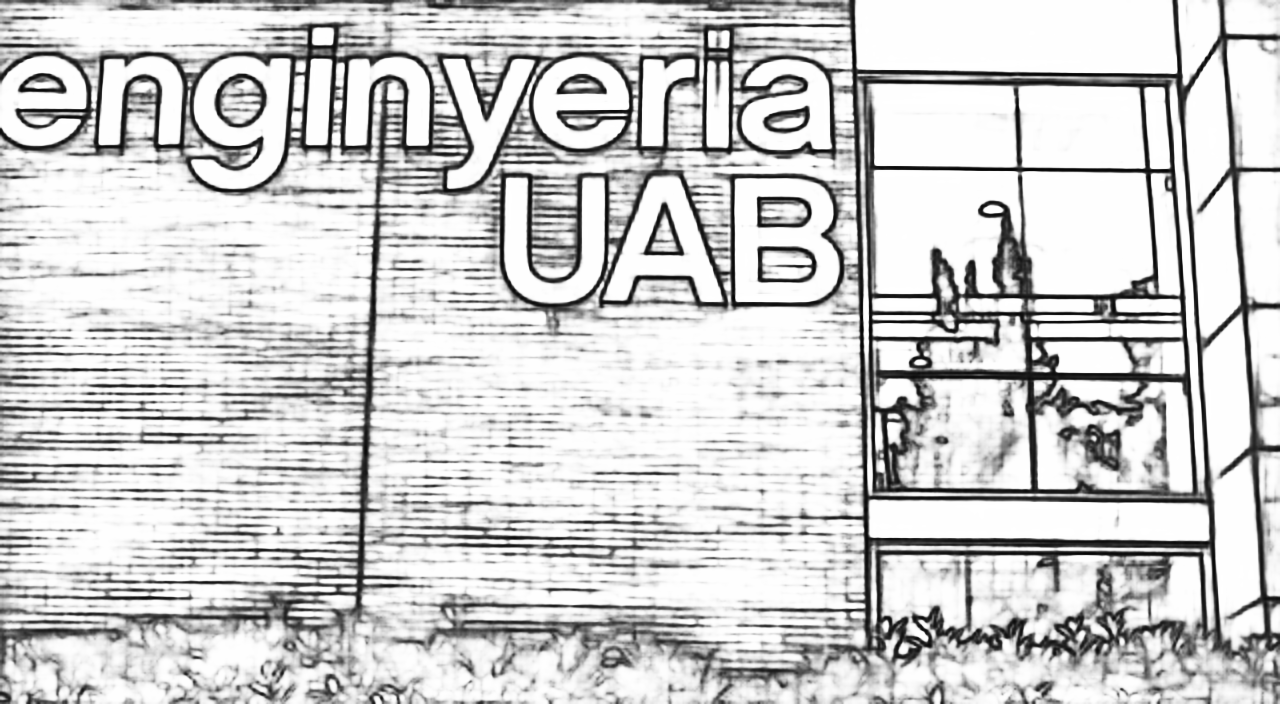} &
    \includegraphics[width=0.18\linewidth]{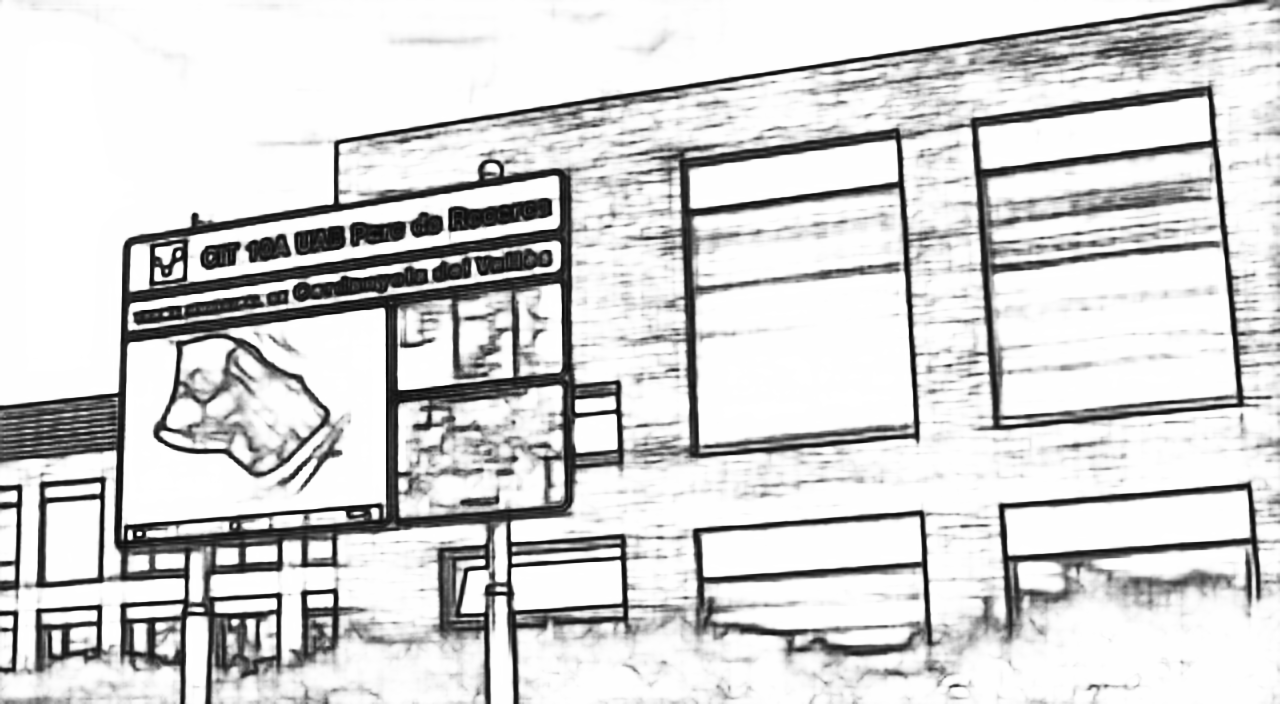}
    \\

     \raisebox{0.8\normalbaselineskip}[0pt][0pt]{\rotatebox[origin=c]{0}{(g)}} &  
    \includegraphics[width=0.18\linewidth]{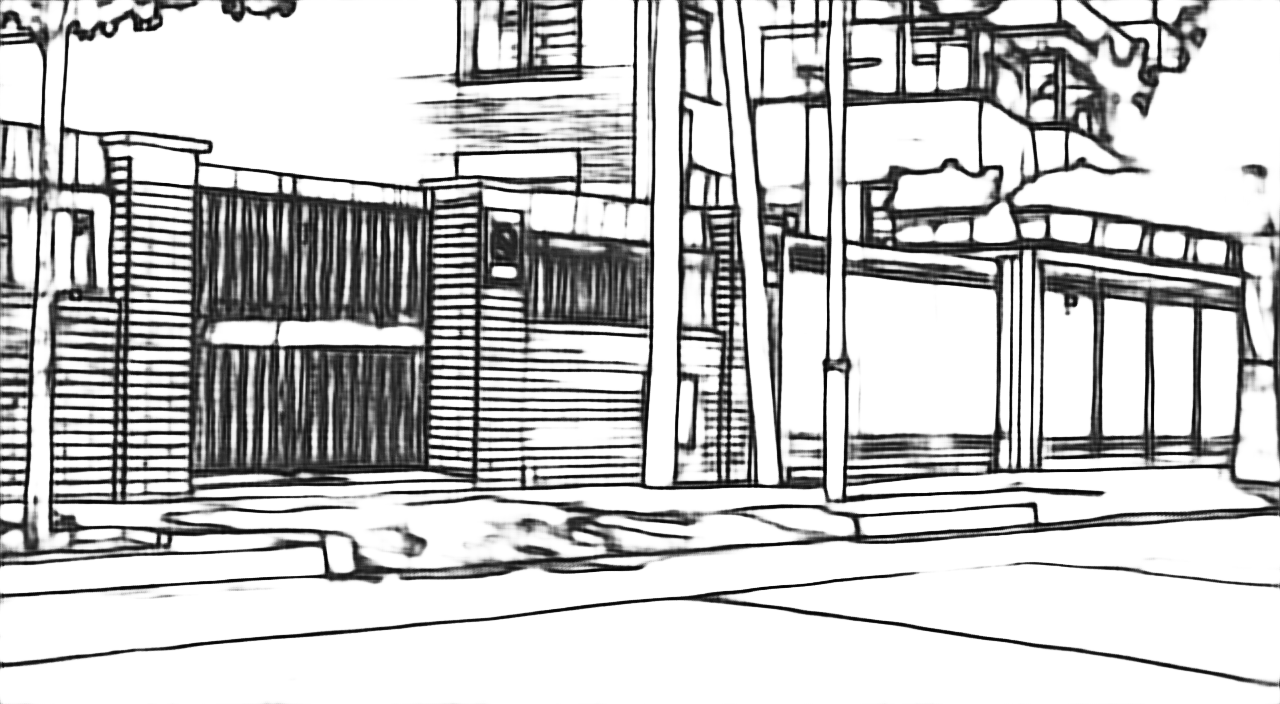} &
    \includegraphics[width=0.18\linewidth]{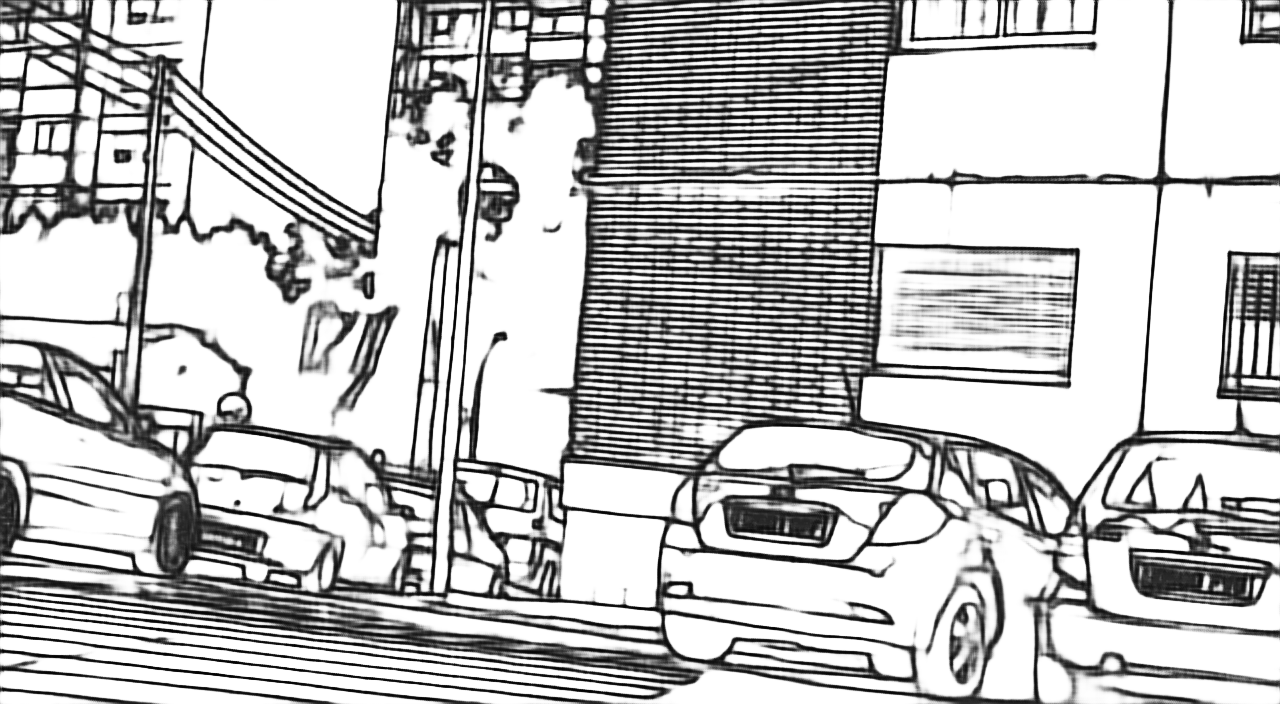} & 
    \includegraphics[width=0.18\linewidth]{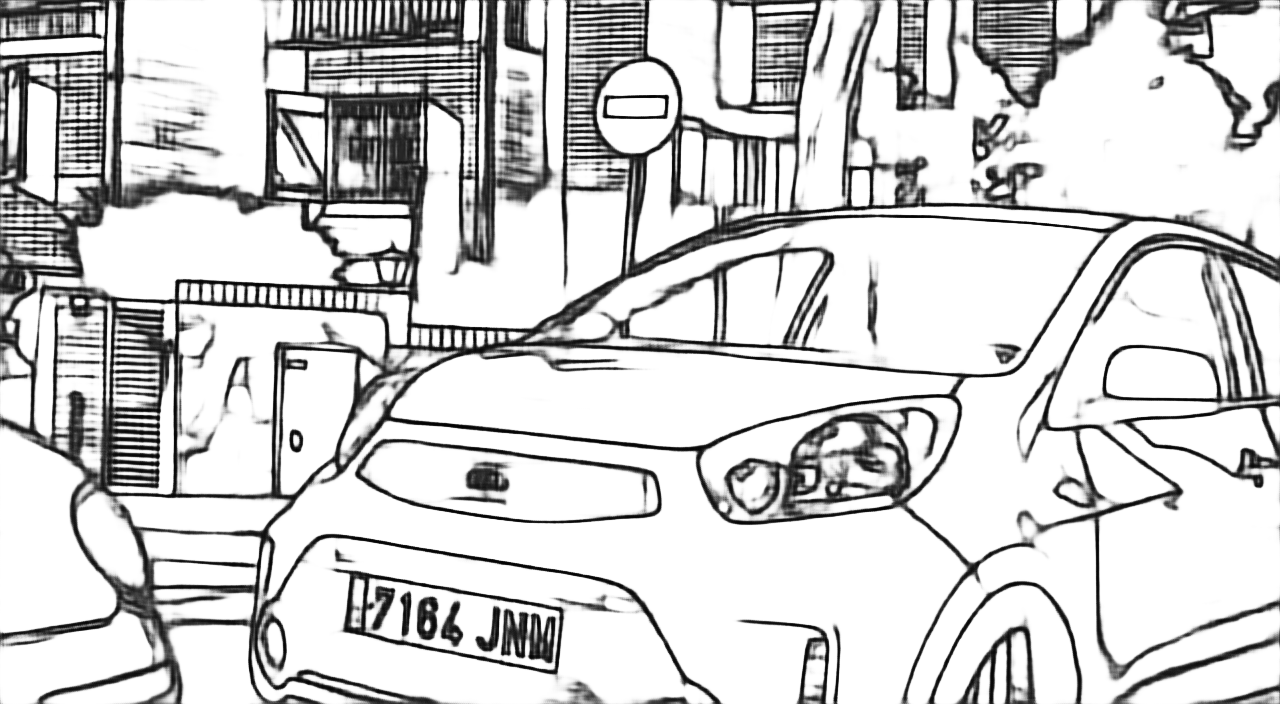} & 
    \includegraphics[width=0.18\linewidth]{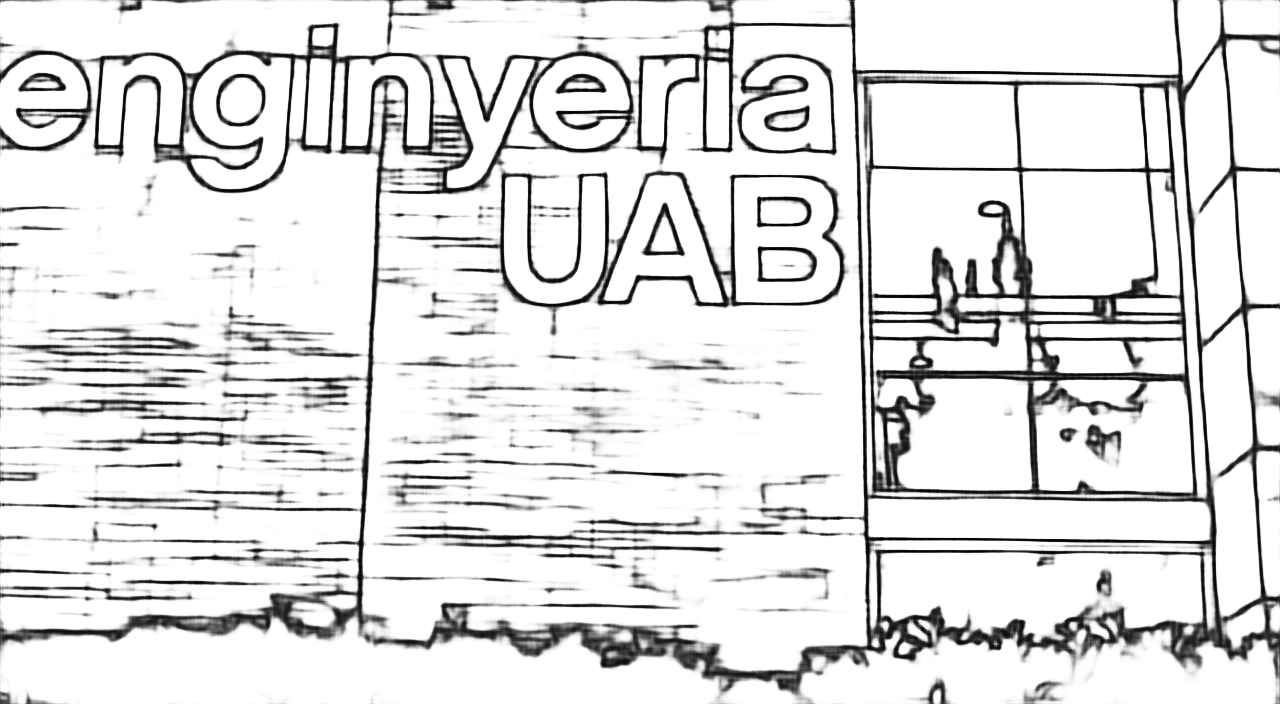} &
    \includegraphics[width=0.18\linewidth]{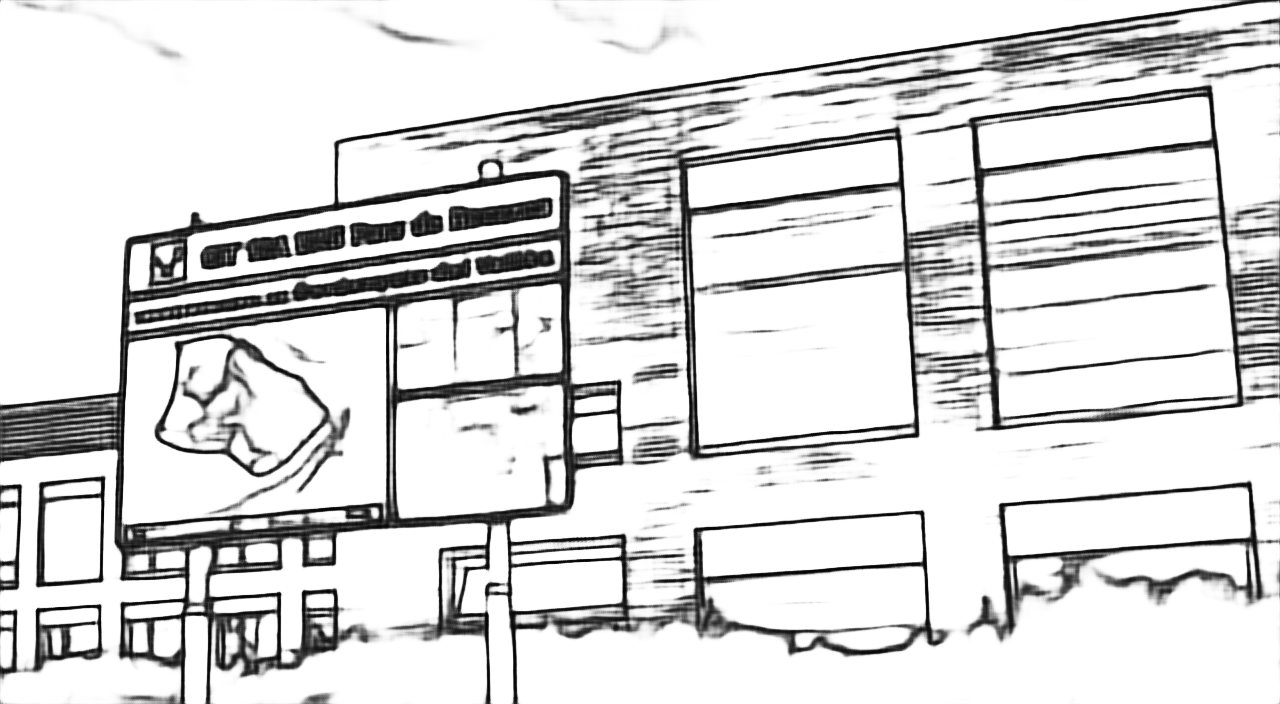}
    \\

   % \raisebox{2.5\normalbaselineskip}[0pt][0pt]{\rotatebox[origin=c]{0}{(k)}} &  
    %\includegraphics[width=0.14\linewidth]{figs/biped/rcf/RGB_025_ss.png} &
    %\includegraphics[width=0.14\linewidth]{figs/biped/rcf/RGB_056_ss.png} & 
    %\includegraphics[width=0.14\linewidth]{figs/biped/rcf/RGB_071_ss.png} & 
    %\includegraphics[width=0.14\linewidth]{figs/biped/rcf/RGB_138_ss.png} & 
    %\includegraphics[width=0.14\linewidth]{figs/biped/rcf/RGB_206_ss.png} &
    %\includegraphics[width=0.14\linewidth]{figs/biped/rcf/RGB_234_ss.png} &
    %\includegraphics[width=0.14\linewidth]{figs/biped/rcf/RGB_228_ss.png}\\
\end{tabular}
\caption{ \footnotesize {\textbf{Visual results on BIPEDv2.} (a) Input image. (b) Ground truth. (c) MTS-DR-1. (d) TEED \cite{Soria2023TinyGeneralization}. (e) XYW-Net \cite{pang2024bio}. (f) Pidinet \cite{Sun2021PixelDetection}. (g) DexiNed \cite{Soria2020DenseDetection}.}
}
\label{fig:visual_biped}
\end{figure}

\subsection{Evaluation Metrics}
We implement the qualitative evaluation with four metrics: optimal dataset scale (ODS), optimal image scale (OIS), mean Accuracy, and mean IOU. ODS and OIS are both based on F-measures, which are widely used for edge detection evaluation. Given a dataset, the ODS is calculated with a fixed threshold for all images \cite{zhou2024muge}, and the OIS is the average F-score of all images with the best threshold \cite{Xu2024RevisitingDatasets}. The number of parameters and GFLOPs are used for complexity evaluation. In this work, we do not adopt the post-processing NMS for evaluation as in \cite{Liu2019RicherDetection, Sun2021PixelDetection, pang2024bio} but use the ``Thin" and ``Raw" settings directly, because our proposal can generate edges with sufficient accuracy. The tolerance values are 0.0075 and 0.0011 for BSDS500 and BIPEDv2, respectively.

%%%%%%%%%%%%%%%%%%%%%%%%%%%%%%%%%%%%%%%%%%%%%%%%%%%cr Results on bsds
\begin{table}[!htp]\centering
\begin{adjustbox}{width=0.5\textwidth,center}
\begin{tabular}{l|cccccc}
\hline
\hline
    \multirow{2}{*}{\thead{CR}}&
    \multicolumn{2}{c}{\thead{Thin}} &
    \multirow{2}{*}{\thead{mean\\ Precision}}&
    \multirow{2}{*}{\thead{mean\\ IOU}} &
     \multirow{2}{*}{\thead{Params}}&
    \multirow{2}{*}{\thead{GFLOPs}}\\
\cline{2-3}
& ODS  & OIS  & &  \\\midrule
0.2    &\textcolor{blue}{0.739} & \textcolor{blue}{0.754}  & \textcolor{blue}{0.8436} & \textcolor{blue}{0.6076} & \textcolor{blue}{1.566M}&\textcolor{blue}{13.380}  \\
0.4    & 0.734&0.750  &0.8404  & 0.5751 &1.580M &15.112    \\
\hline
\hline
\end{tabular}
\end{adjustbox}
\caption{Results Comparison with MTS-DR-1 using different compressing ratios on BSDS500 }\label{tab: bsds_cr}
\end{table}

%%%%%%%%%%%%%%%%%%%%%%%%%%%%%%%%%%%%%%%%%%%%%%%%%%%cr Results on biped
\begin{table}[!htp]\centering
\begin{adjustbox}{width=0.5\textwidth,center}
\begin{tabular}{l|cccccc}
\hline
\hline
    \multirow{2}{*}{\thead{CR}}&
    \multicolumn{2}{c}{\thead{Thin}} &
    \multirow{2}{*}{\thead{mean\\ Precision}}&
    \multirow{2}{*}{\thead{mean\\ IOU}} &
    \multirow{2}{*}{\thead{Params}} &
    \multirow{2}{*}{\thead{GFLOPs}}
    \\
\cline{2-3}
& ODS  & OIS  & &  \\\midrule
0.2    & \textcolor{blue}{0.900}& \textcolor{blue}{0.906}&0.8643 & 0.6032&\textcolor{blue}{1.566M}&\textcolor{blue}{30.104} \\
0.4    &0.897 &0.905  &\textcolor{blue}{0.8689}& \textcolor{blue}{0.6038} &1.599M &33.999  \\
\hline
\hline
\end{tabular}
 \end{adjustbox}
\caption{Results Comparison with MTS-DR-1 using different compressing ratios on BIPEDv2 }\label{tab: biped_cr}
\end{table}

\subsection{Comparison with other SOTA works}
We compare our MTS-DR-Net with TEED \cite{Soria2023TinyGeneralization}, Pidinet \cite{Sun2021PixelDetection}, DexiNed \cite{Soria2020DenseDetection}, and XYW-Net \cite{pang2024bio}  on BSDS500 and BIPEDv2. The baseline of our method is MTS-DR-1. The best result is colored blue, and the second best is colored red.

\textbf{BSDS500:} The qualitative evaluation results are shown in Table \ref{tab:bsds500}. Our MTS-DR-2 achieves the best ODS and OIS (for both ``Thin" and ``Raw"), Mean Precision, and Mean IOU. The MTS-DR-3 and MTS-DR-4 obtain the second-best ODS and OIS. The baseline MTS-DR-1 outperforms XYW-Net \cite{pang2024bio} with ODS by $1.9\%$ and OIS by $3.1\%$ in the ``Thin" setting, by $5.9\%$ and OIS by $5.4\%$ in the ``Raw" setting. The MTS-DR-3 obtains $2.6\%$ higher Mean Precision and $2.8\%$ higher mean IOU than XYW-Net but with $40.9\%$ less parameters and $56.9\%$ lower GFLOPs.  As shown in Figure \ref{fig:visual_bsds}, the baseline produces edge maps with richer details and less noise compared to other methods. Figure \ref{fig:feature_bsds} illustrates two examples of side feature maps and MTS feature maps on BSDS500. It is noteworthy that the MTS-DR backbone can remove homogeneous features and highlight discontinuities preliminarily. Hence, the raw edge maps learned from the MTS feature maps achieve a comparable performance as shown in Table \ref{tab:bsds500}. 

\textbf{BIPEDv2:} The results on BIPEDv2 are illustrated in Table \ref{tab:biped}, in which the baseline achieves the best mean IOU as $0.6038$ and the second-best mean Precision as $0.8689$.  MTS-DR-2 achieves the best ODS and OIS (for both ``Thin" and ``Raw"). The baseline outperforms XYW-Net \cite{pang2024bio} with ODS by $11.1\%$ and OIS by $3.6\%$ in the ``Raw" setting. The visual result and examples of feature maps on BIPEDv2 are illustrated in Figure \ref{fig:visual_biped} and Figure \ref{fig:feature_biped}, respectively. As shown, our method produces edge maps containing more exquisite details and less unnecessary features. In particular, the MTS feature map in \ref{fig:feature_biped} indicates the strong ability of the MTS-DR backbone to learn the required subspace.

\subsection{Ablation Studies}
To further explore the effectiveness of the MTS-DR module, we perform ablation studies using MTS-DR-Net variants and MTS-DR modules with different CR.

\textbf{Effects of Network Scalability:} Firstly, we investigate the scalability of the network with different settings of $M$ and $C$. As shown in Table \ref{tab:bsds500} and Table \ref{tab:biped}, the MTS-DR-2 with the most complex scalability has achieved the best performance using the four metrics on BSDS500 and BIPEDv2. It is noted that the small structure MTS-DR-3 still outperforms XYW-Net \cite{pang2024bio} by $2.1\%$ with ODS and by $1.8\%$ with OIS in the "Thin" setting, by $5.5\%$ with ODS and by $5.2\%$ with OIS in the "Raw" setting on BSDS500. Compared to MTS-DR-2, the computational complexity of MTS-DR-3 decreases from 19.735 GFLOPs to 4.758 GFLOPs on BSDS500 and from 44.403 GFLOPs to 10.706 GFLOPs on BIPEDv2.

\textbf{Effects of Compressing Ratio:} We further explore the effects of different CR with MTS-DR-1. We set CR as $0.2$ on BSDS500 and BIPEDv2 and make a comparison under the ``Thin" setting. As shown in Table \ref{tab: bsds_cr} and Table \ref{tab: biped_cr}, the performance of ODS and OIS with $CR=0.2$ is better on both datasets. In addition, the computational complexity of the network is lower with a smaller CR. 

\section{Conclusions}
In this work, we present a multi-scale tensorial summation and dimensional reduction guided neural network for edge detection. The proposed MTS-DR-Net comprises two modules: an MTS-DR backbone and a lightweight refinement network. The MTS-DR backbone neglects the influence of unnecessary features and further emphasizes the importance of edges with effectiveness and efficiency. Therefore, MTS-DR-Net learns the necessary subspaces representing edges without pre-training and large-scale datasets. Unlike previous related works, our method is capable of removing unnecessary information preliminarily and preserving richer edge features. Moreover, we verify the potential of the multi-scale tensorial summation factorization operator as an efficient backbone replacement of complex DNN structures. Experiments on two benchmark datasets demonstrate that MTS-DR-Net achieves SOTA performance without post-processing.
command\Authands{ and }
{\small
\bibliographystyle{ieee}
\bibliography{egbib}
}

\end{document}